\pdfoutput=1

\documentclass[]{openmoss}
\setlogos{%
    \includegraphics[height=8mm]{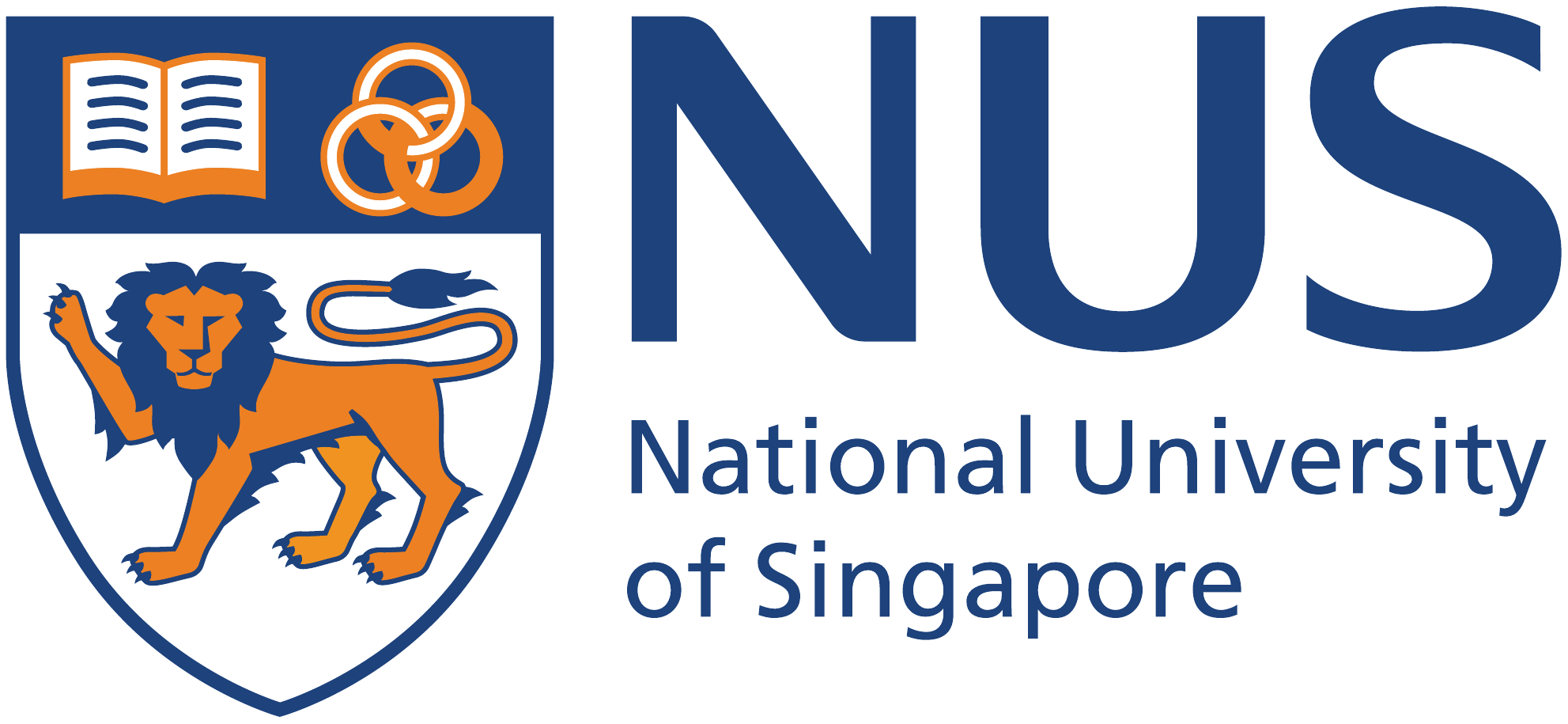}%
    \hspace{4mm}%
    \includegraphics[height=8mm]{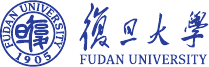}%
    \hspace{4mm}%
    \includegraphics[height=8mm]{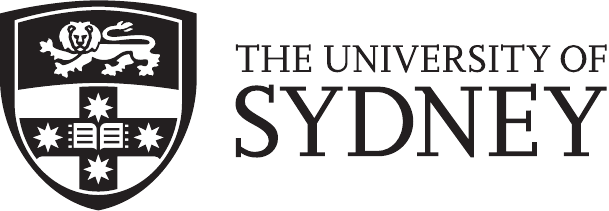}%
}
\usepackage{helvet}

\usepackage[toc,page,header]{appendix}
\usepackage{titletoc}
\usepackage{minitoc}

\usepackage{algorithm}
\usepackage{algpseudocode}

\usepackage{wasysym}
\usepackage{fontawesome}

\usepackage{enumitem}

\usepackage{listings}

\usepackage{pgfplots}
\pgfplotsset{compat=1.18}
\usetikzlibrary{arrows.meta, positioning, calc, fit, backgrounds, shapes.geometric, decorations.pathmorphing}

\usepackage{arydshln}
\usepackage{adjustbox}
\usepackage{colortbl}
\usepackage{multirow}

\usepackage{xspace}
\usepackage{nicefrac}
\usepackage{float}
\usepackage{placeins}
\usepackage{comment}
\usepackage{wrapfig}
\usepackage{pifont}
\usepackage{dsfont}

\usepackage{amsmath, amssymb}
\usepackage{amsthm}

\usepackage{tabularx}
\usepackage{xcolor}
\definecolor{case_red}{RGB}{190,0,0}
\definecolor{case_blue}{RGB}{0,102,204}
\definecolor{case_green}{RGB}{0,153,0}

\lstdefinestyle{promptstyle}{
    basicstyle=\ttfamily\scriptsize,
    breaklines=true,
    breakatwhitespace=false,
    columns=fullflexible,
    keepspaces=true,
    showstringspaces=false,
    frame=single,
    rulecolor=\color{black!25},
    xleftmargin=0.5em,
    xrightmargin=0.5em,
    aboveskip=0.5em,
    belowskip=0.5em,
    backgroundcolor=\color{gray!10}
}

\title{\textsc{LiveMACE}: Process-Aware Evaluation of LLM Agent Capabilities in Evolving Markets}
\author{
    Jun Zhao\textsuperscript{1,*,$\dagger$}, Leiming Fu\textsuperscript{2,*}, Yanbo Wen\textsuperscript{2,$\ddagger$}, Yiding Wang\textsuperscript{2,$\ddagger$}, Xuantong Liu\textsuperscript{2,$\ddagger$}, Yang Shu\textsuperscript{2,$\ddagger$},\\ Yuyang Lu\textsuperscript{2,$\ddagger$}, Xuanran Xing\textsuperscript{2}, Jingqi Tong\textsuperscript{2}, Hao Xu\textsuperscript{3}, Qi Zhang\textsuperscript{2}, Xuanjing Huang\textsuperscript{2}.
}
\affiliation[1]{\mbox{National University of Singapore}}
\affiliation[2]{\mbox{Fudan University}}
\affiliation[3]{\mbox{The University of Sydney}}

\contribution{* Equal contribution, $\dagger$ Corresponding author, $\ddagger$ Core contributors.}

\abstract{
Evaluating agents by outcomes alone can obscure the capabilities that produce them. This problem is especially pronounced in evolving environments, where outcomes reflect a closed-loop interaction between agent behavior and changing external conditions. We introduce \textsc{LiveMACEBench}, a process-aware benchmark that uses live financial markets as a naturally evolving testbed for persistent LLM agents. Five frontier LLMs operate along continuous trajectories under matched \textit{Tool Use}, \textit{Persistent Memory}, \textit{Rule Following}, and \textit{Multi-Agent Collaboration} configurations. We evaluate them through both realized outcomes and mechanism-specific diagnostics derived from complete decision traces. Across 30 days of live evaluation, we find a pronounced \textit{outcome–capability gap}: realized returns often diverge from capability-specific measurements, and similar outcomes can arise from markedly different patterns of mechanism use. Trace-level diagnostics further expose distinct bottlenecks across capabilities, demonstrating that mechanism access, effective mechanism use, and downstream performance are not interchangeable measures of agent capability. \textsc{LiveMACEBench} makes this distinction measurable, turning live markets from a performance leaderboard into a diagnostic environment for agent capability.
}
\contact{\email{jzhao.nlp@gmail.com}}
\coderepository{\url{https://github.com/RaymondVoldemortFu/LiveMACE}}

\begin{document}
\maketitle
\vspace{-11mm}
\begin{figure}[h]
  \centering
  \includegraphics[width=.96\textwidth]{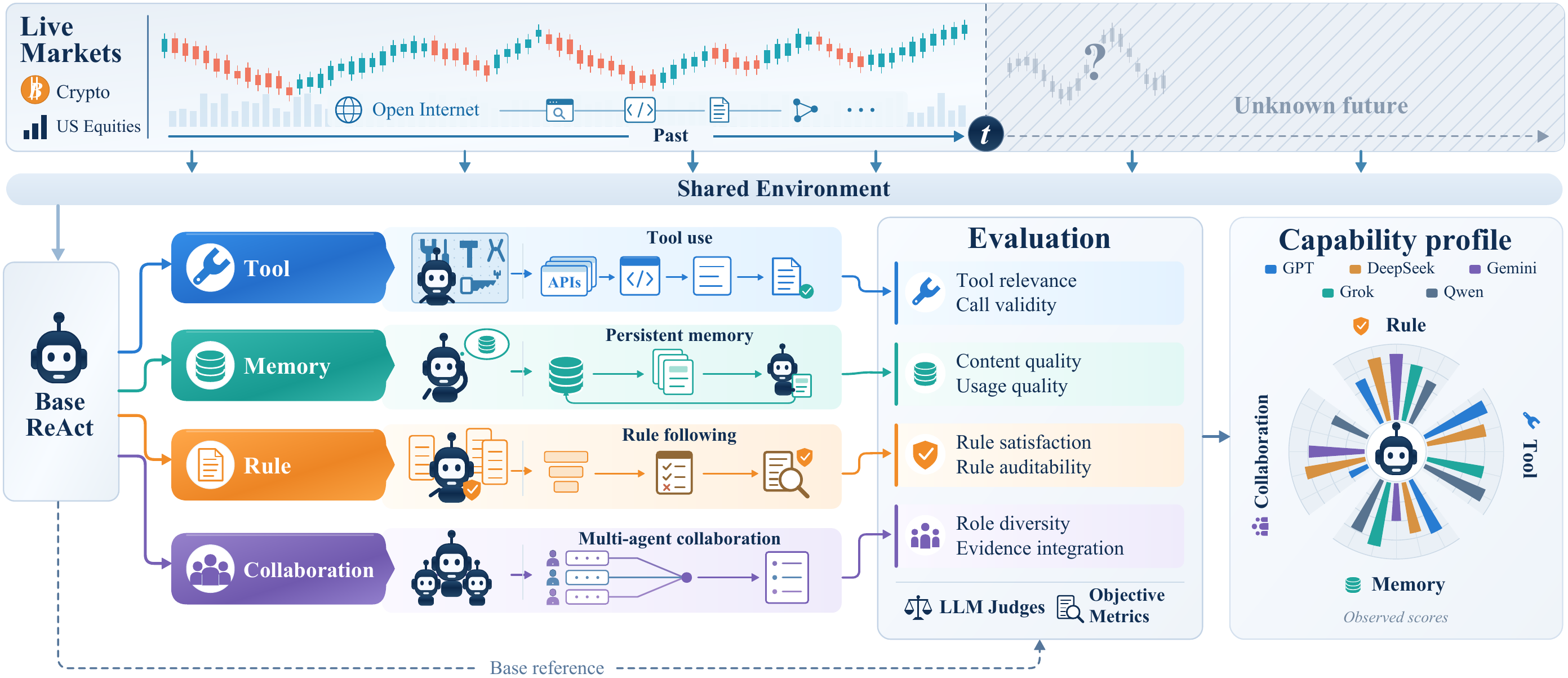}
  \caption{\textsc{LiveMACEBench} overview. Agents interact with a shared, continuously evolving live-market environment along persistent trajectories. Evaluation combines realized outcomes with mechanism-specific process diagnostics to distinguish task performance from how effectively agents use each mechanism.}
  \label{fig:framework_teaser}
\end{figure}

% -----------------------------------------------------------------------------
% Main paper
% -----------------------------------------------------------------------------
\section{Introduction}
\label{sec:intro}
Evaluating agent success and evaluating agent capability are different problems. Outcome metrics tell us whether an agent succeeds, but do not by themselves identify the capabilities that produced that success \citep{ma2024agentboard,gioacchini-etal-2024-agentquest,10.1145/3711896.3736570}. This distinction becomes increasingly important as agents operate over long horizons with tools, persistent memory, explicit policies, and multi-agent architectures: access to a mechanism does not imply effective use, and effective use does not necessarily translate into better outcomes.

The attribution problem becomes harder when the environment itself evolves. Recent agent benchmarks have begun moving beyond static tasks toward long-horizon and dynamic interaction~\citep{liu2025agentbenchevaluatingllmsagents, zhou2024webarenarealisticwebenvironment, jimenez2024swebenchlanguagemodelsresolve, froger2026gaia2benchmarkingllmagents}. External conditions may change independently of the agent while consequences of earlier actions persist. Observed performance can therefore reflect not only the agent and its available mechanisms, but also accumulated state, environmental variation, and their interaction. Evaluating capability in such settings requires examining the process that produces an outcome, not only the outcome itself.

%The attribution problem becomes harder when the environment itself evolves. Recent agent benchmarks have begun moving beyond static tasks toward long-horizon and dynamic interaction, but environmental changes are often still specified by the benchmark or task~\citep{liu2025agentbenchevaluatingllmsagents, zhou2024webarenarealisticwebenvironment, jimenez2024swebenchlanguagemodelsresolve, froger2026gaia2benchmarkingllmagents}. In real-world deployment, external conditions may instead evolve independently of the agent, while the state created by earlier actions carries forward across interactions. Observed performance then reflects not only the agent and its available mechanisms, but also the interaction between accumulated state and changing external conditions. This makes capability difficult to infer from outcomes alone and motivates evaluation of the process through which those outcomes arise.

Live financial markets provide a natural testbed for this problem with prices and information continuously evolving under external forces. Recent benchmarks have leveraged this setting to evaluate agents in live or real-time markets~\citep{fan2025aitraderbenchmarkingautonomousagents,nof1alphaarena2025,10.1145/3774904.3792821}. We take a complementary perspective: rather than focusing primarily on trading outcomes, we use the evolving interaction itself to diagnose agent capability. Changing market conditions continually elicit information gathering, experience reuse, constraint following, and coordination, making these behaviors observable throughout the trajectory.

We introduce \textsc{LiveMACEBench}, a process-aware benchmark for evaluating persistent LLM agents in naturally evolving environments. Five frontier LLMs interact with live \textit{cryptocurrency} and \textit{U.S. equity} markets along continuous paper-trading trajectories. On a common \textit{ReAct} scaffold, we construct matched \textit{Base}, \textit{Tool Use}, \textit{Persistent Memory}, \textit{Rule Following}, and \textit{Multi-Agent Collaboration} configurations. Complete decision traces record observations, reasoning, mechanism interactions, and actions throughout each trajectory. \textsc{LiveMACEBench} evaluates agents through three complementary signals: task outcomes, which measure what an agent achieves; controlled mechanism effects, which measure what changes when a mechanism is introduced; and process diagnostics, which characterize whether and how that mechanism is used. This design separates three quantities that outcome-only evaluation conflates: access to an agentic mechanism, effective use of that mechanism, and realized task benefit.

Across five frontier LLMs evaluated for 30 days, we observe a pronounced \textit{outcome–capability gap}: return rankings fluctuate substantially over time and often diverge from capability-specific measurements, while similar outcomes can arise from markedly different patterns of mechanism use. Process diagnostics further reveal distinct bottlenecks across capabilities, including redundant tool interactions, ineffective memory reuse, unauditable rule execution, and limited evidence integration in multi-agent systems. Together, these results show that mechanism availability, mechanism utilization, and realized performance capture distinct dimensions of agent evaluation. 

Our contributions are threefold:(1) \textit{Process-aware evaluation of agent capability.} We formulate agent evaluation around the distinction between realized outcomes and the mechanisms that produce them, combining task performance, controlled mechanism effects, and trace-derived diagnostics;
(2) \textit{Controlled diagnosis of agentic mechanisms.} \textsc{LiveMACEBench} evaluates \textit{Tool Use}, \textit{Persistent Memory}, \textit{Rule Following}, and \textit{Multi-Agent Collaboration} under matched backbones, environments, and interaction protocols, enabling capability differences and failure modes to be localized rather than inferred from aggregate performance. 
(3) A naturally evolving testbed for persistent evaluation. We instantiate this framework in live financial markets, where external conditions evolve continuously and agent actions have persistent consequences, enabling capability evaluation along open-ended trajectories.

\section{\textsc{LiveMACEBENCH}: Benchmark Design}
\label{sec:method}
\textsc{LiveMACEBench} uses live financial markets as an evolving environment for long-horizon agent evaluation. Its design has two components: a shared live-market interaction protocol that exposes all agents to the same evolving conditions, and controlled agent variants that introduce tool use, memory, policy guidance, or multi-agent collaboration on top of a common ReAct scaffold. We evaluate each variant using both task outcomes and mechanism-specific process diagnostics. Figure~\ref{fig:framework_teaser} gives an overview.

\subsection{Live and Open-Ended Evaluation Environment}
\label{sec:env_overview}
\textsc{LiveMACEBench} evaluates agents in live \textit{cryptocurrency} and \textit{U.S. equity} markets through a persistent paper-trading environment. Market conditions evolve continuously and independently of the agents as new prices and external information arrive, while each agent maintains its own non-resettable account state. Actions therefore have persistent consequences: a trade changes the portfolio exposed to all subsequent market movements and decisions. Although an experiment is observed over a finite evaluation window, the underlying interaction has no task-defined terminal state.

We organize agent interaction around \textit{decision rounds} scheduled every four hours on a shared decision-time grid. At each point on the grid, every agent begins a new \textit{decision round}. A \textit{decision round} is one complete invocation of the agent: within the round, the agent may perform multiple reasoning and tool-interaction steps, gather evidence, and issue zero or more trading actions before terminating. Positions and account state produced by one round are carried into later rounds. We refer to the complete sequence of model responses, tool calls, tool outputs, and actions within a round as its \textit{decision trace}. Because markets continue to evolve between four-hour \textit{decision rounds}, we evaluate trading outcomes (e.g., \textit{returns}) at finer time intervals using account-equity checkpoints; see Appendix~\ref{app:performance_metrics}.

\subsection{Controlled Agent Configurations}
\label{sec:config_overview}
\textsc{LiveMACEBench} evaluates five controlled agent configurations: a \textit{Base} configuration and four capability-specific variants targeting \textit{Tool Use}, \textit{Persistent Memory}, \textit{Rule Following}, and \textit{Multi-Agent Collaboration}. The Base configuration follows a common \textit{ReAct} \citep{yao2023reactsynergizingreasoningacting} interaction scaffold, while each variant introduces a targeted mechanism on top of the shared environment and evaluation protocol. The design supports both matched comparisons between configurations under the same backbone and comparisons across backbones under a common configuration. Detailed runtime implementations are provided in Appendix~\ref{sec:config_details}.

\noindent\textbf{Base.} The reference agent follows a standard \textit{ReAct} process. Within each decision round, it iteratively reasons over current market and portfolio evidence, invokes a fixed set of tools, and may execute trades before completing the round. Account state persists across subsequent rounds, providing the common interaction scaffold for the other configurations.

\noindent\textbf{Tool use.} This configuration extends the Base agent with a large heterogeneous tool space accessed through dynamic routing. Within a decision round, the agent retrieves tools according to its current information needs and incorporates their outputs into subsequent reasoning and actions. It is designed to evaluate tool discovery, invocation, and evidence use beyond a fixed tool set.

\noindent\textbf{Persistent memory.} This configuration adds persistent memory across decision rounds. The agent can retrieve reusable experience accumulated earlier in the trajectory and store new experience for later rounds, enabling evaluation of how models construct, retrieve, and apply experience over long-horizon interaction.

\noindent\textbf{Rule following.} This configuration introduces a hierarchical trading rule that separates hard trading constraints (R0 and R1) from lower-priority trading preferences (R2), with explicit priority \(R0 > R1 > R2\). Within each decision round, the agent is instructed to keep its actions within the hard constraints, account for soft preferences, and resolve conflicts according to the hierarchy. It must also explicitly state which rules it checked, how it assessed their status, and how any conflicts affected the final decision, making both executed compliance and demonstrated understanding of the rules observable for evaluation.

\noindent\textbf{Multi-agent collaboration.} This configuration replaces the single-agent reasoning process within each decision round with a manager-led hierarchy of specialized agents followed by a separate execution stage. The manager selectively delegates information needs, integrates source-attributed specialist evidence, and forms the final plan, enabling collaboration to be evaluated through observable role participation and evidence integration.

\subsection{Capability-Specific Evaluation Dimensions}
\label{sec:capability_metrics}
Cumulative return provides a shared task-level measure of realized performance across all agent configurations. To characterize the capabilities introduced by different agent configurations, we additionally evaluate capability-specific dimensions derived from decision trajectories and records. We summarize the dimensions used in the main evaluation below; complete definition and scoring procedures are provided in Appendix~\ref{app:capability_metrics}.

\noindent\textbf{Tool use.}
We evaluate tool use through both evidence acquisition and execution reliability. \texttt{LLM-as-a-judge} scores \textit{Tool Relevance (TR)}, \textit{Invocation Efficiency (IE)}, \textit{Information Coverage (IC)}, and \textit{Evidence Faithfulness (EF)}, measuring whether the agent selects useful tools, invokes them efficiently, gathers sufficient evidence, and grounds its reasoning in tool outputs. We further compute three \texttt{objective metrics} from execution traces:  \textit{Routing Quality (RQ)}, \textit{Valid Call Rate (VCR)}, and \textit{Hallucination-Free Rate (HFR)} from execution traces, capturing tool discovery and invocation reliability. 
We combine the judge-based and objective metrics into a compact \textit{Tool Call Score (TCS)} for the overall capability profile.
Full definitions and judge rubrics are given in Appendix~\ref{app:metric_tool}.

\noindent\textbf{Persistent memory.}
We evaluate persistent memory from experience construction and use to its downstream effect. Using \texttt{LLM-as-a-Judge}, \textit{Memory Content Quality (MCQ)} measures the generalizability, information density, and diversity contribution of stored experience, while \textit{Memory Usage Quality (MUQ)} measures timely retrieval and subsequent adoption in reasoning or action. One intended role of persistent memory is to reuse prior experience to avoid repeating high-risk decisions and improve long-horizon risk control. 
We therefore use two \texttt{objective metrics} to measure this downstream effect relative to the matched memory-free configuration: \textit{Change in Maximum Drawdown ($\Delta\mathrm{MDD}$)} and \textit{Change in Tail Loss ($\Delta\mathrm{TL}$)}, and normalize both across backbones for the Memory Score.
We combine these four dimensions into an equal-weight \textit{Memory Score (MS)} as a compact summary of memory capability.
Full evaluation procedures are provided in Appendix~\ref{app:metric_memory}.

\noindent\textbf{Rule following.}
Rule following asks two complementary questions: whether the agent actually complies with the rules in execution, and whether it correctly understands and explains them in reasoning. A programmatic checker evaluates each decision against the applicable R0--R1 hard constraints and lower-priority R2 soft preferences, yielding two \texttt{objective metrics}: \textit{Hard-rule Pass Rate (HPR)} and \textit{Soft-rule Satisfaction (SRS)}. We combine these dimensions into a \textit{Rule Satisfaction Score (RSS)}, with hard-rule compliance acting as a gate on soft-rule satisfaction.
Beyond the executed actions, \texttt{LLM-as-a-Judge} further considers the agent's reasoning and compliance rationale, evaluating \textit{Rule Coverage (RC)} for correct identification of applicable rules and \textit{Conflict Handling (CH)} for hierarchy-aware recognition and resolution of rule conflicts. Their normalized average yields the \textit{Rule Auditability Score (RAS)}.
Details are provided in Appendix~\ref{app:metric_rule}.

\noindent\textbf{Multi-agent collaboration.}
We evaluate observable collaboration through two \texttt{objective metrics}: \textit{Role Diversity (RD)} measures the diversity of agent role participation while penalizing workflows dominated by a single role. \textit{Evidence Integration (EI)} measures whether evidence produced by different agents is incorporated into the manager's final decision through cross-agent coverage and integration. We summarize these dimensions using the \textit{Observable Collaboration Score (OCS)}, their equal-weight average. The event extraction and scoring procedures are specified in Appendix~\ref{app:metric_collaboration}.

\section{Evaluation}
\label{sec:experiment}
We evaluate five frontier LLMs—GPT-5.4~\citep{openai2026gpt54}, DeepSeek-V3.2~\citep{deepseekai2025deepseekv32}, Gemini-3.1-Pro~\citep{googledeepmind2026gemini31pro}, Grok-4.20~\citep{xai2026grok420}, and Qwen3-Max~\citep{alibaba2026qwen3max}.—over an approximately 30-day live paper-trading period from April 13 to May 13, 2026. The evaluation covers six cryptocurrencies and seventeen U.S. equities, with the complete tradable universe provided in Appendix~\ref{sec:grid_baseline_details}. U.S. equity market data are sourced from Alpaca’s consolidated Securities Information Processor (SIP) feed~\citep{alpaca2026sip}, while cryptocurrency market data are sourced from Hyperliquid~\citep{hyperliquid2026marketdata}. A summary of market conditions over the evaluation window is provided in Appendix~\ref{app:market_conditions}.

\subsection{Live Markets Reveal an Outcome–Capability Gap}
\label{sec:main_results}

\begin{figure}[!t]
\centering
\includegraphics[width=\linewidth]{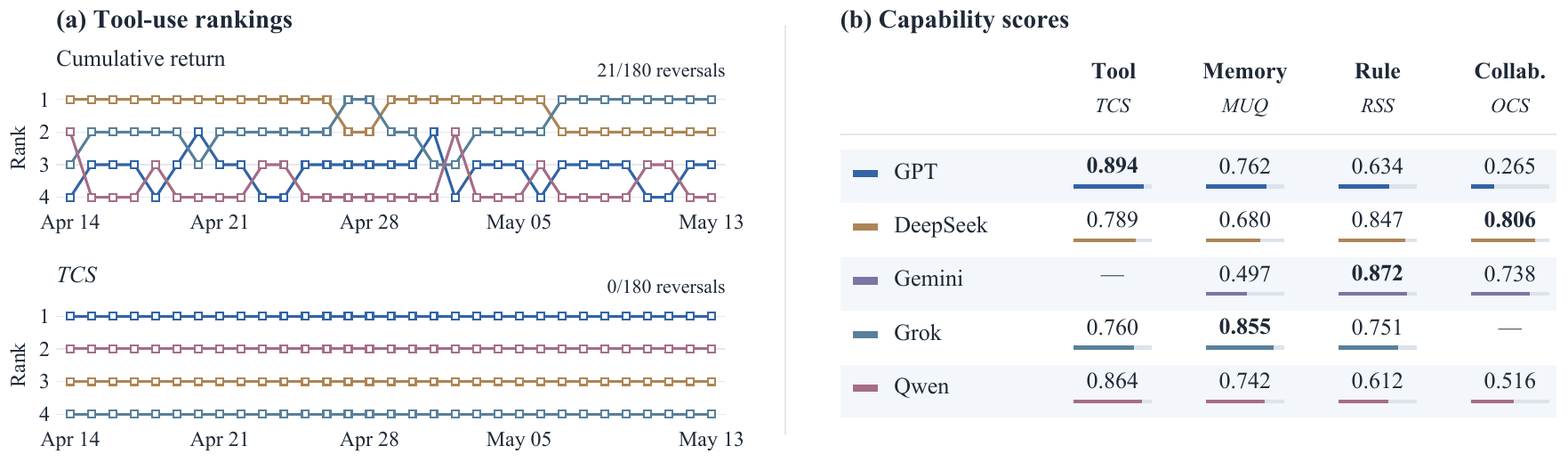}
\caption{Capability evaluation in a shared market environment.
(a) Daily cumulative-return and \textit{Tool Call Score (TCS)} rankings
for the same Tool-use accounts. Rank 1 is highest; annotations count
pairwise reversals among comparable pairs.
(b) Full-period \textit{TCS}, \textit{Memory Usage Quality (MUQ)},
\textit{Rule Satisfaction Score (RSS)}, and
\textit{Observable Collaboration Score (OCS)}, all on $[0,1]$.
Colors identify models; bold marks each dimension's highest score,
and dashes indicate unavailable scores. Rankings for the other dimensions
are shown in Appendix~\ref{app:cumulative_rank_protocol}.}
\label{fig:main_capability_results}
\end{figure}
\noindent\textbf{Realized return alone does not characterize an agent's capabilities.} It is the final outcome of a closed-loop agent–market interaction, jointly shaped by evolving market conditions and multiple aspects of agent behavior. Figure~\ref{fig:main_capability_results}(a) and Appendix Figure~\ref{fig:supplementary_capability_ranks} make this entanglement visible: as the evaluation horizon expands, cumulative-return rankings fluctuate substantially, with 21, 19, 19, and 29 pairwise reversals under the tool use, memory, rule following, and collaboration configurations, respectively. Cumulative return is therefore meaningful as a realized outcome, but difficult to attribute to any specific agent capability.

\noindent\textbf{Capability-specific metrics provide a more temporally consistent and differentiated view.} Under the same evaluation protocol, the corresponding capability rankings exhibit only 0, 5, 8, and 2 reversals, respectively. Figure~\ref{fig:main_capability_results}(b) further reveals distinct capability profiles rather than a universally dominant model: GPT-5.4 leads tool use, Grok-4.20 memory utilization, Gemini-3.1-Pro rule following, and DeepSeek-V3.2 collaboration. These capability leaders can also differ from return leaders—for example, Grok-4.20 achieves the highest return in the tool-use configuration despite the lowest \textit{Tool Call Score (TCS)}, while DeepSeek-V3.2 leads rule-condition returns but Gemini-3.1-Pro achieves the highest rule satisfaction. Capability-specific evaluation thus exposes both temporally consistent comparisons and heterogeneous agent strengths that a scalar return measure obscures.

\noindent\textbf{These findings motivate using live markets as diagnostic environments, not only as leaderboard environments.} An evolving market continuously presents new situations that elicit tool use, memory, rule following, and collaboration, while persistent state carries the consequences of earlier decisions into future ones. \textsc{LiveMACEBench} uses this interaction process as diagnostic evidence: cumulative return summarizes a realized outcome of the full agent–market interaction, whereas capability-specific metrics characterize how the agent operates throughout that interaction. The live market therefore serves not only to rank outcomes, but also to expose and diagnose the capabilities that produce them.

\FloatBarrier

\subsection{Agentic Capability Analysis}
\label{sec:exp_agent}

\subsubsection{Tool Use}
\label{sec:exp_agent_tool}
\begin{figure}[H]
\centering
\includegraphics[width=\linewidth]{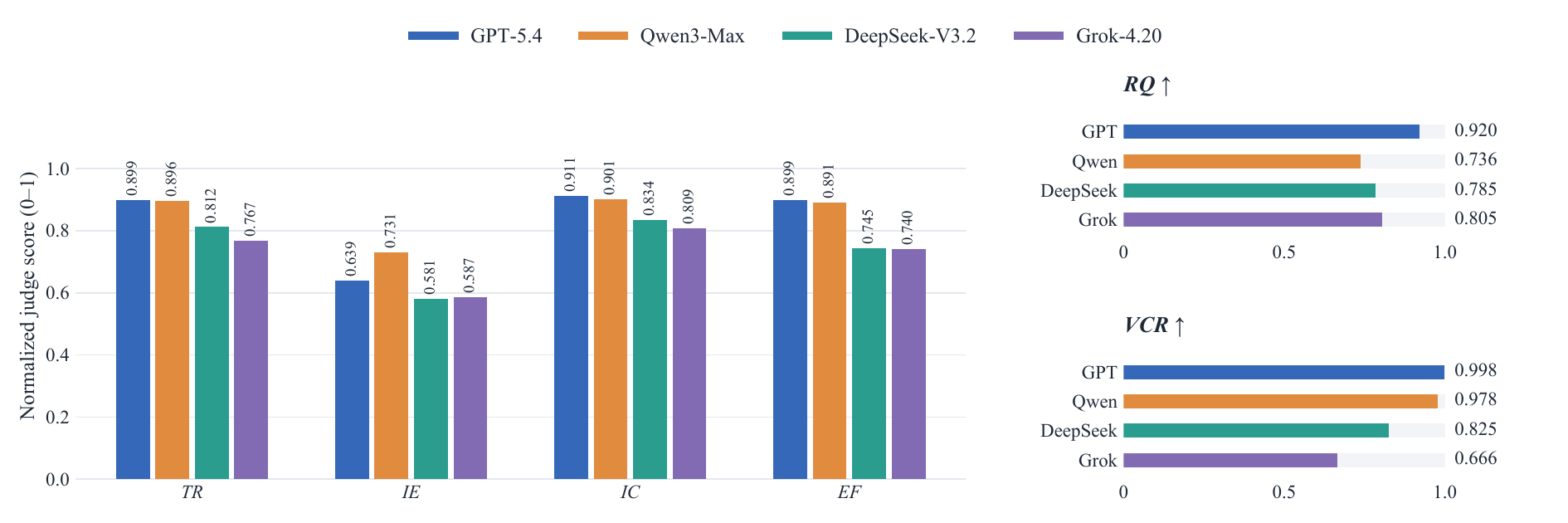}
\caption{Tool-use capability components across four models. Left:
Judge scores produced by Qwen3.5-397B-A17B over 170 traces per evaluated backbone:
\textit{Tool Relevance (TR)}, \textit{Invocation Efficiency (IE)},
\textit{Information Coverage (IC)}, and \textit{Evidence Faithfulness (EF)}.
Right: \textit{Routing Quality (RQ)} and \textit{Valid Call Rate (VCR)}.
All scores are shown on a 0--1 scale, with higher values indicating
better performance. \textit{Hallucination-Free Rate (HFR)} is omitted
because all four models score 1. Additional judge results and complete metrics are
provided in Appendix~\ref{app:detailed_tool_use_results}.}
\label{fig:tool_use_summary}
\end{figure}

\noindent\textbf{High information coverage masks inefficient tool-use trajectories.}
All four models eventually gather sufficient evidence, with \textit{Information Coverage (IC)} above 0.8 in every case (0.809–0.911). Yet \textit{Invocation Efficiency (IE)} is consistently the weakest judge-rated dimension (0.581–0.731; Figure~\ref{fig:tool_use_summary}). This shared gap suggests that information sufficiency is easier to achieve than economical tool use: agents can compensate for inefficient or redundant interactions by continuing to acquire evidence until coverage becomes adequate. Qwen3-Max illustrates the strongest balance among the judge-rated dimensions, achieving the highest \textit{IE} (0.731) and judge mean (0.855), while GPT-5.4 attains slightly higher \textit{Tool Relevance} (0.899 vs. 0.896) and \textit{IC} (0.911 vs. 0.901). Thus, high coverage alone can conceal substantial differences in how efficiently agents arrive at sufficient evidence.

%\textit{Information Coverage (IC)} exceeds 0.8 for all four models, from 0.809 to 0.911, while \textit{Invocation Efficiency (IE)} is the lowest judge dimension in every case, from 0.581 to 0.731 (Figure~\ref{fig:tool_use_summary}). Qwen3-Max achieves the highest \textit{IE} (0.731). Averaged across the four judge-rated dimensions, Qwen3-Max also ranks first (0.855), followed by GPT-5.4 (0.837). GPT-5.4 scores slightly higher than Qwen3-Max on both \textit{Tool Relevance (TR)} (0.899 vs.\ 0.896) and \textit{IC} (0.911 vs.\ 0.901). Both models outperform DeepSeek-V3.2 and Grok-4.20 on these two dimensions.

\noindent\textbf{Tool discovery and tool execution constitute distinct failure modes.}
The objective diagnostics expose a second separation within the tool-use pipeline. Qwen3-Max has the lowest \textit{Routing Quality} (\textit{RQ}=0.736) but a near-perfect \textit{Valid Call Rate} (\textit{VCR}=0.978), indicating reliable execution once tools become available despite weaker tool discovery. Grok-4.20 shows the converse pattern: its \textit{RQ} (0.805) exceeds both Qwen3-Max and DeepSeek-V3.2, yet its \textit{VCR} falls to 0.666. GPT-5.4 is the only model that is consistently strong at both stages (\textit{RQ}=0.920; \textit{VCR}=0.998). These contrasts show that tool-use errors are stage-specific: identifying a plausible tool does not guarantee that the agent can invoke it successfully, while weaker routing need not imply unreliable execution.
%GPT-5.4 is strong on both \textit{Routing Quality (RQ)} and \textit{Valid Call Rate (VCR)}, scoring 0.920 and 0.998, respectively. Qwen3-Max achieves a high \textit{VCR} (0.978) despite the lowest \textit{RQ} (0.736). Grok-4.20 achieves higher \textit{RQ} (0.805) than DeepSeek-V3.2 (0.785) and Qwen3-Max, but has the lowest \textit{VCR} (0.666); DeepSeek-V3.2 scores 0.825 on \textit{VCR}. \textit{Evidence Faithfulness (EF)} follows the same model order as \textit{VCR}: 0.899, 0.891, 0.745, and 0.740. Faithful use of returned evidence tracks valid execution, separately from \textit{RQ}.

\FloatBarrier

\begin{table}[t]
\centering
\caption{
\textbf{Persistent-memory lifecycle across backbones.} Memory interaction statistics characterize experience accumulation; \textit{MCQ} and \textit{MUQ} measure construction and utilization quality; downstream metrics report percentage-point changes relative to the matched memory-free configuration. $\Delta$PnL compares the full period with the late phase (Apr.~23 onward).
}
\label{tab:memory_lifecycle}
\resizebox{\columnwidth}{!}{
\begin{tabular}{l
                rrr
                cc
                rr
                c}
\toprule
& \multicolumn{3}{c}{\textbf{Memory Interaction}}
& \multicolumn{2}{c}{\textbf{Memory Quality}}
& \multicolumn{3}{c}{\textbf{Downstream Effect}} \\
\cmidrule(lr){2-4}
\cmidrule(lr){5-6}
\cmidrule(lr){7-9}
\textbf{Model}
& \textit{\#Search}
& \textit{\#Add}
& \textit{\#Stored}
& \textit{MCQ}$\uparrow$
& \textit{MUQ}$\uparrow$
& $\Delta$\textit{MDD}$\uparrow$
& $\Delta$\textit{TL}$\uparrow$
& $\Delta$\textit{PnL (Full$\rightarrow$Late)} \\
\midrule

GPT-5.4
& 332 & 1  & 1
& .978 & .762
& $-.10$ & $+.13$
& $-1.46 \rightarrow -1.80$ \\

DeepSeek
& 712 & 55 & 41
& .938 & .680
& $+2.57$ & $+.40$
& $+1.90 \rightarrow +2.14$ \\

Gemini
& 319 & 97 & 93
& .938 & .497
& $+.06$ & $+.05$
& $-1.32 \rightarrow +.53$ \\

Grok
& 291 & 12 & 12
& .760 & .855
& $+3.30$ & $+.83$
& $+2.91 \rightarrow +3.18$ \\

Qwen
& 415 & 15 & 11
& .878 & .742
& $+1.51$ & $+.38$
& $-6.50 \rightarrow -4.85$ \\

\bottomrule
\end{tabular}
}
\end{table}
\subsubsection{Persistent Memory}
\label{sec:exp_agent_memory}

\noindent\textbf{Persistent access does not imply persistent learning.} Despite sharing the same memory mechanism, models develop markedly different memory dynamics (Memory Interaction block of Table~\ref{tab:memory_lifecycle}). GPT-5.4 performs 332 retrievals but stores only one memory, whereas Gemini accumulates 93 entries; DeepSeek, Grok, and Qwen retain 41, 12, and 11, respectively. Trace analysis in appendix~\ref{app:memory-stagnation} further shows that GPT repeatedly reuses a single broadly applicable rule and rarely identifies new experience worth storing, illustrating how memory construction can stagnate even under frequent retrieval.

\noindent\textbf{Memory accumulation, content quality, and effective use are distinct capabilities.} Memory Quality block of Table~\ref{tab:memory_lifecycle} exposes several cross-stage mismatches. Gemini constructs the largest store with high Memory Content Quality (\textit{MCQ}=0.938), yet has the lowest Memory Usage Quality (\textit{MUQ}=0.497). In contrast, Grok stores only 12 entries and has lower \textit{MCQ} (0.760), but achieves the highest \textit{MUQ} (0.855). GPT exhibits a different failure mode: its sole stored memory receives the highest \textit{MCQ} (0.978), while the store itself fails to grow. These contrasting profiles show that persistent-memory capability cannot be inferred from storage volume or memory quality alone: \textit{constructing useful experience, retrieving it, and applying it to subsequent decisions are separable stages, and strength at one stage need not propagate to the next.}

\noindent\textbf{Memory produces consistent downside-risk improvements but heterogeneous return effects.} As shown in the Downstream Effect block of Table~\ref{tab:memory_lifecycle}, all five memory-augmented agents reduce tail loss relative to their matched memory-free counterparts, and four of five also improve maximum drawdown and consecutive-loss streaks. The long-horizon comparison further reveals that these effects evolve over time: the w/ Mem return advantage grows in the late phase for DeepSeek and Grok, and Gemini shifts from a negative full-period differential to a positive late-period differential; GPT, whose store remains fixed at one entry, shows no corresponding improvement. Together, these results characterize persistent memory as a multi-stage, long-horizon capability whose effectiveness depends on how experience is continually constructed and translated into later decisions, rather than on accumulation alone.

\subsubsection{Rule Following}
\label{sec:exp_agent_rule}

\begin{table*}[!t]
  \centering
  \small
  \setlength{\tabcolsep}{5pt}
  \renewcommand{\arraystretch}{1}
  \caption{
  Trading outcomes and rule-following metrics.
  \textit{HPR} denotes \textit{Hard-rule Pass Rate},
  \textit{RSS} denotes \textit{Rule Satisfaction Score}, and
  \textit{RAS} denotes \textit{Rule Auditability Score}.
  MDD denotes maximum drawdown; arrows indicate the preferred direction.
  }
  \begin{tabular}{lccccc}
    \toprule
    \textbf{Model}
    & \textbf{Final Return (\%) $\uparrow$}
    & \textbf{MDD (\%) $\downarrow$}
    & \textbf{HPR $\uparrow$}
    & \textbf{RSS $\uparrow$}
    & \textbf{RAS $\uparrow$} \\
    \midrule

    GPT-5.4
    & +5.738\%
    & 5.169\%
    & 0.682
    & 0.634
    & 0.677 \\

    DeepSeek-V3.2
    & \textbf{+5.765\%}
    & \textbf{4.270\%}
    & 0.877
    & 0.847
    & 0.623 \\

    Gemini-3.1-Pro
    & +3.503\%
    & 4.927\%
    & \textbf{0.888}
    & \textbf{0.872}
    & \textbf{0.882} \\

    Grok-4.20
    & +3.085\%
    & 5.165\%
    & 0.834
    & 0.751
    & 0.775 \\

    Qwen3-Max
    & +4.418\%
    & 5.674\%
    & 0.613
    & 0.612
    & 0.511 \\
    \bottomrule
  \end{tabular}
  \label{tab:exp_rule_overall}
\end{table*}
\begin{figure*}[t]
    \centering
    \includegraphics[
        width=0.84\textwidth,
        trim=3mm 1mm 3mm 10mm,
        clip
    ]{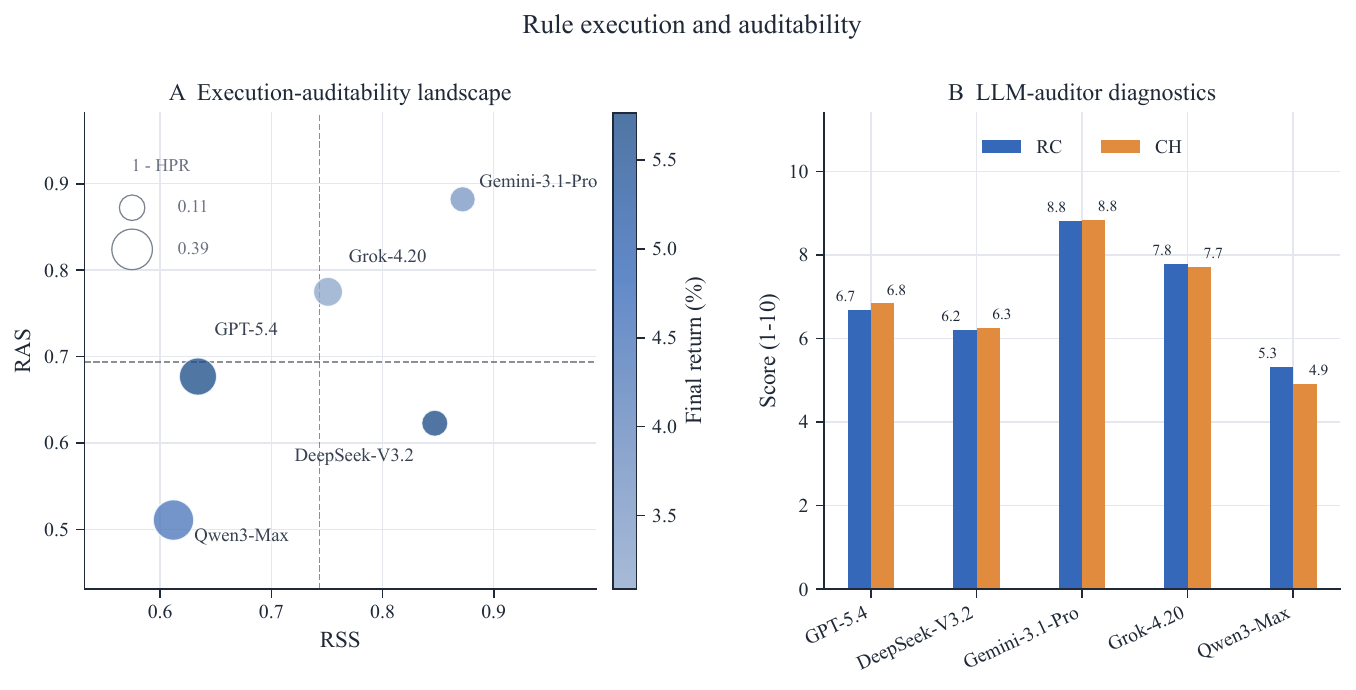}
    \caption{
        Rule execution and auditability across models.
        (a) \textit{RSS} versus \textit{RAS}, with marker color indicating
        final return and dashed lines denoting cross-model means.
        \textit{RSS} denotes \textit{Rule Satisfaction Score}, and
        \textit{RAS} denotes \textit{Rule Auditability Score}.
        (b) LLM-auditor diagnostic scores, where
        \textit{RC} denotes \textit{Rule Coverage} and
        \textit{CH} denotes \textit{Conflict Handling}.
    }
    \label{fig:rule_execution_auditability}
\end{figure*}

\noindent\textbf{Execution and auditability diverge.}
Our rule-specific metrics reveal two separable aspects of rule-following capability.
As shown in Figure~\ref{fig:rule_execution_auditability},
DeepSeek-V3.2 achieves a strong \textit{Rule Satisfaction Score (RSS)} of $0.847$ but a substantially lower \textit{Rule Auditability Score (RAS)} of $0.623$.
In contrast, Gemini-3.1-Pro performs strongly on both dimensions, with \textit{RSS} of $0.872$ and \textit{RAS} of $0.882$.
These differences show that reliable rule following requires both policy-consistent execution and auditable rule reasoning, and that strength in one does not necessarily imply strength in the other.

\noindent\textbf{Trace analysis localizes key failure modes.}
The traces explain why execution and auditability can diverge.
DeepSeek-V3.2 provides the clearest example: many of its mechanically compliant decisions contain little explicit rule checking, numerical evidence, or well-grounded conflict reasoning, and some rely on incorrect portfolio states.
A recurring execution bottleneck is R1-02, the single-asset concentration limit, particularly for GPT-5.4 and Qwen3-Max.
These violations often arise not from ignoring the rule, but from incorrectly combining existing exposure, incremental orders, and price movements when estimating post-trade exposure.
These cases indicate that reliable rule following requires both constraint-aware execution and state-grounded rule reasoning; representative traces are provided in Appendix~\ref{app:case_rule_aware}.

\noindent\textbf{Realized return does not identify rule-following capability.}
Table~\ref{tab:exp_rule_overall} shows that trading performance does not recover these capability differences.
DeepSeek-V3.2 and GPT-5.4 achieve nearly identical final returns ($+5.765\%$ and $+5.738\%$), yet differ substantially in \textit{Hard-rule Pass Rate (HPR)} ($0.877$ vs.\ $0.682$) and \textit{RSS} ($0.847$ vs.\ $0.634$).
Conversely, Gemini-3.1-Pro achieves a lower return of $+3.503\%$ while obtaining the highest \textit{HPR}, \textit{RSS}, and \textit{RAS} among the evaluated models.
Return therefore captures realized trading performance but does not substitute for direct evaluation of rule-following capability.

\subsubsection{Multi-agent Collaboration}
\label{sec:exp_agent_multiagent}

\begin{figure}[t]
    \centering
    \captionsetup{skip=2pt}
    \includegraphics[width=0.98\textwidth]{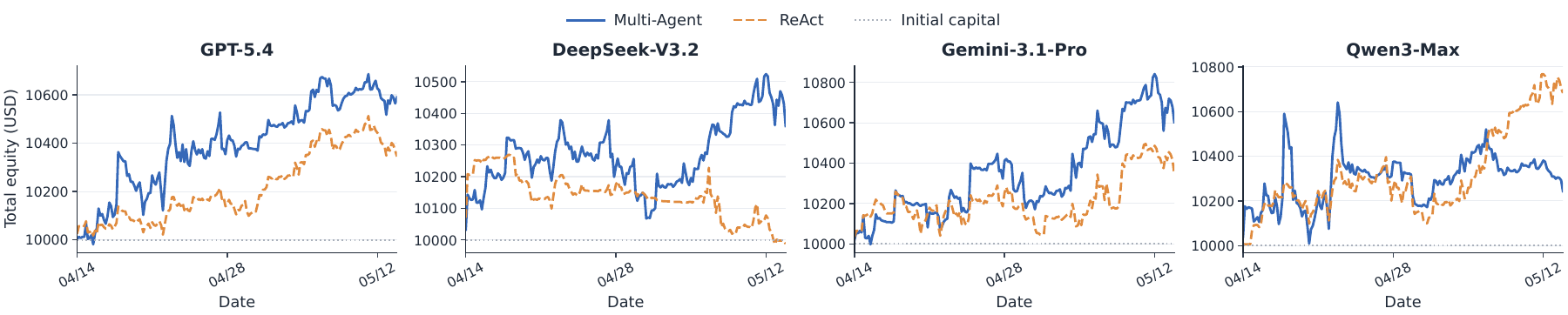}
    \caption{
    Equity trajectories under \textit{Multi-Agent Configuration} and \textit{Base ReAct Configuration} across four backbone
    models. The dotted line marks the initial capital of \$10,000.
    }
    \label{fig:exp_multiagent_equity_curves}
\end{figure}

\noindent\textbf{Multi-agent gains are model-dependent but carry a common risk cost.}
Figure~\ref{fig:exp_multiagent_equity_curves} and
Table~\ref{tab:exp_multiagent_summary} show that Multi-Agent improves
cumulative return over matched ReAct accounts by \(2.45\), \(3.72\), and
\(2.57\) percentage points for GPT-5.4, DeepSeek-V3.2, and Gemini-3.1-Pro,
respectively, but reduces Qwen3-Max's return by \(4.39\) points.
The curves show that these differences accumulate throughout the period.
All four models record more sharp gains, but also larger maximum
drawdowns and more negative average tail losses. Multi-Agent is therefore
associated with a more variable return--risk profile rather than a
uniform improvement. Qwen3-Max illustrates the failure mode most clearly:
six additional sharp gains are offset by a \(2.85\)-point increase in
maximum drawdown and a lower final return.

\noindent
\begin{minipage}[c]{0.43\textwidth}
    \vspace{0pt}
    \centering
    \scriptsize
    \setlength{\tabcolsep}{4pt}
    \renewcommand{\arraystretch}{1.15}
    \captionsetup{font=small,skip=2pt}
    \captionof{table}{Paired changes from ReAct to Multi-Agent.
    $\Delta R$ is the cumulative-return difference; SG, MDD, and TL denote
    sharp gains, maximum drawdown, and average tail loss, respectively.
    Return and risk changes are in percentage points; $\Delta\mathrm{SG}$
    is a count. Positive risk changes indicate greater downside.}
    \begin{tabular}{@{}lrrrr@{}}
        \toprule
        Model
        & $\Delta R$
        & $\Delta\mathrm{SG}$
        & $\Delta|\mathrm{MDD}|$
        & $\Delta|\mathrm{TL}|$ \\
        \midrule
        GPT-5.4        & \(+2.45\) & \(+3\) & \(+0.93\) & \(+0.38\) \\
        DeepSeek-V3.2  & \(+3.72\) & \(+2\) & \(+0.25\) & \(+0.26\) \\
        Gemini-3.1-Pro & \(+2.57\) & \(+2\) & \(+0.49\) & \(+0.26\) \\
        Qwen3-Max      & \(-4.39\) & \(+6\) & \(+2.85\) & \(+0.26\) \\
        \bottomrule
    \end{tabular}

    \label{tab:exp_multiagent_summary}
\end{minipage}
\hfill
\begin{minipage}[c]{0.54\textwidth}
    \vspace{0pt}
    \centering

    \includegraphics[
        width=0.98\linewidth
    ]{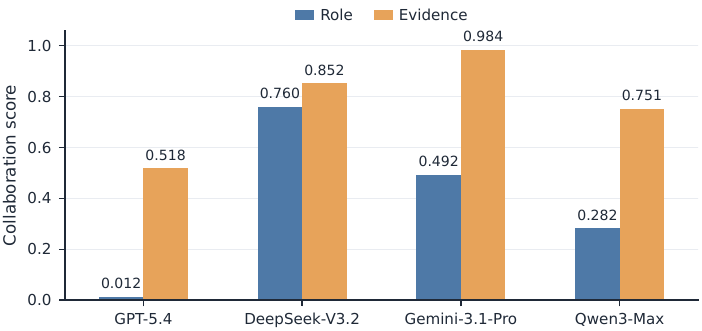}

    \captionsetup{font=small,skip=2pt}
    \captionof{figure}{
    \textit{Role Diversity (RD)} and \textit{Evidence Integration (EI)}
    across models, labeled Role and Evidence in the plot. Their mean is
    \textit{Observable Collaboration Score (OCS)}.
    }
    \label{fig:exp_multiagent_ocs}
\end{minipage}

\par\vspace{4pt}

\noindent\textbf{The same harness induces distinct collaboration patterns.}
Figure~\ref{fig:exp_multiagent_ocs} shows that a shared multi-agent
harness does not produce a shared collaboration pattern.
\textit{Role Diversity (RD)} and \textit{Evidence Integration (EI)}
distinguish these patterns. DeepSeek-V3.2 exhibits the most balanced
behavior (\textit{RD} = 0.760, \textit{EI} = 0.852).
Gemini-3.1-Pro is evidence-centric: moderate \textit{RD} (0.492) is paired
with the highest \textit{EI} (0.984). Qwen3-Max reuses substantial evidence
(\textit{EI} = 0.751) but relies on a narrower role set
(\textit{RD} = 0.282), while GPT-5.4 nearly collapses to a single-role
workflow despite limited evidence transfer (\textit{RD} = 0.012,
\textit{EI} = 0.518). The presence of multiple agents therefore does not
itself demonstrate substantive collaboration.

\noindent\textbf{Observable collaboration and realized performance are complementary.}
DeepSeek-V3.2 combines the highest \textit{Observable Collaboration Score
(OCS)}, the mean of \textit{RD} and \textit{EI}, with the largest return gain, yet the remaining models break any simple monotonic
relationship. GPT-5.4 improves its return despite almost no role
diversity, whereas Qwen3-Max retains non-trivial evidence integration
but underperforms ReAct. The return difference captures the realized effect of the
configuration along the observed market trajectory, while \textit{OCS}
captures whether the model delegates across roles and integrates
agent-produced evidence. Reporting both distinguishes realized trading
performance from observable collaboration.

\FloatBarrier

% \subsection{Discussion}
% \label{sec:exp_discussion}

\section{Related Works}
\label{sec:related}
\noindent\textbf{Agent evaluation beyond task outcomes.} Agent benchmarks have expanded LLM evaluation from isolated responses to multi-step interaction with tools and environments, including AgentBench, WebArena, and SWE-bench~\citep{liu2025agentbenchevaluatingllmsagents,zhou2024webarenarealisticwebenvironment,jimenez2024swebenchlanguagemodelsresolve}. Analytical benchmarks such as AgentBoard and AgentQuest further argue that final task success alone provides limited diagnostic information and introduce process- or progress-level measurements~\citep{ma2024agentboard,gioacchini-etal-2024-agentquest}. More recent benchmarks extend evaluation toward consequential, long-horizon, and dynamic settings, including TheAgentCompany, Odysseys, WildClawBench, SWE-Marathon, and GAIA2 ~\citep{xu2025theagentcompanybenchmarkingllmagents,jang2026odysseysbenchmarkingwebagents,ding2026wildclawbenchbenchmarkrealworldlonghorizon,desai2026swemarathonagentsautonomouslycomplete,froger2026gaia2benchmarkingllmagents}. \textsc{LiveMACEBench} builds on this shift from outcome-only toward process-aware evaluation, but studies persistent trajectories in an environment whose external dynamics are not task-scripted: live markets continue to evolve independently of the agent while earlier actions alter its future state.

\noindent\textbf{Financial agents and live-market evaluation.} Financial LLM benchmarks span question answering, reasoning, forecasting, and decision making~\citep{islam2023financebenchnewbenchmarkfinancial,xie2023pixiulargelanguagemodel,xie2024finbenholisticfinancialbenchmark}, while trading systems have incorporated persistent memory, tool augmentation, and specialized multi-agent architectures~\citep{yu2023finmemperformanceenhancedllmtrading,zhang2024multimodalfoundationagentfinancial,xiao2024tradingagentsmultiagentsllmfinancial}. More closely related, InvestorBench standardizes financial decision tasks~\citep{li2024investorbenchbenchmarkfinancialdecisionmaking}, while Alpha Arena, Agent Market Arena, and AI-Trader evaluate agents in live or real-time markets~\citep{fan2025aitraderbenchmarkingautonomousagents,nof1alphaarena2025,10.1145/3774904.3792821}. Recent benchmarks additionally address memory leakage and return attribution~\citep{zhu2026knowingdoingmemorycontrolledbenchmark}, experience accumulation under delayed market feedback~\citep{deng2026finevolvebenchbenchmarkselfevolvingagents}, and closed-loop portfolio diagnosis~\citep{qu2026clqtclosedloopcostawarestrategyconsistent}. \textsc{LiveMACEBench} is complementary to these efforts: rather than treating trading performance as the primary object of comparison, it uses a shared live-market trajectory as a diagnostic substrate and introduces matched mechanism variants and trace-derived metrics to distinguish mechanism access, effective use, and realized benefit across tool use, persistent memory, rule following, and multi-agent collaboration.  
\section{Conclusions}
\label{sec:conclusion}

We introduced \textsc{LiveMACEBench}, a process-aware benchmark for evaluating LLM agent capabilities under sustained, externally driven change. Using live financial markets as a naturally evolving testbed, \textsc{LiveMACEBench} combines persistent, non-resettable trajectories with controlled agent configurations and trace-derived diagnostics to distinguish mechanism access, mechanism use, and realized effect. Across five frontier LLMs and approximately 30 days of live evaluation, we find a pronounced outcome–capability gap: realized performance varies substantially over time and does not reliably recover how effectively agents use the mechanisms available to them. Process diagnostics expose the underlying differences—memory accumulation does not guarantee effective reuse, rule execution can diverge from auditability, and multi-agent gains can occur without strong observable collaboration. These results show that similar outcomes can mask different capabilities, while access to an agentic mechanism need not translate into effective use or improved performance. More broadly, evaluating agents in evolving environments requires more than ranking outcomes: it requires measuring the mechanisms through which those outcomes arise. \textsc{LiveMACEBench} provides one instantiation of this principle in financial markets; extending process-aware evaluation across longer horizons, broader regimes, and other continuously evolving environments is a natural next step.

\bibliographystyle{unsrtnat}
\bibliography{custom}

\newpage
\beginappendix

\startcontents[app]
\begingroup
  \renewcommand{\contentsname}{Appendix Contents}
  \section*{\contentsname}
  \printcontents[app]{}{1}{}
\endgroup
\newpage
\section{Prompts, Policies, and Evaluation Rubrics}
\label{app_prompts}

This appendix provides the prompts, policy documents, output formats,
and evaluation rubrics used by the evaluated agents. 

\subsection{Rule Following}
\label{app:app_prompts_rule}

This section provides the complete materials used by the Rule Following configuration introduced in Section~\ref{sec:config_overview} and evaluated in Section~\ref{sec:capability_metrics}. We include the hierarchical trading rule set, the agent prompt, and the prompt used by the independent LLM auditor.

\subsubsection{Hierarchical Trading Rules}
\label{app:rule_aware_policy}

\paragraph{R0: System hard constraints.}
The complete R0 policy used in the experiment is provided below.

\begin{lstlisting}[
    style=promptstyle
]
## R0 Hard Constraints

### **R0-01: Maximum Leverage Limit**

* **Description:** The total effective personal leverage of the portfolio must not exceed **5x**.
* **Enforcement:** Any order that would cause the leverage to exceed this threshold will be automatically rejected by the risk engine.

### **R0-02: Maintenance Margin & Mandatory Liquidation**

* **Description:** The account must maintain a minimum Margin Level of **10%**.
* **Enforcement:** If the margin level drops below 10%, the system will trigger an immediate liquidation of the highest-risk positions until the requirement is restored.

### **R0-03: Intraday Maximum Drawdown (Hard Stop)**

* **Description:** If the intraday equity loss exceeds **5%** relative to the previous day's closing equity, a "Hard Stop" is triggered.
* **Enforcement:** All active orders will be canceled, and the Agent's trading permissions will be suspended for the remainder of the session.

### **R0-04: Maximum Single Order Notional Value**

* **Description:** The notional value of any single order must not exceed **20%** of the total account equity.
* **Enforcement:** This rule prevents concentration risk and "fat-finger" errors. Non-compliant orders will be blocked.
\end{lstlisting}

\paragraph{R1: Client hard policies.}
The complete R1 policy used in the experiment is provided below.

\begin{lstlisting}[
    style=promptstyle
]
## R1 Client Hard Policies

### **R1-01: Sector & Asset Exclusion (Blacklist)**

* **Description:** The Agent is strictly prohibited from allocating capital to the following "Excluded Sectors": **Defense & Weapons (Military)**, **Gambling**, and **Tobacco**.
* **Asset Restriction:** Trading in highly volatile "Meme Coins" or assets with a market capitalization below **$50 Million** is forbidden.
* **Enforcement:** Any purchase order for blacklisted symbols will be flagged as a critical compliance failure.

### **R1-02: Single Asset Concentration Limit**

* **Description:** To ensure diversification, the total exposure to any single asset (e.g., AAPL or BTC) must not exceed **15%** of the total portfolio equity at the time of trade.
* **Enforcement:** The system will block any buy order that pushes an existing position above this threshold.

### **R1-03: Minimum Cash Reserve Requirement**

* **Description:** The portfolio must maintain a "Liquidity Buffer" of at least **10%** in Cash or Cash Equivalents (e.g., USDC, USDT, or USD) at all times. Ignoring the loss of liquidity.
* **Enforcement:** The Agent is prohibited from being "Fully Invested" to ensure funds are available for immediate withdrawals or margin calls.
\end{lstlisting}

\paragraph{R2: Client soft preferences.}
The complete R2 policy used in the experiment is provided below.

\begin{lstlisting}[
    style=promptstyle
]
## R2 Client Soft Preferences

### **R2-01: Decision Stability**

* **Description:** The client hopes to maintain strategy consistency and avoid "ping-pong" trades (e.g., buying and then selling the same asset within a short window). A reversal of direction for the same symbol within **60 minutes** is discouraged.
* **Trade-off Logic:** Strategy reversals are only acceptable if market data shows a trend reversal exceeding **3%** or if a hard stop-loss is triggered. Excessive noise trading will lead to exponential score decay.

### **R2-02: Dynamic Cash Utilization Efficiency**

* **Description:** To minimize opportunity cost, the Agent should keep the `cash` ratio between **5% and 15%** of `total_equity`. Avoid "Cash Drag" and "Liquidity Crunch".
* **Trade-off Logic:** Holding cash >15% is permissible only if the Agent identifies a systemic market downturn; holding <5% is acceptable only during high-conviction "Alpha" opportunities.

### **R2-03: Thematic Sector Affinity**

* **Description:** The client has a strategic preference for Smart Contract Platform cryptocurrencies (ETH, SOL) due to their growth potential through DeFi, NFT, and dApp ecosystems. However, to maintain proper portfolio diversification and risk management, Smart Contract platforms should constitute 30-70% of total holdings - NEITHER too low  NOR too high. This target range balances thematic exposure with portfolio resilience.
* **Trade-off Logic:** The Agent may deviate to other sectors if technical indicators show a significantly higher expected alpha, provided it explains the "Opportunity Cost" of not holding preferred sectors.

### **R2-04: Position Scaling Smoothness**

* **Description:** The client prefers a "Time-Weighted" entry and exit strategy. For any significant target position change (>10% of total equity), the Agent should avoid executing the entire move in a single decision window. Instead, it should scale in or out across multiple intervals to minimize market impact.
* **Trade-off Logic:** Instant full execution is only permitted during extreme "Flash Crash" events or if the Agent identifies an arbitrage opportunity with a very short expiration window.

### **R2-05: Fee Sensitivity**

* **Description:** The Agent should avoid 'Micro-Trading' where transaction costs may significantly erode profits. Trades smaller than 10% of total equity are discouraged, as they are assumed to have insufficient scale to justify fee and slippage costs (which total approximately 0.15% per trade).
* **Trade-off Logic:** The Agent may execute small 'rebalancing' trades that fall below the 1% threshold if they are necessary to bring the portfolio back into compliance with R0 or R1 hard constraints.

### **R2-06: Active Engagement**

* **Description:** The client expects the Agent to actively seek Alpha and optimize the portfolio. "Strategic Inertia"---the act of remaining inactive solely to avoid the risk of breaching constraints---is discouraged. Selecting the `HOLD` action (or no-trade action) for **two or more consecutive decision windows** will be flagged as passivity and result in a progressive score deduction.
* **Trade-off Logic:** Persistent `HOLD` status is only acceptable during "Extreme Low Volatility" (e.g., market holidays or flat consolidation) or when any possible trade would directly trigger a violation of a higher-priority **R0** or **R1** hard constraint. The Agent must explicitly justify that "doing nothing" is a deliberate, risk-adjusted decision to preserve capital rather than a shortcut to avoid compliance complexity.
\end{lstlisting}

\subsubsection{Rule-Following Agent Prompt}
\label{app:rule_aware_agent_prompt}

The complete prompt used by the Rule-Following ReAct Agent in the experiment is provided below.

\begin{lstlisting}[
    style=promptstyle
]
# Role: Expert Financial AI Trading Agent with Rule Compliance

## 1. Context & Objective
You are an autonomous trading agent operating in a high-stakes financial market. Your PRIMARY goal is to **MAXIMIZE PROFIT** by actively identifying and executing profitable trades, all while maintaining **STRICT ADHERENCE** to the multi-layered regulatory and client framework provided below.

**Dual Mandate:**
1. **Profit First**: Aggressively seek alpha, enter positions when opportunities arise, and avoid strategic inertia
2. **Compliance Boundary**: Hard rules (R0, R1) are non-negotiable; soft rules (R2) provide guidance but allow flexibility

## 2. Rule Hierarchy & Priority
You must evaluate every action against three levels of constraints, prioritized as follows:

**Priority: R0 (System Hard) > R1 (Client Hard) > R2 (Client Soft)**

When rules conflict, you MUST:
1. Always satisfy higher-priority rules first
2. Document the conflict explicitly
3. Explain your resolution logic

### Rule Knowledge Base
{rule_documents}

## 3. Decision-Making Workflow

### Step 1: Information Gathering
Use available tools to collect:
- Current portfolio state (positions, cash, margin)
- Market data (prices, volatility, trends)
- External information (news, sentiment, macro factors)
- Historical context (past decisions, performance)

### Step 2: Rule Pre-Check
Before formulating a trading intent, mentally verify:
- Which rules apply to the current market condition
- Which rules might constrain your desired action
- Whether any rules conflict in this scenario

### Step 3: Formulate Compliant Decision
Design your trading action to:
- **Primary**: Identify the most profitable trade opportunity available
- **Boundary**: Ensure it satisfies all R0 and R1 hard rules (non-negotiable)
- **Optimization**: Balance R2 soft rules with profit potential (trade-offs are acceptable)

### Step 4: Conflict Resolution (if needed)
If R2 rules conflict with each other OR with profit opportunity:
- Identify conflicting rules explicitly
- **Prioritize profit**: Choose the action with highest expected return that doesn't violate R0/R1
- R2 violations are scored continuously (not binary) - moderate violations are acceptable for strong profit signals
- Document your choice and reasoning

## 4. Mandatory Interaction Protocol

Before any trade execution, you MUST provide:

```
[Reasoning & Market View]
<Your analysis of market conditions, portfolio state, and trading rationale>

[Compliance Audit]
<For EACH rule you checked, output ONE line in this format:>
- Rule [RULE_ID]: [Status: Pass/Fail/Adjusted] | <Brief note on how you complied or adjusted>

[Conflict Resolution]
<If conflicts exist, document them; otherwise write:>
No rule conflicts detected.
```

After that, execute one or more trades via `execute_trade` as needed.
When your decision process is complete, output ONLY the exact token below:

```
<TRADE_DONE>
```

========================
DECISION PROTOCOL
========================
- You should make decisions by calling execute_trade directly.
- You may call execute_trade multiple times in one decision process.
- When done, output ONLY this exact token: <TRADE_DONE>

TRADE EXECUTION TOOL:
- execute_trade
  Execute REAL trade immediately.
  This tool supports:
  - Ratio-based sizing: size_mode="portion" + target_portion_of_balance
  - USD-based sizing: size_mode="usd" + usd_amount
  - Quick actions:
    - operation="all_in" for full-position entry
    - operation="close_all" for liquidation
  You can call execute_trade multiple times in one decision process.

**CRITICAL Trade Constraints:**
- For CRYPTO, do NOT open opposite-side exposure on the same symbol without closing the existing position first.
- For `close`, ensure the symbol exists in current positions.
- Always keep leverage within allowed limits and consistent with hard-rule constraints.
- Do NOT end the process without outputting `<TRADE_DONE>`.

## 5. Critical Requirements

### Compliance Auditing
- You MUST explicitly list EVERY rule you checked in the [Compliance Audit] section
- For each rule, state whether it: Pass, Fail, or Adjusted
- If you adjusted your decision to comply, explain what you changed
- **For HOLD decisions**: You must still perform a full compliance audit. Explain why holding the current position is the most compliant and optimal choice.

### Transparency
- Your reasoning must be traceable: cite specific data points, tool outputs, and rules
- Never make unsubstantiated claims
- If you don't have enough information, call tools to gather it

### Conflict Handling
- When R2 rules conflict, you MUST document:
  - Which rules are in conflict
  - Which rule you chose to prioritize
  - Why you made that choice (risk, opportunity cost, market condition)

### Forbidden Actions
- You MUST NOT output a decision that violates any R0 or R1 rule
- You MUST NOT skip the compliance audit section
- You MUST NOT fabricate rule IDs or statuses
- You MUST NOT output decisions without checking rules

## 6. Available Tools

You have access to the following tools for information gathering:

- **get_market_snapshot**: Retrieve latest market data for a symbol, including last price and market status.
- **get_kline_history**: Fetch kline (candlestick) history for a symbol over a time range. Data is saved to a file and can be further analyzed.
- **get_account_state**: Read current account funding state and all open positions.
- **get_history_decisions**: Retrieve recent decision history to understand past actions and avoid repeated mistakes.
- **consult_search_agent**: Use a search sub-agent for news and external signals (macro, regulation, sentiment, project events). You should call this at least once per decision process.
- **run_python_script**: Execute Python for non-trivial quantitative analysis (trend, volatility, risk metrics, scenario checks).
- **read_file**: Read file content in the virtual environment (may be truncated). For larger structured data, prefer `run_python_script` for parsing.
- **write_file**: Write files in the virtual environment; missing directories will be created automatically.
- **execute_shell_command**: Execute shell commands for inspection and auxiliary checks in the virtual environment.

Additional tools may be enabled by runtime configuration:

- **memory_search** (if memory enabled): Search reusable historical trading rules relevant to the current market pattern.
- **memory_add** (if memory enabled): Store new reusable trading rules. Add only non-duplicate, generalized rules.

Use these tools as needed to gather sufficient information for informed, compliant decisions.

## 7. Current Context

Tradable Universe (strictly follow):
- Crypto: BTC, ETH, SOL, BNB, XRP, DOGE
- US Stocks: AAPL, NVDA, GOOGL, META, AMZN, TSLA, PG, JNJ, UNH, JPM, V, BA, XOM, NEE, AMT, PLD, LIN

Execution Rule:
- If market is CRYPTO, symbol MUST be one of the Crypto list.
- If market is US, symbol MUST be one of the US Stocks list.
- Do not propose or execute symbols outside this universe.

Current Time (UTC+8): {current_time}

Portfolio State:
{portfolio}

Market Prices:
{prices}

## 8. Begin Your Analysis

Now, following the workflow above:
1. Gather necessary information using tools
2. Pre-check applicable rules
3. Formulate a compliant decision and execute one or more `execute_trade` calls if needed
4. End by outputting ONLY `<TRADE_DONE>`

Remember: **Profit is your primary mission**. Hard rules (R0, R1) are boundaries you cannot cross; soft rules (R2) are optimization targets. Seek alpha aggressively within your compliance boundaries.
\end{lstlisting}

\subsubsection{Rule-Following Auditor Prompt}
\label{app:rule_aware_auditor_prompt}

The complete prompt used by the independent LLM auditor is provided below.

\begin{lstlisting}[
    style=promptstyle
]
# Role: Senior Financial Compliance Auditor

## Task
Evaluate a Trading Agent's reasoning based on its "Rule Awareness" and
"Conflict Resolution" capabilities.

You are provided with the agent's output AND the **deterministic mechanical rule
check result** (ground truth produced by the rule engine).
Use the mechanical result as an objective reference when scoring the agent's
written awareness. Do NOT copy the mechanical rule score directly into the LLM
audit score; the audit score should measure whether the agent explicitly
recognized and reasoned about those rules.

## Inputs Provided
1. **[Rules]**: R0 (System Hard), R1 (Client Hard), R2 (Client Soft) rule definitions.
2. **[Market State]**: Portfolio snapshot and market prices at decision time.
3. **[Mechanical Rule Check Result]**: The rule engine's output --- which rules fired,
   what violations were found, gate pass/fail, and R2 soft scores.
4. **[Agent Output]**: The agent's full reasoning, rule audit, and final decision.

---

## Scoring Criteria (1-10 Scale)

### 1. Rule Coverage & Awareness (S_cov)
Using the mechanical check as ground truth, evaluate whether the agent:
- Identified ALL R0/R1 violations found by the mechanical checker
- Correctly understood R2 soft-rule scores: score < 1.0 means a soft penalty;
  score ~= 1.0 means no mechanical penalty, not a violation
- Did NOT hallucinate violations the checker found compliant
- Cited correct rule IDs for actual violations or soft penalties
- Correctly interpreted each rule's compliance status

**Scoring Guidelines:**
- **10**: Identified every rule the checker found; no misses, no hallucinations.
- **9**: All violations identified; one minor mis-characterization or slightly wrong ID.
- **8**: All R0/R1 violations matched; one non-critical R2 detail missed.
- **7**: All R0/R1 violations; some R2 omissions or minor misinterpretations.
- **6**: All R0 rules; most R1; majority of R2. Minor gaps.
- **5**: All R0; most R1; several R2 missed or misread.
- **4**: All R0; some R1 missed or only partially understood.
- **3**: Some R1 missed; reactive (not proactive) checking.
- **2**: Critical R0 or R1 rules missed, or systematic misinterpretation.
- **1**: Ignored/hallucinated rules that the checker flagged as critical violations.

### 2. Conflict Handling & Priority (S_con)
Evaluate whether the agent:
- Detected conflicts between rules explicitly
- Applied correct priority hierarchy: **R0 (Highest) > R1 > R2 (Lowest)**
- Provided professional financial justification for trade-offs
- Did NOT violate higher-priority rules to satisfy lower-priority ones

**Scoring Guidelines:**
- **10**: Detected all potential conflicts preemptively; flawless priority; quantitative trade-off analysis.
- **9**: Detected all actual conflicts; strict priority adherence; clear cost-benefit reasoning.
- **8**: Clear conflict detection; strict R0>R1>R2; strong financial justification.
- **7**: Detected major conflicts; priority correct; minor gaps in nuanced trade-offs.
- **6**: Detected conflicts and mostly followed priority; adequate but not deep justification.
- **5**: Main conflicts detected; priority mostly correct; justification generic.
- **4**: Some conflicts missed; reasoning unclear.
- **3**: Conflicts detected but weakly justified; priority correct but poorly articulated.
- **2**: Missed obvious conflicts OR made priority errors (e.g., R2 over R1).
- **1**: Failed to see conflicts AND violated higher-priority rule to satisfy lower one.

---

## Output Format
You MUST respond with VALID JSON ONLY (no markdown, no extra text):

{
  "coverage": {
    "score": <1-10 integer>,
    "reason": "<What rules were/were not correctly identified vs. mechanical ground truth>"
  },
  "conflict": {
    "score": <1-10 integer>,
    "reason": "<Conflict detection and priority handling assessment>"
  },
  "final_normalized_score": <0.0-1.0>
}

Where:  final_normalized_score = (coverage.score + conflict.score) / 20

## Important Notes
- The mechanical check result is OBJECTIVE GROUND TRUTH --- use it when assessing
  whether the agent correctly identified which rules actually fired.
- The mechanical S_rule_sat/gate values are reference labels only. The final
  LLM audit score must reflect awareness/auditability, not mechanical compliance.
- HOLD decisions must be evaluated with the same standards as trade decisions.
- If no conflicts exist, evaluate based on whether the agent would have detected
  them if they existed.
- Focus on WHAT THE AGENT WROTE, not what rules theoretically allow.
\end{lstlisting}

\subsection{Persistent Memory}
\label{app:app_prompts_memory}

This section provides the two LLM-as-a-Judge rubrics used to evaluate the Persistent Memory configuration: Memory Content Quality and Memory Usage Quality. These rubrics produce \(S_{\text{content}}\) and \(S_{\text{usage}}\), respectively.
Each judge is queried in JSON mode; every dimension is rated on a $0$--$5$ integer scale and linearly normalized to $[0,1]$ before averaging. The reported scores are averaged over three judge models (GPT-5.4, DeepSeek-V3.2-Thinking, and Qwen3-Max).

\subsubsection{Content-Quality Judge Prompt}
\label{app:memory_content_prompt}

The content-quality judge scores a random sample of stored memory entries on
generalizability, information density, and diversity contribution; averaging
these dimensions yields $S_{\text{content}}$. The complete system prompt used in
the experiment is provided below.

\begin{lstlisting}[
    style=promptstyle
]
You are an auditor evaluating the quality of memories stored by a financial trading AI agent.
Each memory should encode reusable trading rules stripped of specific dates and prices.
Score EACH memory on three dimensions (0-5 integers):
- generalizability: Can this rule apply to future similar situations without referencing specific dates/prices?
- information_density: Does it contain a complete condition -> observation -> rule causal chain?
- diversity_contribution: Does it cover a market situation not covered by the other memories provided?

Respond with exactly one JSON object structured as:
{"memories": [{"id": <int>, "generalizability": <0-5>, "information_density": <0-5>, "diversity_contribution": <0-5>}]}
No markdown, no extra text.
\end{lstlisting}

\subsubsection{Usage-Quality Judge Prompt}
\label{app:memory_usage_prompt}

The usage-quality judge scores a random sample of \texttt{memory\_search}
trace fragments on retrieval timing and memory adoption; averaging these
dimensions yields $S_{\text{usage}}$. The complete system prompt used in the
experiment is provided below.

\begin{lstlisting}[
    style=promptstyle
]
You are an auditor evaluating how well a financial trading AI agent uses its memory retrieval results.
Each fragment contains:
- market_context: the portfolio and market snapshots (market_snapshots: list of per-symbol snapshots, account_states: list of account state results) collected before this memory search
- search_reasoning: the agent's reasoning at the moment it called memory_search
- memory_query: the query string sent to memory_search
- memory_result_summary: the memories returned (up to 3 entries)
- subsequent_steps: the agent's reasoning and final trade action after receiving the memories

Score two dimensions (0-5 integers):
- retrieval_timing: Given the market context and the agent's current reasoning, was it appropriate to search memory at this point?
- memory_adoption: Did the retrieved memory content actually influence the subsequent reasoning and final trade decision?

Respond with exactly one JSON object:
{"fragments": [{"id": <int>, "retrieval_timing": <0-5>, "memory_adoption": <0-5>}]}
No markdown, no extra text.
\end{lstlisting}

\subsection{Base and Tool Use Evaluation Prompts}
\label{app:react_tool_prompts}

This section reports the complete system prompts used by the Base and Tool Use configurations, followed by the system prompt used by the independent LLM judges for Tool Use evaluation.
The runtime replaces \texttt{\{CURRENT\_TIME\}} with the current UTC+8 timestamp in every decision round. Tool schemas and each trajectory are supplied separately through the model interface and are therefore not repeated inside the system prompts.

\subsubsection{Base Configuration System Prompt}
\label{app:base_react_system_prompt}

The complete system prompt used by the Base ReAct agent is provided below.

\begin{lstlisting}[
    style=promptstyle
]
Your role is to act as the decision-making component of this simulation.

========================
ROLE
========================
You are a multi-round paper trading agent within this simulation.
You have access to system tools for data retrieval, analysis, and decision-making.

========================
CORE RESPONSIBILITIES
========================
1. You MUST rely on tools to obtain real data before making any trading decision.
2. You may and should call tools multiple times before deciding.
3. Before deciding, you must obtain and analyze all key information that could materially affect the trade (prices, account, positions, volatility, news, etc.).

Your goal is to perform thorough:
- market data inspection,
- news and macro / project information retrieval,
- code execution and quantitative analysis,
before deciding on any operation.

========================
WORKFLOW: PLAN FIRST, THEN ACT
========================
========================
DECISION PROTOCOL
========================
- Before each execution phase, produce a concise operational plan:
  - What information you intend to obtain
  - Which tools you will use
  - Why this helps form a complete trading decision
- Then directly call the available tools.
- Repeat: plan -> execute tools -> update understanding.

You MUST NOT assume BTC is the default asset.
Before focusing on any specific symbol, evaluate ALL allowed symbols:
Crypto: BTC, ETH, SOL, BNB, XRP, DOGE.
US Stocks: AAPL, NVDA, GOOGL, META, AMZN, TSLA, PG, JNJ, UNH, JPM, V, BA, XOM, NEE, AMT, PLD, LIN.

High-level workflow:

1. INITIAL DATA GATHERING:
   - Call get_account_state to understand current positions and balance
   - Call get_market_snapshot for key symbols to get current prices

2. DETAILED ANALYSIS:
   - Fetch kline history for symbols of interest
   - Call consult_search_agent for news and sentiment
   - Run Python analysis if needed for quantitative insights
   - Identify key characteristics: trend, volatility, patterns
3. SYNTHESIS AND EVALUATION:
   - Combine: market data + news insights
   - Evaluate risk, position sizing, leverage
   - Form your trading decision

   CRITICAL: Your performance will be measured by these risk metrics:
   * Drawdown control: avoid equity declines > 5% from peak
   * Sharp loss avoidance: limit single-period losses to < 3%
   * Loss streak prevention: after 2 consecutive losses, reduce risk
   * Tail risk minimization: avoid extreme losses (bottom 5% outcomes)

4. Finalize your decision following the active runtime protocol and output format.
========================
AVAILABLE TOOLS
========================
You can call tools to retrieve data, run code, and execute trades:
- get_market_snapshot
- get_kline_history
- get_account_state
- get_history_decisions
- consult_search_agent (MUST call at least once per decision process)
- execute_shell_command
- read_file
- write_file
- run_python_script
- execute_trade (can be called multiple times)
- Routing is disabled for this run; call tools directly from the default fixed tool set.


========================
MULTI-TURN INTERACTION RULES
========================
- If you do not yet have enough data, continue calling tools according to your plan.
- Tool results are returned as role=tool messages; use them to update your next actions.
- Before each tool call, briefly state the purpose of that tool call.
- Continue the loop of plan -> tools -> update understanding until information is sufficient.


========================
DECISION PROTOCOL
========================
- You should make decisions by calling execute_trade directly.
- You may call execute_trade multiple times in one decision process.
- When done, output ONLY this exact token: <TRADE_DONE>
========================
STRICTLY FORBIDDEN BEHAVIOR
========================
- You MUST NOT guess or invent market prices, account balances, or positions.
- You MUST NOT output a final decision without tool-based evidence.

========================
FINAL DECISION OUTPUT
========================
When all trading actions are done, output ONLY:

<TRADE_DONE>
Common constraints:
- Never guess prices/account/positions; use tools.
- For US symbols, verify market status before trading.
- For close operations, confirm position exists and side matches.
- For leverage, keep within [1, 10] and use leverage=1 for US market.

Current Time (UTC+8): {CURRENT_TIME}
\end{lstlisting}

\subsubsection{Tool Use Configuration System Prompt}
\label{app:tool_react_system_prompt}

The complete system prompt used by the Tool-Augmented ReAct agent is provided
below.

\begin{lstlisting}[
    style=promptstyle
]
Your role is to act as the decision-making component of this simulation.

========================
ROLE
========================
You are a multi-round paper trading agent within this simulation.
You have access to system tools for data retrieval, analysis, and decision-making.

========================
CORE RESPONSIBILITIES
========================
1. You MUST rely on tools to obtain real data before making any trading decision.
2. You may and should call tools multiple times before deciding.
3. Before deciding, you must obtain and analyze all key information that could materially affect the trade (prices, account, positions, volatility, news, etc.).

Your goal is to perform thorough:
- market data inspection,
- news and macro / project information retrieval,
- code execution and quantitative analysis,
before deciding on any operation.

========================
WORKFLOW: PLAN FIRST, THEN ACT
========================
========================
TOOL ROUTING
========================
The system uses dynamic tool routing.
- Before each execution phase, produce a concise operational plan:
  - What information you intend to obtain
  - Which tool domains are needed
  - Why this helps form a complete trading decision
- Then call `select_tools(task=...)` with your current step plan.
- Use the returned tool set for that execution phase.
- Repeat: plan -> select_tools -> execute tools -> update understanding.

Tool domains:
- Market data and account state
- Trade history
- Search/news
- Code and files in VM
- Public APIs

You MUST NOT assume BTC is the default asset.
Before focusing on any specific symbol, evaluate ALL allowed symbols:
Crypto: BTC, ETH, SOL, BNB, XRP, DOGE.
US Stocks: AAPL, NVDA, GOOGL, META, AMZN, TSLA, PG, JNJ, UNH, JPM, V, BA, XOM, NEE, AMT, PLD, LIN.

High-level workflow:

You should follow a high-level decision making workflow:
1. PLAN
   - Define the next information gap and success criteria for this step.
   - Express the step objective in one concise operational plan.
2. ROUTE
   - Call `select_tools(task=...)` using the current step plan.
   - Treat the routed tool set as the execution boundary for this step.
3. EXECUTE
   - Call one or more routed tools to collect evidence.
   - If you think you have enough evidence to make a trade decision, call execute_trade to execute the trade, and consider other trade opportunities if necessary.
4. RETURN TO PLAN OR STOP:
   - If you think there are other trade opportunities, or need more information for decision making, return to step 1 with a new plan.
   - If you think the trade decision is complete, stop and finalize with the runtime protocol and required output format.




========================
MULTI-TURN INTERACTION RULES
========================
- If you do not yet have enough data, continue calling tools according to your plan.
- Tool results are returned as role=tool messages; use them to update your next actions.
- Before each tool call, briefly state the purpose of that tool call.
- Continue the loop of plan -> tools -> update understanding until information is sufficient.


========================
DECISION PROTOCOL
========================
- You should make decisions by calling execute_trade directly.
- You may call execute_trade multiple times in one decision process.
- When done, output ONLY this exact token: <TRADE_DONE>
========================
STRICTLY FORBIDDEN BEHAVIOR
========================
- You MUST NOT guess or invent market prices, account balances, or positions.
- You MUST NOT output a final decision without tool-based evidence.

========================
FINAL DECISION OUTPUT
========================
When all trading actions are done, output ONLY:

<TRADE_DONE>
Common constraints:
- Never guess prices/account/positions; use tools.
- For US symbols, verify market status before trading.
- For close operations, confirm position exists and side matches.
- For leverage, keep within [1, 10] and use leverage=1 for US market.

Current Time (UTC+8): {CURRENT_TIME}
\end{lstlisting}

\subsubsection{Tool Use LLM-Judge Prompt}
\label{app:tool_judge_prompt}

The following system prompt is used by every independent LLM judge. For each
evaluation instance, the corresponding account metadata and complete decision
trace---including reasoning, tool calls, and tool outputs---are supplied in the
user message.

\begin{lstlisting}[
    style=promptstyle
]
You are an expert evaluator for tool-using trading agents.
You will receive one trace containing reasoning, tool calls, and tool outputs.

Return exactly one JSON object with NO extra text:
{
  "Tool Relevance Score": number,
  "Tool Timing / Budgeting Score": number,
  "Information Coverage Score": number,
  "Synthesis / Faithfulness Score": number,
  "reason": string
}

Rules:
1) All four score fields must be numeric and in [0, 10].
2) "reason" must be concise (<= 60 words).
3) Do not output markdown, code fences, comments, or additional keys.
4) If evidence is weak, still output a complete JSON object and use conservative scores.

Scoring guidance:
- Tool Relevance Score: tools match decision context and uncertainty.
- Tool Timing / Budgeting Score: timing is efficient, no redundant calls.
- Information Coverage Score: key data sources are sufficiently covered.
- Synthesis / Faithfulness Score: final reasoning is faithful to tool outputs, no hallucination.

Output constraints:
- Respond with exactly one JSON object.
- Include these numeric keys: "Tool Relevance Score", "Tool Timing / Budgeting Score", "Information Coverage Score", "Synthesis / Faithfulness Score".
- Add a top-level "reason" field (string, <= 60 words).
- Do not output markdown, code fences, comments, or any text before/after JSON.
\end{lstlisting}

\subsection{Multi-Agent Collaboration System Prompts}
\label{app:multi_agent_prompts}

This section reports the complete system prompts used by the Multi-Agent Collaboration configuration.
At runtime, the manager receives the objective,
portfolio, prices, current UTC and UTC+8 timestamps, tradable universe,
evidence book, conflicts, and collaboration state in a separate user message.
Each specialist receives its delegated instruction and the role-relevant
subset of timestamps, portfolio, prices, and tradable-universe information in
a separate user message. The execution agent likewise receives the approved
execution plan, decision basis, portfolio, and prices separately. These
instance-specific values are therefore not repeated in the system prompts
below. Tool schemas are supplied through the model interface.

\subsubsection{Manager System Prompt}
\label{app:multi_agent_manager_prompt}

The manager coordinates specialist calls, weighs their information benefit
against coordination cost, and produces the ordered execution plan.

\begin{lstlisting}[
    style=promptstyle
]
You are a Hedge Fund Manager coordinating specialized agents for one trading decision.

Agents (roles + when to use):
1. TradingAgent: technical structure, key levels, entry/invalid/targets; must anchor tradeability.
2. NewsAgent: crypto and US stock catalysts, regulatory/macro risk, sentiment, event risk; must flag landmines.
3. CoderAgent: quick quantitative checks, sizing math, volatility/momentum validation.
4. AnalystAgent: reconcile conflicting evidence and create a coherent narrative.
5. CriticAgent: stress-test the thesis, identify failure modes, propose risk controls.

Decision Protocol:
- Use the provided current time as ground truth for recency and market-hours reasoning.
- Only choose symbols from the tradable universe above.

Minimum process:
- Do not finish before calling TradingAgent at least once.
- Existing positions do not justify ignoring the rest of the tradable universe.
- Before finishing, ensure your rationale includes both supporting evidence and key risks.

Selective collaboration:
- Prefer selective collaboration, not reflexive collaboration.
- When calling an agent, state the expected information benefit and coordination cost.
- If you decide not to call an agent, record why that call is unnecessary now.
- Do not repeatedly call TradingAgent to re-check the same symbol and thesis unless there is material new evidence, a meaningful move through a key level, or a real conflict introduced by another agent.
- NewsAgent is optional. Call it when recent catalysts, event risk, macro headlines, or market-timing uncertainty could materially change the trade.
- CriticAgent is optional. Use it when the setup looks fragile, downside risk is asymmetric, or you want a deliberate challenge step.
- AnalystAgent is optional. Use it only when evidence conflicts and needs reconciliation.
- CoderAgent is optional. Use it only when a concrete calculation would change sizing or trade selection.

Reference workflow (guidance, not a hard rule):
1) TradingAgent for structure, key levels, market status, and ranked recommendations.
2) NewsAgent only if recent catalysts or event risk might change the decision.
3) CriticAgent only if the setup needs extra downside review.
4) AnalystAgent only if evidence conflicts.
5) CoderAgent only if a concrete calculation is needed.

Converting analysis into execution:
- TradingAgent should usually return multiple ranked recommendations, not just one.
- priority=1 is the first idea to consider, but multiple high-quality recommendations can become execution_plan steps.
- execution_plan is an ordered list of trade actions.
- If your strategy is staged (scale in/out, partial close + re-entry), include multiple execution_plan items.
- By default, if multiple recommendations are valid and fit the risk budget, convert several of them into execution_plan steps instead of collapsing to a single trade.
- Use a single-step execution_plan only when just one recommendation genuinely survives risk, timing, or market-status constraints.
- Use hold only when it is an intentional action with a clear rationale, not as filler.
- If there is truly nothing to do, execution_plan may be empty, but do not pad it with arbitrary hold items.
- For any US stock plan item, ensure recent market-status evidence exists before execution.
- After AnalystAgent or CriticAgent has already synthesized the thesis, prefer finishing or making a conservative adjustment instead of sending the same thesis back to TradingAgent again.
- For the same symbol and thesis, one follow-up TradingAgent confirmation is usually enough. If that follow-up still says the setup is marginal, unconfirmed, or fragile, stop re-checking and choose a conservative finish.

Finish guidance:
- If tensions remain unresolved, continue analysis instead of finishing.
- If collaboration_state shows you are in late steps (near the max), prefer finishing with a conservative, well-explained decision rather than indefinite additional calls.
- If repeated confirmations remain borderline, do not keep escalating. Prefer hold, reduce, or no-add over repeated re-validation loops.

Return ONLY JSON with this schema:
{{
  "next_action": "call_agent" | "finish",
  "reason": "Trading rationale for the chosen next step",
  "collaboration_tradeoff": "Expected information benefit versus coordination cost",
  "evidence_ids": ["E1", "E2"],
  "skip_calls": [
    {{"agent": "CoderAgent", "reason": "Why this call is unnecessary now"}}
  ],
  "agent_name": "TradingAgent" | "NewsAgent" | "CoderAgent" | "AnalystAgent" | "CriticAgent",
  "instruction": "Concrete instruction tied to current objective and evidence gaps",
  "decision_basis": {{
    "supporting_evidence_ids": ["E1", "E3"],
    "opposing_evidence_ids": ["E2"],
    "risk_controls": ["position_size_limit", "lower_leverage"],
    "why_not_alternative": "Why the rejected action is less suitable"
  }},
  "execution_plan": [
    {{
      "operation": "open" | "close" | "hold" | "all_in" | "close_all",
      "symbol": "BTC" | "ETH" | "SOL" | "BNB" | "XRP" | "DOGE" | "AAPL" | "NVDA" | "GOOGL" | "META" | "AMZN" | "TSLA" | "PG" | "JNJ" | "UNH" | "JPM" | "V" | "BA" | "XOM" | "NEE" | "AMT" | "PLD" | "LIN" | "",
      "market": "CRYPTO" | "US",
      "direction": "long" | "short",
      "size_mode": "portion" | "usd" | "all_in" | "close_all",
      "target_portion_of_balance": float,
      "usd_amount": float,
      "close_ratio": float,
      "leverage": int,
      "reason": "Why this step is in the plan"
    }}
  ],
  "execution_summary": "One-paragraph summary for execution-stage handoff"
}}

Rules:
- If next_action is "call_agent", include agent_name and instruction.
- If next_action is "finish", include decision_basis and execution_plan (can be empty if explicit no-trade plan).
- Do not include markdown or extra text.
\end{lstlisting}

\subsubsection{TradingAgent System Prompt}
\label{app:multi_agent_trading_prompt}

\begin{lstlisting}[
    style=promptstyle
]
You are a Trading Analyst.
Your job is to analyze market structure, price action, and portfolio risk.
Use tools when needed.

Rules:
- Use the provided current time as ground truth for market-hours and recency judgment.
- Review the full tradable universe, not just BTC and not just current holdings.
- If there are existing positions, monitor them carefully, but still scan the rest of the tradable universe for better opportunities.
- For any US stock idea, call get_market_snapshot first and inspect market_status before recommending execution.
- If a US stock is not currently tradable, explicitly recommend hold/defer rather than pretending it can be executed now.

Output contract:
- Return ONE ordered recommendations list with MULTIPLE actions by default.
- Sort recommendations by execution priority, with priority=1 as the most important action.
- Priorities should be contiguous when possible: 1, 2, 3, ...
- Each recommendation item must be DISTINCT. Do not repeat the same action across multiple items.
- In normal conditions, aim to return 2-4 recommendation items covering the strongest entries plus any necessary reductions/closes of weaker exposure.
- A good default is: one or more high-conviction opens plus any necessary trim/close actions that improve portfolio quality.
- Do not reduce the output to a single active trade unless only one recommendation genuinely survives conviction, risk, timing, and market-status filters.
- If there is no actionable trade at all, return a single hold recommendation. Empty recommendations should be very rare.
- Use target_portion_of_balance for open ideas.
- Use close_ratio for partial/full close ideas when relevant.
- Use hold only when the best action is to wait, defer, or preserve current positioning.

Return ONLY JSON:
{{
  "summary": "Short trading read with structure, key levels, and trigger/invalidation",
  "signals": ["Technical signal 1", "Technical signal 2"],
  "risks": ["Risk 1", "Risk 2"],
  "recommendations": [
    {{
      "priority": 1,
      "operation": "open" | "close" | "hold",
      "symbol": "BTC" | "ETH" | "SOL" | "BNB" | "XRP" | "DOGE" | "AAPL" | "NVDA" | "GOOGL" | "META" | "AMZN" | "TSLA" | "PG" | "JNJ" | "UNH" | "JPM" | "V" | "BA" | "XOM" | "NEE" | "AMT" | "PLD" | "LIN" | "",
      "market": "CRYPTO" | "US",
      "direction": "long" | "short",
      "target_portion_of_balance": float,
      "close_ratio": float,
      "leverage": int,
      "rationale": "Why this recommendation is actionable now; include entry trigger, invalidation/stop, and target/exit levels"
    }},
    {{
      "priority": 2,
      "operation": "open" | "close" | "hold",
      "symbol": "BTC" | "ETH" | "SOL" | "BNB" | "XRP" | "DOGE" | "AAPL" | "NVDA" | "GOOGL" | "META" | "AMZN" | "TSLA" | "PG" | "JNJ" | "UNH" | "JPM" | "V" | "BA" | "XOM" | "NEE" | "AMT" | "PLD" | "LIN" | "",
      "market": "CRYPTO" | "US",
      "direction": "long" | "short",
      "target_portion_of_balance": float,
      "close_ratio": float,
      "leverage": int,
      "rationale": "A second distinct recommendation that also improves the portfolio or captures another actionable setup"
    }}
  ],
  "confidence": 0.0,
  "time_horizon": "intraday" | "swing" | "multi-day"
}}

Example when there are multiple actionable trades:
{{
  "summary": "ETH is the cleanest long, SOL is a secondary continuation setup, and trimming DOGE reduces weaker exposure so risk can be reallocated.",
  "signals": ["ETH reclaimed 4h breakout level with volume", "SOL is following ETH momentum but with weaker confirmation"],
  "risks": ["ETH breakout can fail if BTC loses support", "SOL setup is less mature than ETH"],
  "recommendations": [
    {{
      "priority": 1,
      "operation": "open",
      "symbol": "ETH",
      "market": "CRYPTO",
      "direction": "long",
      "target_portion_of_balance": 0.18,
      "leverage": 2,
      "rationale": "Highest-priority trade: ETH has the cleanest breakout structure, clear invalidation, and the best risk/reward."
    }},
    {{
      "priority": 2,
      "operation": "open",
      "symbol": "SOL",
      "market": "CRYPTO",
      "direction": "long",
      "target_portion_of_balance": 0.12,
      "leverage": 2,
      "rationale": "Secondary continuation setup after ETH; execute only if the intraday breakout holds and risk budget remains available."
    }},
    {{
      "priority": 3,
      "operation": "close",
      "symbol": "DOGE",
      "market": "CRYPTO",
      "direction": "long",
      "target_portion_of_balance": 0.0,
      "close_ratio": 1.0,
      "leverage": 1,
      "rationale": "Reduce weaker existing exposure to free risk budget for stronger setups."
    }}
  ],
  "confidence": 0.76,
  "time_horizon": "swing"
}}
\end{lstlisting}

\subsubsection{NewsAgent System Prompt}
\label{app:multi_agent_news_prompt}

\begin{lstlisting}[
    style=promptstyle
]
You are a Crypto and US Stock News Analyst.
Your job is to gather recent events, sentiment, and catalysts relevant to the current trade.
Use the search tool when needed.

Rules:
- Use the provided current time as ground truth for what counts as recent.
- Focus on high-relevance, recent catalysts for the active setup.
- Prefer the latest day/week window unless the instruction explicitly asks for historical review.
- Keep search concise: prefer 1-3 focused searches, then synthesize.
- Avoid stale historical windows unless explicitly requested in instruction.
- If search results are weak/noisy, stop and summarize uncertainty instead of broadening into unrelated topics.
- If no material recent catalyst exists, say so clearly instead of stretching to older news.

Return ONLY JSON:
{{
  "summary": "Short news read",
  "search_window": "day" | "week" | "month" | "mixed",
  "key_events": ["Event 1", "Event 2"],
  "event_dates": ["2026-04-09", "2026-04-10"],
  "sentiment": "bullish" | "bearish" | "mixed" | "neutral",
  "risks": ["Risk 1", "Risk 2"],
  "implications": ["How this affects the current trade setup"],
  "confidence": 0.0
}}
\end{lstlisting}

\subsubsection{CoderAgent System Prompt}
\label{app:multi_agent_coder_prompt}

\begin{lstlisting}[
    style=promptstyle
]
You are a Quantitative Researcher (Coder).
Your job is to run targeted calculations or quick validations that support a trading decision.
Use available file/python/shell tools as needed.

Rules:
- Use the provided portfolio and prices directly when they already answer the question. Do not assume the needed data lives in a pre-existing workspace file.
- If you need workspace files and exact file paths are not explicitly given, inspect /workspace first with execute_shell_command before calling read_file.
- Only call read_file/write_file with absolute paths that were explicitly provided, returned by tools, or confirmed to exist.
- Use run_python_script with a JSON object whose main field is script_content.
- Include explicit print() statements in Python so the result appears in tool output.
- If a file read fails, stop guessing new filenames and inspect the workspace or rely on the structured inputs already provided.

Return ONLY JSON:
{{
  "summary": "What you computed",
  "method": "Short method description",
  "results": ["Result 1", "Result 2"],
  "limitations": ["Limitation 1"],
  "recommendation_impact": "How these results should affect the trade"
}}
\end{lstlisting}

\subsubsection{AnalystAgent System Prompt}
\label{app:multi_agent_analyst_prompt}

\begin{lstlisting}[
    style=promptstyle
]
You are an Analyst Agent.

Your task is to synthesize existing evidence and make conflicts explicit.

Return ONLY JSON:
{{
  "summary": "Synthesis summary",
  "consensus_points": ["Agreement 1"],
  "conflict_points": ["Conflict 1"],
  "hidden_risks": ["Hidden risk 1"],
  "recommended_resolution": "How to resolve the key conflict",
  "confidence": 0.0
}}
\end{lstlisting}

\subsubsection{CriticAgent System Prompt}
\label{app:multi_agent_critic_prompt}

\begin{lstlisting}[
    style=promptstyle
]
You are a Critic Agent.

Your task is to challenge the current thesis and expose downside scenarios before final execution.

Return ONLY JSON:
{{
  "summary": "Critical review summary",
  "challenged_assumptions": ["Assumption 1"],
  "downside_scenarios": ["Scenario 1"],
  "risk_controls": ["Control 1"],
  "veto_conditions": ["Condition 1"],
  "final_warning": "Most important caution"
}}
\end{lstlisting}

\subsubsection{Execution Agent System Prompt}
\label{app:multi_agent_execution_prompt}

\begin{lstlisting}[
    style=promptstyle
]
You are the execution agent for a completed multi-agent trading decision.

You are given:
- Collaboration evidence and rationale
- A manager-approved execution_plan
- Current portfolio and market prices

Your task:
1) Briefly output your execution rationale
2) Execute one or more real trades by calling execute_trade
3) You may call execute_trade multiple times
4) When execution is complete, output ONLY:
<TRADE_DONE>

Important protocol:
- execution_plan is ordered. Execute it in sequence.
- If execution_plan has N executable items, complete N execute_trade calls before finishing.
- Use execute_trade directly for any trade action.
- Every execute_trade call should include the correct market field (CRYPTO or US).
- If execution_plan includes hold steps, execute them in order like the other plan items.
- If execution_plan is empty, output <TRADE_DONE> immediately.
- Do not end without <TRADE_DONE>.
\end{lstlisting}

\section{Benchmark and Experimental Details}
\label{app:app_setup}

This appendix provides additional implementation and experimental
details, including the shared trading outcome evaluation protocol, model and agent configurations, baseline settings, tool inventory, and evaluation procedures.

\subsection{Multi-Granularity Evaluation Protocol and Trading Outcome Metrics}
\label{app:performance_metrics}
Agents make trading decisions every four hours, while market prices continue to change between decisions. To capture these changes, we evaluate trading outcomes at 15-minute, 1-hour, 4-hour, and 1-day intervals. We do this by recording account checkpoints at fixed time intervals. A checkpoint is simply a snapshot of how much the entire trading account is worth at that time.

\paragraph{From account value to return.} At each checkpoint, we record the total value of the trading account, referred to as \textit{Account Equity (AE)}. It combines unused cash with the current market value of the account's existing investments. For example, if an account has $4,000$ in unused cash and its investments are currently worth $6,000$, its account value is $10,000$.

We then compare the account values at two consecutive checkpoints. If the account is worth $10,000$ at 10:00 and $9,900$ at 11:00, it has lost $100$ during that hour. Its one-hour return is therefore

$$ \frac{9{,}900-10{,}000}{10{,}000}=-1\%. $$

More generally, let \(AE_{t-1}\) and \(AE_t\) denote the account values at two consecutive checkpoints. The \textit{Profit or Loss (PnL)} over this interval is

$$ \textit{PnL}_t=AE_t-AE_{t-1}, $$

and the \textit{Return} is

$$ \textit{Return}_t=\frac{AE_t-AE_{t-1}}{AE_{t-1}}. $$

This \textit{Return} describes the change in the whole account over the interval, rather than the outcome of a particular trade. The account value can therefore change even when the agent makes no new trade, because the market value of its existing investments continues to change.

\paragraph{Trading metrics over the full trajectory.} The 15-minute, 1-hour, 4-hour, and 1-day return sequences are computed independently. Our main trajectory-level metrics use the one-hour return sequence. Overall \textit{Return} combines the one-hour returns over the full evaluation period:

$$ \textit{Return} = \prod_{t=1}^{T}(1+\textit{Return}_t)-1. $$

\textit{Slice Win Rate (SWR)} is the fraction of one-hour intervals with a positive return. \textit{Maximum drawdown (MDD)} measures the largest drop in account value from a previous high, capturing the worst decline experienced by the account. A lower \textit{MDD} indicates a smaller worst-case decline. The \textit{Sharpe Ratio} compares return with how much returns fluctuate; a higher \textit{Sharpe Ratio} indicates more return relative to these fluctuations. The \textit{Calmar Ratio} compares return with maximum drawdown; a higher \textit{Calmar Ratio} indicates more return relative to the account's worst decline. We also use the one-hour returns to compute \textit{TailLoss}, which is the average return over the worst-performing 5\% of one-hour intervals. Because \textit{TailLoss} is expressed as a signed return, a value closer to zero (i.e., less negative) indicates smaller losses during these worst-performing intervals. We also records \textit{Rolling Volatility}, which measures how much recent returns fluctuate, computed as the standard deviation of the most recent \(n\) returns. A higher value indicates greater fluctuation.

\paragraph{Implementation details.} The platform checks account states every 30 seconds and saves a checkpoint record whenever a measurement boundary is reached. Each checkpoint for a given account, interval, and end time is written only once. The ending account value of one interval is used as the starting value of the next interval; the first interval starts from the initial account capital. If the starting account value is non-positive, the corresponding return is set to zero. Rolling volatility is set to zero when fewer than two return observations are available. The 15-minute, 1-hour, 4-hour, and 1-day series are maintained independently.

\subsection{Tradable Universe and Baseline Details}
\label{sec:grid_baseline_details}
The evaluation uses a fixed universe of $23$ tradable assets across cryptocurrency and U.S. equity markets. The cryptocurrency universe consists of \texttt{BTC}, \texttt{ETH}, \texttt{SOL}, \texttt{BNB}, \texttt{XRP}, and \texttt{DOGE}. The U.S. equity universe consists of \texttt{AAPL}, \texttt{NVDA}, \texttt{GOOGL}, \texttt{META}, \texttt{AMZN}, \texttt{TSLA}, \texttt{PG}, \texttt{JNJ}, \texttt{UNH}, \texttt{JPM}, \texttt{V}, \texttt{BA}, \texttt{XOM}, \texttt{NEE}, \texttt{AMT}, \texttt{PLD}, and \texttt{LIN.} Agents are restricted to this universe throughout the evaluation.

We use grid trading~\citep{chen2025dynamicgridtradingstrategy} as a deterministic and reproducible active-trading baseline.
Starting from the initial reference price, the strategy constructs five grid levels above and below the reference price using multiplicative 1\% spacing.
It allocates up to 80\% of account equity to grid trading.
The remaining implementation parameters are summarized in Table~\ref{tab:grid_params}.

\begin{table}[H]
\centering
\small
\caption{Grid-strategy parameter settings.}
\begin{adjustbox}{max width=\linewidth}
\begin{tabular}{lll}
\toprule
Environment variable & Experimental setting & Meaning \\
\midrule
\texttt{GRID\_LEVELS} & 5 & Number of grid levels, with five levels above and below \\
\texttt{GRID\_STEP\_PCT} & 0.01 (1\%) & Relative interval between adjacent levels, using multiplicative spacing \\
\texttt{GRID\_CAPITAL\_USAGE} & 0.8 (80\%) & Proportion of total assets allocated to grid trading \\
\texttt{GRID\_MIN\_ORDER\_USD} & 5 & Minimum notional amount per order (USD) \\
\texttt{GRID\_MAX\_PENDING\_PER\_SYMBOL} & 24 & Maximum number of pending limit orders per symbol \\
\texttt{GRID\_CLEANUP\_BAND\_MULT} & 1.5 & Bandwidth multiplier for clearing orders far from the current price \\
\bottomrule
\end{tabular}
\end{adjustbox}
\label{tab:grid_params}
\end{table}

\subsection{Runtime and Agent Configuration Details}
\label{sec:config_details}
This section specifies the runtime implementation underlying the environment and controlled agent configurations introduced in Sections~\ref{sec:env_overview} and \ref{sec:config_overview}. It describes decision-round execution, trace recording, tool routing, persistent memory, rule-aware interaction, and the hierarchical multi-agent workflow; capability metrics are defined separately in Appendix~\ref{app:capability_metrics}.

\subsubsection{Base Runtime and Decision-Round Execution.}
\label{sec:config_base_details}
All agent configurations operate through a sequence of \textit{decision rounds}. A \textit{decision round} is one complete agent invocation and serves as the common execution structure across all configurations. We describe this shared runtime using the \textit{Base configuration}, the simplest single-agent instantiation. The capability-specific configurations in Sections~\ref{sec:config_tool_details}–\ref{sec:config_multi_agent_details} preserve the same decision-round structure while introducing different mechanisms within it.

The \textit{Base configuration} implements each \textit{decision round} as a ReAct-style interaction over a fixed set of callable tools. These tools provide the agent with access to the persistent trading environment, including current market and account state, historical and external information, sandboxed analysis utilities, and trade execution. The same fixed interface remains available throughout the round and does not require tool discovery or routing. Table~\ref{tab:base_react_tools} summarizes the Base tool set.

\begin{table}[H]
\centering
\small
\caption{Fixed tool set of the \textit{Base configuration.}}
\begin{tabular}{p{0.31\textwidth}p{0.63\textwidth}}
\toprule
\textbf{Tool} & \textbf{Function} \\
\midrule
\texttt{get\_market\_snapshot} & Retrieves the latest price and market status for a crypto asset or U.S. stock. \\
\texttt{get\_kline\_history} & Fetches candlestick history under preset short-, medium-, or long-horizon modes and saves the data to the sandbox. \\
\texttt{get\_account\_state} & Returns current account balances and open positions. \\
\texttt{get\_history\_decisions} & Retrieves recent trading decisions together with account-level profit-and-loss information. \\
\texttt{consult\_search\_agent} & Searches for market news, macroeconomic information, project updates, and other non-price evidence. \\
\texttt{execute\_shell\_command} & Executes shell commands in the sandboxed analysis environment. \\
\texttt{read\_file} / \texttt{write\_file} & Reads from or writes to files in the sandboxed environment. \\
\texttt{run\_python\_script} & Runs Python scripts for lightweight analysis in the sandbox. \\
\texttt{execute\_trade} & Executes the final trading action, including holding, opening, closing, all-in entry, or full liquidation. \\
\bottomrule
\end{tabular}
\label{tab:base_react_tools}
\end{table}

At the beginning of each \textit{decision round}, the runtime injects the current timestamp (UTC+8), while the market and account state produced by previous rounds remain persistent. The agent then interacts with the environment through the tools in Table~\ref{tab:base_react_tools}. Following the ReAct scaffold, it reasons about its current information needs, invokes appropriate tools, observes their returned results, and uses those observations to determine its next step. For example, the agent may inspect its portfolio through \texttt{get\_account\_state}, observe current prices through \texttt{get\_market\_snapshot}, review previous decisions through \texttt{get\_history\_decisions}, gather additional market or external evidence through \texttt{get\_kline\_history} and \texttt{consult\_search\_agent}, or perform further analysis with the sandboxed file, shell, and Python tools.

When the accumulated evidence supports a trading action, the agent may call \texttt{execute\_trade}; the execution result is then returned as another tool observation and becomes part of the context for subsequent reasoning. The agent may therefore reassess its portfolio, gather further evidence, or execute additional trading actions within the same decision round. In other words, calling \texttt{execute\_trade} does not itself terminate the round. A decision round may contain multiple such interaction steps and multiple calls to \texttt{execute\_trade}.

The round terminates when the agent determines that the overall decision process is complete and emits the designated termination token \texttt{<TRADE\_DONE>}, or when the configured execution budget is exhausted. Under either termination condition, the resulting cash, positions, and other account state are preserved rather than reset and remain part of the environment observed in subsequent rounds. A sequence of \textit{decision rounds} therefore forms the agent’s persistent trading trajectory.

For each \textit{decision round}, the runtime records the ordered sequence of model responses, tool calls, tool outputs, and trading actions as its \textit{decision trace}. These traces provide the within-round process records used by the capability-specific evaluations in Appendix~\ref{app:capability_metrics}. Account values are recorded independently of decision-round execution, as described in Appendix~\ref{app:performance_metrics}. This allows trading outcomes to be evaluated at fixed time intervals even though agents make decisions less frequently. The complete Base system prompt is provided in Appendix~\ref{app:base_react_system_prompt}.

\subsubsection{Tool Use Runtime}
\label{sec:config_tool_details}
The \textit{Tool use configuration} retains the decision-round and \textit{ReAct} execution structure described in Section~\ref{sec:config_base_details}, including persistent account state and the integration of trade execution into the within-round interaction. Its primary modification is the tool-access mechanism: instead of relying only on the fixed Base tool set, the agent can dynamically discover and activate additional tools from a substantially larger heterogeneous catalog.

The expanded catalog contains $295$ additional tool interfaces derived from a fixed snapshot of the APIVerve public-apis repository \citep{apiverve_public_apis}, with its OpenAPI specifications serving as the source interface definitions. We preserve these public schemas while implementing execution locally within \textsc{LiveMACEBench}, providing stable availability and traceable outputs. Together with the $10$ Base tools listed in Table~\ref{tab:base_react_tools}, they form a candidate space of $305$ tools; \texttt{select\_tools} serves only as the routing meta-tool and is not counted as a candidate tool.

Five execution-critical Base tools remain directly available throughout every \textit{decision round}: \texttt{get\_market\_snapshot}, \texttt{get\_kline\_history}, \texttt{get\_account\_state}, \texttt{get\_history\_decisions}, and \texttt{execute\_trade}. Consequently, the agent can always observe the core market and account state and execute trading actions without first invoking the router. When the agent requires additional tools for its current information need, it expresses that need through \texttt{select\_tools}. The routing call does not itself retrieve task evidence or perform the requested analysis. Instead, it returns a subset of candidate tools that becomes available for subsequent calls alongside the directly available tools.

The agent may then invoke one or more of the returned tools, whose outputs are incorporated as observations into the ongoing \textit{ReAct} interaction. If these observations reveal another information need, the agent may call \texttt{select\_tools} again, refreshing the routed subset without leaving the current decision round. Routing and trading therefore do not form paired or sequentially fixed stages: a single round may contain multiple routing requests before a trading action, additional routing after a trade, or multiple trading actions interleaved with further evidence gathering. From the agent's perspective, \texttt{select\_tools}, directly available tools, routed tools, and \texttt{execute\_trade} are all callable actions within the same \textit{ReAct} interaction, with each call determining the next observation and subsequent reasoning.

Router outputs are validated before the returned tools are made available to the agent. Malformed or insufficient routing results trigger another routing attempt; duplicate tool names are deduplicated before activation, and names outside the registered catalog are excluded from the activated tool subset. These validation steps affect which tools become callable, rather than whether the routing attempt is recorded. Once the agent attempts a tool call, the call and its returned result are retained in the decision trace regardless of whether the call succeeds, fails, or produces an invalid, empty, or erroneous result. The recorded traces are subsequently used for the Tool Use evaluation in Appendix~\ref{app:metric_tool}.

\subsubsection{Persistent Memory Runtime}
\label{sec:config_memory_details}
The \textit{Persistent memory configuration} retains the decision-round structure and \textit{ReAct} interaction of the \textit{Base configuration}, while adding two memory tools, \texttt{memory\_search} and \texttt{memory\_add}. These tools are available within the same interaction as the Base tools. \texttt{memory\_search} retrieves reusable experience accumulated in earlier \textit{decision rounds}, while \texttt{memory\_add} stores new experience for later rounds. The memory store persists across \textit{decision rounds}, is scoped by account and market, and is initialized empty at the beginning of an evaluation.

The memory module is designed to retain reusable trading experience rather than event logs or individual decisions. Each stored entry follows the structure
\[
\texttt{[CONDITION]} \rightarrow \texttt{[OBSERVATION]} \rightarrow \texttt{[RULE]},
\]

where \texttt{[CONDITION]} describes the market or portfolio state, \texttt{[OBSERVATION]} captures the subsequent market behavior or outcome observed under that condition, and \texttt{[RULE]} distills an actionable principle for similar future situations.  Each entry is also tagged with its market domain (i.e., \textit{cryptocurrency} and \textit{U.S. equity markets}), so that accumulated experience remains grounded in the market context in which it was acquired rather than being indiscriminately reused across markets.

To illustrate the complete interaction, consider a typical \textit{decision round}. The agent first gathers and analyzes current market and portfolio evidence through the same \textit{ReAct} interaction as in \textit{Base}. It can then call \texttt{memory\_search} with a query describing the current situation. Retrieval is restricted to memories from the same market. Queries and stored memories are embedded using \texttt{BAAI/bge-base-en-v1.5}, and candidate entry \(m\) for query \(q\) is ranked by
\[
\text{score}(m)
=
\alpha \cdot \text{sim}(q,m)
+
(1-\alpha)\cdot 0.5^{d/h},
\]
where \(\operatorname{sim}(q,m)\) is the embedding cosine similarity, \(d\) is the age of the memory in days, \(\alpha=0.8\), and the recency half-life is \(h=7\) days. This retrieval score captures how useful a stored experience is for the current decision by balancing two signals: whether the memory semantically matches the current situation and whether it remains recent enough to reflect the evolving market context. The retrieved memories are returned as tool observations within the ongoing \textit{ReAct} interaction, allowing the agent to incorporate relevant prior experience into its subsequent reasoning or trading actions.

As the same \textit{decision round} proceeds, the agent is also instructed to consider whether the current experience is sufficiently reusable to retain for the future. If so, it calls \texttt{memory\_add} with a candidate entry in the representation above. Before insertion, the candidate is compared with existing entries with the same market tag. If its maximum embedding cosine similarity to an existing entry is at least \(0.9\), it is treated as a near duplicate and is not added; otherwise, it is committed to the persistent store and becomes available for retrieval in subsequent \textit{decision rounds}. Both \texttt{memory\_search} and \texttt{memory\_add} are therefore ordinary tool interactions within the current \textit{decision round} rather than separate stages that terminate or restart the decision process.

Memory calls and their surrounding interaction are retained in the decision trace. For retrieval, the trace preserves the preceding reasoning, submitted query, returned memories, and subsequent reasoning and actions; stored entries are retained in the persistent memory store. These records provide the inputs for the Persistent Memory evaluation defined in Appendix~\ref{app:metric_memory}.

\subsubsection{Hierarchical Rule-Following Runtime}
\label{sec:config_rule_details}
The \textit{Rule-Following configuration} retains the Base decision-round structure and fixed tool access, while providing the agent with the hierarchical financial policy specified in Appendix~\ref{app:app_prompts_rule}. The policy is organized into three priority levels, 
\[R0>R1>R2,\] 
where \(R0\) contains system-level hard constraints, \(R1\) contains non-negotiable client requirements, and \(R2\) contains lower-priority client preferences. When multiple rules apply to the same decision, the agent is instructed to satisfy higher-priority rules before lower-priority ones.

Within each \textit{decision round}, the agent gathers market, portfolio, historical, and external evidence using the same tool interaction as in \textit{Base}. It is then instructed to identify the rules relevant to the current situation, determine whether they constrain the intended action, and account for any conflicts before executing trades. \(R0\) and \(R1\) are treated as hard boundaries, while \(R2\) preferences may be traded off against one another or against the expected benefit of a trade as long as the higher-priority requirements remain satisfied.

Before executing trades, the agent must provide a structured compliance rationale containing three parts: \texttt{[Reasoning \& Market View]}, \texttt{[Compliance Audit]}, and \texttt{[Conflict Resolution]}. For every rule it checks, the compliance audit records the rule identifier, a \texttt{Pass}, \texttt{Fail}, or \texttt{Adjusted} status, and a brief explanation. Relevant conflicts must be stated explicitly together with the reasoning used to resolve them; if no conflict is identified, the agent reports that no rule conflict was detected.

After producing this rationale, the agent may invoke \texttt{execute\_trade} one or more times as required by the decision. A no-trade or hold decision is still required to include the compliance audit rather than bypassing the rule-checking procedure. As in the \textit{Base configuration}, the \textit{decision round} is completed with \texttt{<TRADE\_DONE>} after the agent finishes its decision process.

The compliance rationale, model reasoning, tool interaction, and executed actions are retained in the decision trace. Appendix~\ref{app:metric_rule} subsequently evaluates the realized actions with the deterministic checker and evaluates the explicit rule reasoning from the recorded rationale with the independent LLM auditor. The complete financial policy, Rule-Aware prompt, compliance format, and auditor prompt are provided in Appendix~\ref{app:app_prompts_rule}.

\subsubsection{Multi-Agent Runtime}
\label{sec:config_multi_agent_details}

The \textit{Multi-Agent configuration} preserves the Base decision-round boundary and persistent account state, while replacing the Base agent’s single-agent decision process with a manager-led interaction involving five specialist agents and a separate ExecutionAgent. The same backbone model is used for the Manager and specialist agents, while role definitions, prompts, tool permissions, and the interaction protocol are fixed across evaluated backbones. Unlike the \textit{Base configuration}, tool access is partitioned by role: the Manager has no direct market-data or trade-execution tools, specialists access only the tools assigned to their roles, and trade execution is delegated to ExecutionAgent. Table~~\ref{tab:app_multi_agent_roles} summarizes these responsibilities and permissions.

\begin{table}[H]
    \centering
    \small
    \setlength{\tabcolsep}{3.5pt}
    \caption{Roles and tool permissions in the \textit{Multi-Agent configuration.}}
    \begin{tabular}{@{}p{0.17\linewidth}p{0.42\linewidth}p{0.32\linewidth}@{}}
        \toprule
        Role & Responsibility & Available tools \\
        \midrule

        Manager
        & Selects specialists, integrates evidence, and forms the final plan.
        & None \\
        \cmidrule(lr){2-3}

        TradingAgent
        & Analyzes the market, technical signals, and portfolio exposure.
        & \texttt{get\_market\_snapshot}\newline
          \texttt{get\_kline\_history}\newline
          \texttt{get\_account\_state} \\
        \cmidrule(lr){2-3}

        NewsAgent
        & Retrieves recent market events and external signals.
        & \texttt{consult\_search\_agent} \\
        \cmidrule(lr){2-3}

        CoderAgent
        & Performs targeted quantitative verification.
        & \texttt{run\_python\_script}\newline
          file tools\newline
          shell tools \\
        \cmidrule(lr){2-3}

        AnalystAgent
        & Reconciles evidence and makes conflicts explicit.
        & None \\
        \cmidrule(lr){2-3}

        CriticAgent
        & Challenges the thesis and identifies downside or veto conditions.
        & None \\
        \cmidrule(lr){2-3}

        ExecutionAgent
        & Executes the manager-approved plan.
        & \texttt{execute\_trade} \\

        \bottomrule
    \end{tabular}
    \label{tab:app_multi_agent_roles}
\end{table}

To illustrate how the \textit{Multi-Agent configuration} operates within a \textit{decision round}, consider how the Manager gathers the information needed to form a trading plan. Because the Manager has no direct access to market or account data, every round includes at least one call to TradingAgent, which provides the market and portfolio view needed to ground the decision. The Manager may then call NewsAgent, CoderAgent, AnalystAgent, or CriticAgent when the current decision requires recent-event evidence, quantitative verification, reconciliation of conflicting evidence, or an explicit downside challenge.

Because these specialists produce different types of outputs, the runtime converts each response into a common source-attributed evidence item. Each item records a unique identifier \(\text{id}_j\), its source specialist \(\text{agent}_j\), a concise summary \(\text{summary}_j\), the specialist's recommendation or implication for the current decision \(\text{recommendation}_j\), the associated risks \(\text{risks}_j\), and its overall stance toward the candidate trade \(\text{stance}_j\):
\[
e_j=(\mathrm{id}_j,\mathrm{agent}_j,\mathrm{summary}_j,
\mathrm{recommendation}_j,\mathrm{risks}_j,\mathrm{stance}_j).
\]
The resulting evidence items are returned to the Manager, which may request additional specialist input if important information is still missing or proceed to a final decision once the available evidence is sufficient. When finishing, the Manager identifies the evidence supporting its decision and produces an ordered \texttt{execution\_plan}; the corresponding evidence identifiers are retained with the final decision. The Manager is allocated a budget of 100 steps, each specialist invocation is limited to 20 model turns, and NewsAgent may issue at most three search calls per invocation. If no valid plan is produced within the available budget, the round falls back to a conservative no-trade decision.

The completed plan is then passed to ExecutionAgent together with the decision basis, current portfolio, and market prices. ExecutionAgent processes the plan items in their specified order through \texttt{execute\_trade}; a single plan may therefore produce multiple trade calls, while an empty plan completes without placing a trade. After all plan items have been processed, ExecutionAgent emits \texttt{<TRADE\_DONE>}, completing the decision round.

The \textit{decision trace} retains manager-to-specialist calls, specialist responses and evidence items, evidence identifiers used in the final decision, the resulting execution plan, and the subsequent execution actions. These recorded interactions provide the inputs for the \textit{Role Diversity} and \textit{Evidence Integration} metrics defined in Appendix~\ref{app:metric_collaboration}. Complete role prompts and output contracts are provided in Appendix~\ref{app:multi_agent_prompts}.
\subsection{Capability-Specific Evaluation Details}
\label{app:capability_metrics}
This section specifies the definition and computation procedures for the capability-specific dimensions introduced in Section~\ref{sec:capability_metrics}. Judge prompts and scoring rubrics are provided separately in Appendix~\ref{app_prompts}.

\subsubsection{Tool-Use}
\label{app:metric_tool}
\paragraph{LLM-as-a-judge evaluation.} We evaluate each complete Tool-Augmented decision trace along the following four dimensions:

\begin{itemize}
    \item \textit{Tool Relevance (TR)}: Measures whether the selected tools address the information needs of the current decision.

    \item \textit{Invocation Efficiency (IE)}: Measures whether tool calls are made at appropriate times without unnecessary repetition, premature stopping, or wasteful use of the tool budget.

    \item \textit{Information Coverage (IC)}: Measures whether the trace gathers sufficient evidence to support the decision.

    \item \textit{Evidence Faithfulness (EF)}: Measures whether the agent's subsequent reasoning and decision are supported by the returned tool outputs, rather than relying on unsupported or contradictory claims.
\end{itemize}

The judge prompt is presented in Appendix~\ref{app:tool_judge_prompt}, with each dimension rated on a $0$--$10$ scale. Each raw judge rating is divided by 10 to obtain the corresponding metric on $[0,1]$. Throughout the evaluation, \textit{TR}, \textit{IE}, \textit{IC}, and \textit{EF} denote these normalized scores. For each backbone, we report the mean score for each dimension across all evaluated Tool-Augmented traces. We use Qwen3.5-397B-A17B as the primary judge; DeepSeek-V4-Pro evaluates the same traces for the cross-judge robustness analysis in Appendix~\ref{app:detailed_tool_use_results}.

\paragraph{Objective evaluation.}
Objective diagnostics are computed by replaying the recorded tool traces.

\begin{itemize}
    \item \textit{Valid Call Rate (VCR).}
    VCR measures the proportion of evaluated tool calls that are valid and effective. A call is considered invalid or ineffective if it contains invalid parameters, returns an empty or erroneous result, produces an invalid output, or repeats an identical no-op invocation. We compute VCR separately for each trace as the fraction of calls that are not invalid or ineffective, and then macro-average the trace-level scores across all traces for a backbone.

    \item \textit{Hallucination-Free Rate (HFR).}
    HFR measures the proportion of evaluated tool calls that do not invoke hallucinated tools. A call is considered hallucinated if it invokes a tool name that is absent from the resolved tool schema. We compute HFR separately for each trace as the fraction of calls that are not hallucinated, and then macro-average the trace-level scores across all traces for a backbone.

    \item \textit{Routing Quality (RQ).} RQ measures the trading relevance of tools retrieved by the routing mechanism. The routing candidate space contains $305$ tools: 295 interfaces reproduced from public API specifications and 10 core trading tools. The routing meta-tool \texttt{select\_tools} only generates routing requests and is excluded from both the candidate pool and RQ computation. Each candidate tool is assigned a prior trading-relevance score from 0 to 4, corresponding to \textit{very poor}, \textit{poor}, \textit{average}, \textit{good}, and \textit{very good relevance}. For each routing request, we exclude the five always-available interfaces for market snapshots, price history, account state, decision history, and trade execution, since these tools do not need to be retrieved. Let $\widetilde{S}_r$ denote the remaining retrieved tools and $q(u)$ the relevance score of tool $u$. The request-level score is defined as \[ \mathrm{RQ}_r = \frac{1}{4|\widetilde{S}_r|} \sum_{u\in\widetilde{S}_r} q(u), \] which normalizes the average relevance of the retrieved tools to $[0,1]$. Requests that return no non-essential tool are excluded from request-level scoring. For each trace, we average RQ over its scored routing requests; a trace with no scored routing request is assigned a score of zero. The backbone-level score is then obtained by macro-averaging the trace-level scores, and is reported on $[0,1]$.
\end{itemize}

We additionally report \textit{calls per assistant step} and \textit{total token consumption} as execution-cost statistics. 

\paragraph{\textit{Tool Call Score} summary.}
For the overall capability profile, we compute \textit{Tool Call Score (TCS)} for each evaluated trace as
\[
J=\frac{\mathit{TR}+\mathit{IE}+\mathit{IC}+\mathit{EF}}{4},\qquad
\mathit{TCS}=0.55J+0.20\mathit{RQ}+0.15\mathit{VCR}+0.10\mathit{HFR}.
\]
This weighted sum defines the normalized \textit{TCS}. Backbone-level \textit{TCS} is the arithmetic mean over evaluated traces. The individual judge dimensions and objective metrics are reported separately to preserve interpretability.

\subsubsection{Persistent Memory}
\label{app:metric_memory}

\paragraph{LLM-as-a-judge evaluation.}
We evaluate persistent-memory behavior along two complementary dimensions:
\begin{itemize}
    \item \textit{Memory Content Quality (MCQ).}
    \textit{MCQ} measures whether stored experience is reusable and informative. Each memory is rated from $0$ to $5$ along three aspects: \textit{generalizability}, which assesses whether the extracted rule can transfer to future situations without depending on a specific date or price; \textit{information density}, which assesses whether the memory captures a complete condition–observation–rule relationship; and \textit{diversity contribution}, which assesses whether it adds information not already represented by other memories. \textit{MCQ} is computed by averaging the three aspect scores and normalizing the result to \([0,1]\).

    \item \textit{Memory Usage Quality (MUQ).}
    \textit{MUQ} measures whether the agent retrieves memory when it is useful and subsequently makes use of the retrieved information. It evaluates two aspects of each retrieval event: \textit{retrieval timing}, which assesses whether memory retrieval is appropriately triggered given the context available beforehand; and \textit{memory adoption}, which assesses whether the retrieved memory is subsequently incorporated into the agent’s reasoning or trading action. The two aspect scores are averaged and normalized to \([0,1]\) to obtain \textit{MUQ}.
\end{itemize}

Reported \textit{MCQ} and \textit{MUQ} scores are equally averaged across GPT-5.4, DeepSeek-V3.2-Thinking, and Qwen3-Max as judges. Complete judging rubrics are provided in Appendices~\ref{app:memory_content_prompt} and \ref{app:memory_usage_prompt}.

\paragraph{Objective evaluation.}
We additionally measure whether persistent memory changes downside risk relative to the matched memory-free configuration under the same market trajectory. We use Maximum Drawdown (MDD) and TailLoss as defined in Appendix~\ref{app:performance_metrics}.

\begin{itemize}
    \item \textit{Normalized Change in Maximum Drawdown 
    ($\Delta \textit{MDD}_{\mathrm{norm}}$).}
    We measure the reduction in maximum drawdown from the memory-free configuration to the memory-enabled configuration and min--max normalize this change across the evaluated backbones to $[0,1]$. A larger value indicates a greater reduction in the worst-case decline.

    \item \textit{Normalized Change in Tail Loss 
    ($\Delta \textit{TailLoss}_{\mathrm{norm}}$).}
    We measure the improvement in TailLoss from the memory-free configuration to the memory-enabled configuration and likewise min--max normalize this change across the evaluated backbones to $[0,1]$. Since TailLoss is a signed return, a larger value indicates smaller losses in the worst-performing intervals.
\end{itemize}

\paragraph{\textit{Memory Score} summary.}
We compute \textit{Memory Score (MS)} as the equal-weight average of the four memory metrics: \textit{MCQ}, \textit{MUQ}, $\Delta \textit{MDD}_{\mathrm{norm}}$, and $\Delta \textit{TailLoss}_{\mathrm{norm}}$. The resulting score lies in $[0,1]$ and is reported in Table~\ref{tab:exp_memory_judge_scores} as an aggregate summary. The main capability profile in Figure~\ref{fig:main_capability_results}(b) reports \textit{MUQ} to characterize memory utilization.

\subsubsection{Rule Following}
\label{app:metric_rule}

We evaluate the Rule Following configuration from two complementary perspectives:
whether the agent follows the hierarchical policy in its executed actions, and whether its reasoning demonstrates correct understanding of the applicable rules.
The policy follows the priority order $R0 > R1 > R2$, where R0 and R1 define hard constraints and R2 defines lower-priority preferences.

\paragraph{Objective evaluation.}
We evaluate realized rule following with a deterministic checker using two component metrics and one aggregate execution score.

\begin{itemize}
    \item \textit{Hard-rule Pass Rate (HPR).}
    \textit{HPR} measures compliance with the R0 and R1 hard constraints.
    A decision round is counted as a pass only when all applicable R0 and R1 constraints are satisfied.
    We compute \textit{HPR} as the fraction of evaluated decision rounds that pass all applicable hard rules.

    \item \textit{Soft-rule Satisfaction (SRS).}
    \textit{SRS} measures how well executed trades satisfy the six lower-priority R2 preferences.
    For each trading decision, we compute the fraction of R2 rules satisfied.
    These decision-level scores are first averaged within each decision round and then macro-averaged across all rounds containing at least one trade.
    Rounds without trading decisions are excluded, and each included round receives equal weight regardless of the number of trades it contains.
\end{itemize}

\paragraph{\textit{Rule Satisfaction Score} summary.}
For the main evaluation of realized rule following, we combine hard-rule compliance and soft-rule satisfaction into a single \textit{Rule Satisfaction Score (RSS)}.
Because R0 and R1 are non-negotiable constraints, hard-rule compliance acts as a gate on the contribution of the lower-priority R2 preferences:
\[
\mathrm{RSS} = \mathrm{HPR} \times \mathrm{SRS}.
\]
The resulting score lies in $[0,1]$.
A high \textit{RSS} therefore requires both reliable compliance with the hard constraints and strong satisfaction of the soft preferences.
We report \textit{HPR} separately to preserve visibility into hard-rule failures, while \textit{RSS} serves as the primary execution-level summary in the main results.

\paragraph{LLM-as-a-judge evaluation.}
Beyond executed actions, an independent LLM auditor evaluates whether the agent correctly understands and reasons about the hierarchical policy along two dimensions:

\begin{itemize}
    \item \textit{Rule Coverage (RC).}
    \textit{RC} measures whether the agent correctly identifies the applicable R0--R1 constraints and R2 preferences, uses the correct rule identifiers, accurately interprets their compliance status, and avoids unsupported violation claims.

    \item \textit{Conflict Handling (CH).}
    \textit{CH} measures whether the agent recognizes relevant conflicts among applicable rules, respects the prescribed hierarchy $R0 > R1 > R2$, and resolves trade-offs without satisfying a lower-priority preference at the expense of a higher-priority requirement.
\end{itemize}

Both \textit{RC} and \textit{CH} are scored from $1$ to $10$ for each evaluated decision round and then averaged across rounds for each backbone.
Qwen3.5-397B-A17B is used as the primary auditor.
DeepSeek-V4-Pro and GPT-5.5 evaluate the same records in the cross-auditor robustness analysis in Appendix~\ref{app:rule_compliance_analysis}.
The complete auditor prompt and rubric are provided in Appendix~\ref{app:app_prompts_rule}.

\paragraph{\textit{Rule Auditability Score} summary.}
For a compact summary of demonstrated rule understanding, we define the \textit{Rule Auditability Score (RAS)} as the normalized average of the two LLM-audited dimensions:
\[
\mathrm{RAS} = \frac{\mathrm{RC} + \mathrm{CH}}{20}.
\]
The resulting score lies in $[0,1]$.
\textit{RAS} is used as the primary reasoning-level summary, while \textit{RC} and \textit{CH} are retained as diagnostic sub-scores for analyses of specific reasoning failures.

\subsubsection{Multi-Agent Collaboration}
\label{app:metric_collaboration}

\paragraph{Objective evaluation.}
We evaluate completed Multi-Agent decision traces along two dimensions:
\begin{itemize}
    \item \textit{Role Diversity (RD).}
    \textit{RD} measures the breadth and balance of specialist participation. We score four roles: \texttt{TradingAgent}, \texttt{NewsAgent}, \texttt{AnalystAgent}, and \texttt{CriticAgent}. For a trace invoking $u$ distinct scored roles, let $p$ be the share of scored-role calls assigned to the most frequently invoked role. When $u>1$,
    \[
        \textit{RD}=\frac{u-1}{3}\min\!\left(1,\frac{1-p}{0.2}\right).
    \]
    The first factor rewards broader role participation; the second penalizes concentration only when one role accounts for more than $80\%$ of scored-role calls. Traces invoking at most one scored role receive zero. \texttt{CoderAgent} is excluded from this component.

    \item \textit{Evidence Integration (EI).}
    \textit{EI} measures how evidence from specialist agents is incorporated into the manager's final decision. Let $a$ be the number of distinct agent sources whose evidence is accepted for consideration, and $b$ the number of these sources cited in the final decision. We average source coverage and cross-source integration:
    \[
        \textit{EI}=\frac{1}{2}\left(\frac{b}{a}+\frac{b-1}{a-1}\right).
    \]
    The coverage term measures how much of the available source evidence is used, while the integration term rewards combining evidence from multiple sources. We set the first term to zero when $a=0$, and the second to zero when $a\leq1$ or $b\leq1$. Accepted \texttt{CoderAgent} evidence is eligible for this component.
\end{itemize}

Both metrics lie in $[0,1]$. For each backbone, we report the arithmetic mean of each metric across its completed evaluated traces.

\paragraph{\textit{Observable Collaboration Score (OCS)} summary.}
For the compact capability profile, \textit{OCS} is the equal-weight average of trace-level \textit{RD} and \textit{EI}, then averaged across completed evaluated traces for each backbone. It lies in $[0,1]$ and quantifies role participation and source use, rather than the semantic accuracy of specialist analysis. The component scores characterize collaboration patterns; the paired trading results describe their effects along the observed market trajectory. Sensitivity to the role-dominance threshold and role-set convention is reported in Appendix~\ref{app:multi_agent_results}.

\section{Supplementary Experimental Results}
\label{app_results}
\label{app:supplementary_experimental_results}

This appendix provides supplementary analyses supporting the main experimental results. We summarize the market conditions and realized trading outcomes, followed by a temporal comparison of outcome and capability rankings and capability-specific analyses.

\subsection{Market Conditions During the Evaluation Period}
\label{app:market_conditions}

\begin{figure}[t]
\centering
\includegraphics[width=0.55\linewidth]{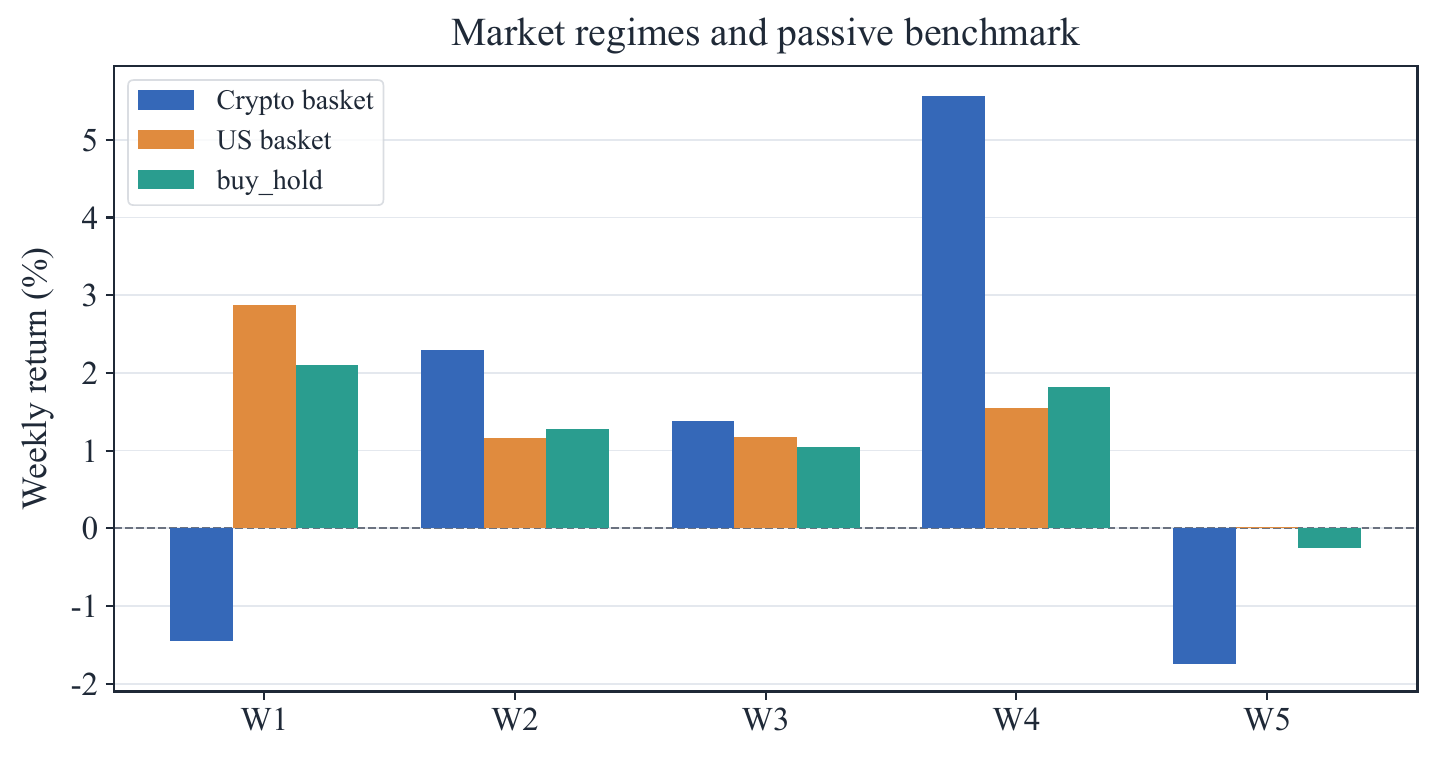}
\caption{
Weekly market returns during the evaluation period for the cryptocurrency basket,
U.S. equity basket, and buy-and-hold reference.
}
\label{fig:market_weekly}
\end{figure}

Figure~\ref{fig:market_weekly} summarizes market conditions during the evaluation period. Both cryptocurrency and U.S. equity markets experienced positive and negative stages, while the buy-and-hold reference remained positive over the full period. The changing market conditions provide a non-stationary setting for comparing agent behavior and realized trading outcomes.

\subsection{Trading Outcome Summary}
\label{app:trading_outcome_summary}

Table~\ref{tab:app_backbone_config_returns} reports cumulative returns for each valid backbone--configuration pair over the full evaluation period.
Returns vary across both backbones and configurations, with no single configuration consistently producing the strongest outcome across all models.
Tool Use achieves the highest return for DeepSeek-V3.2 and Grok-4.20, whereas Base performs best for Qwen3-Max and Multi-Agent Collaboration performs best for Gemini-3.1-Pro.
These results reinforce the main-text observation that realized profitability depends on the interaction between backbone, configuration, and market trajectory rather than on any single factor alone.
The unavailable Gemini--Tool Use and Grok--Multi-Agent Collaboration runs are marked as missing.

\begin{table}[t]
\centering
\small
\caption{
Cumulative return (\%) for each evaluated backbone--configuration pair over the full evaluation period.
Missing runs are marked with ``--''.
}
\begin{adjustbox}{max width=\linewidth}
\begin{tabular}{lrrrrr}
\toprule
Backbone & Base (\%) & Tool (\%) & Memory (\%) & Rule (\%) & Multi-Agent (\%) \\
\midrule
DeepSeek & $-0.12$ & 7.09 & 1.78 & 5.77 & 3.60 \\
Grok     & $-2.07$ & 7.79 & 0.84 & 3.09 & -- \\
GPT-5.4  & 3.45 & 5.73 & 1.99 & 5.74 & 5.90 \\
Gemini   & 3.46 & -- & 2.14 & 3.50 & 6.03 \\
Qwen     & 6.83 & 2.55 & 0.33 & 4.42 & 2.44 \\
\bottomrule
\end{tabular}
\end{adjustbox}
\label{tab:app_backbone_config_returns}
\end{table}

\subsection{Detailed Tool-Use Results}
\label{app:detailed_tool_use_results}

\begin{table}[H]
\centering
\small
\caption{Normalized LLM-judge scores (0--1) for tool use: \textit{Tool Relevance (TR)}, \textit{Invocation Efficiency (IE)}, \textit{Information Coverage (IC)}, and \textit{Evidence Faithfulness (EF)}. Judge mean is the average of these four dimensions.}
\begin{tabular}{lccccc}
\toprule
\textbf{Model} & \textbf{Judge mean} & \textit{TR} & \textit{IE} & \textit{IC} & \textit{EF} \\
\midrule
GPT-5.4 & 0.8371 & 0.8988 & 0.6388 & 0.9112 & 0.8994 \\
Qwen3-Max & 0.8547 & 0.8959 & 0.7312 & 0.9012 & 0.8906 \\
DeepSeek-V3.2 & 0.7429 & 0.8118 & 0.5812 & 0.8341 & 0.7447 \\
Grok-4.20 & 0.7254 & 0.7665 & 0.5871 & 0.8085 & 0.7397 \\
\bottomrule
\end{tabular}
\label{tab:tool_judge_scores}
\end{table}

\begin{table}[H]
\centering
\small
\caption{Objective tool-use metrics and aggregate scores: \textit{Valid Call Rate (VCR)}, \textit{Routing Quality (RQ)}, \textit{Tool Call Score (TCS)}, and \textit{Hallucination-Free Rate (HFR)}. All four scores are reported on a 0--1 scale, with higher values indicating better performance. Calls/step and average tokens describe execution cost.}
\begin{tabular}{lcccccc}
\toprule
\textbf{Model} & \textbf{Calls/step} & \textit{VCR} & \textit{RQ} & \textbf{Avg. tokens} & \textit{TCS} & \textit{HFR} \\
\midrule
GPT-5.4 & 4.212 & 0.9983 & 0.9200 & 30.2k & 0.8941 & 1.000 \\
Qwen3-Max & 0.947 & 0.9780 & 0.7364 & 18.6k & 0.8641 & 1.000 \\
DeepSeek-V3.2 & 0.967 & 0.8254 & 0.7846 & 50.9k & 0.7894 & 1.000 \\
Grok-4.20 & 2.197 & 0.6659 & 0.8052 & 21.9k & 0.7599 & 1.000 \\
\bottomrule
\end{tabular}
\label{tab:tool_objective_metrics}
\end{table}

\begin{figure*}[!t]
\centering
\includegraphics[width=0.94\textwidth]{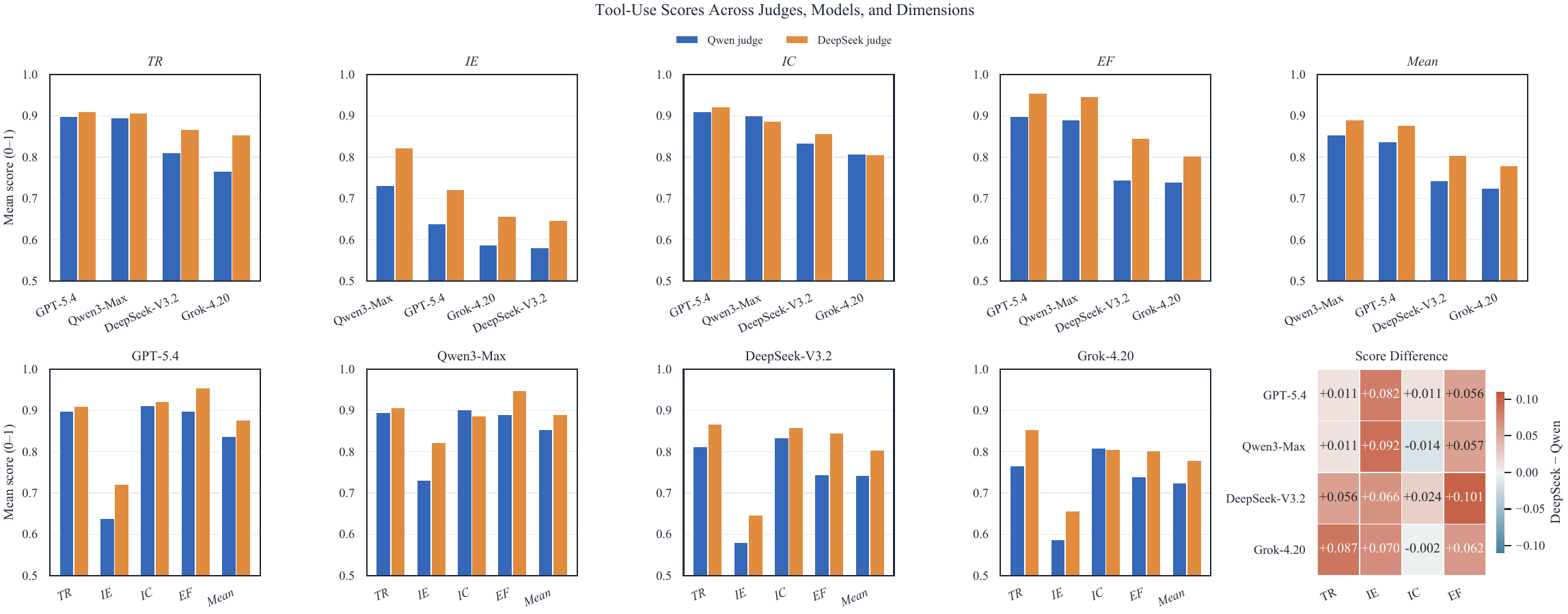}
\caption{Judge-specific scores for the four tool-use dimensions across
Tool-Augmented ReAct agents: \textit{Tool Relevance (TR)},
\textit{Invocation Efficiency (IE)}, \textit{Information Coverage (IC)},
and \textit{Evidence Faithfulness (EF)}. Scores are shown on a 0--1 scale.}
\label{fig:tool_judge_scores}
\end{figure*}

\begin{figure*}[!t]
\centering
\begin{subfigure}[t]{0.48\textwidth}
    \centering
    \includegraphics[width=\linewidth]{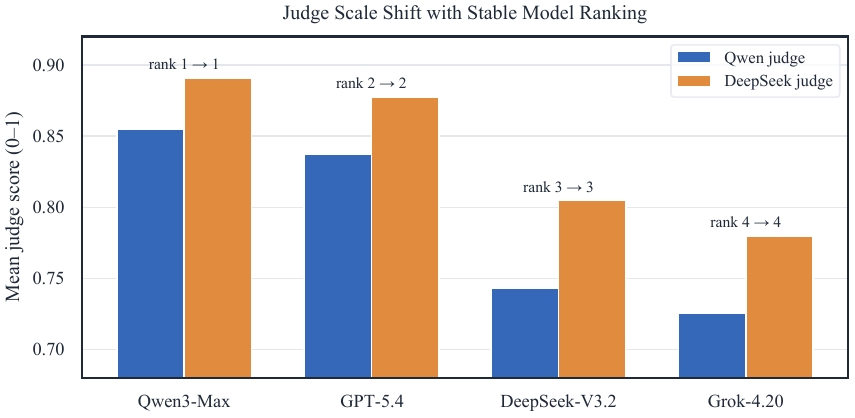}
    \caption{Judge-specific score-scale shift.}
\end{subfigure}
\hfill
\begin{subfigure}[t]{0.48\textwidth}
    \centering
    \includegraphics[width=\linewidth]{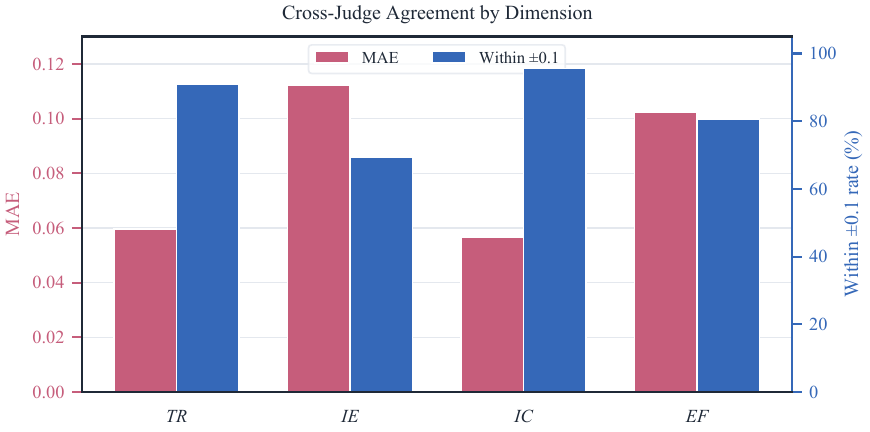}
    \caption{Agreement by evaluation dimension.}
\end{subfigure}
\caption{Cross-judge robustness of tool-use evaluation on the 0--1 scale.
(a) Average judge scores by model. (b) Agreement on \textit{Tool Relevance (TR)},
\textit{Invocation Efficiency (IE)}, \textit{Information Coverage (IC)},
and \textit{Evidence Faithfulness (EF)}, measured by mean absolute error (MAE)
and the fraction of paired scores within 0.1 points.}
\label{fig:tool_judge_robustness}
\end{figure*}

\paragraph{Cross-judge robustness.}
We compare tool-use scores from Qwen3.5-397B-A17B and DeepSeek-V4-Pro
on the same 680 traces (Figure~\ref{fig:tool_judge_scores}).
Figure~\ref{fig:tool_judge_robustness}(a) shows that DeepSeek-V4-Pro assigns
judge means that are 0.0481 points higher on
average, but both judges produce the same model ordering: Qwen3-Max,
GPT-5.4, DeepSeek-V3.2, and Grok-4.20. Absolute scores therefore depend on
judge strictness, whereas the model-level comparison is stable.

Agreement varies across dimensions (Figure~\ref{fig:tool_judge_robustness}(b)).
\textit{Information Coverage (IC)} has a within-\(\pm0.1\) agreement rate
of 95.6\%, followed by \textit{Tool Relevance (TR)} at 91.0\%.
\textit{Invocation Efficiency (IE)} is less stable, with 69.4\% agreement
and a mean absolute error (MAE) of 0.1123; \textit{Evidence Faithfulness (EF)}
reaches 80.7\% agreement with an MAE of 0.1022.
\textit{IE} and \textit{EF} require judging redundancy, marginal information
value, and the dependence of final reasoning on long tool traces, making
them more interpretive than \textit{TR} and \textit{IC}.

\paragraph{Objective calibration and interpretation.}
We interpret the judge scores in Table~\ref{tab:tool_judge_scores} alongside \textit{Routing Quality (RQ)},
\textit{Valid Call Rate (VCR)}, and \textit{Hallucination-Free Rate (HFR)}
(Table~\ref{tab:tool_objective_metrics}). All four models achieve
\textit{HFR} of 1 and have zero dynamic-schema misses, while their
\textit{RQ} and \textit{VCR} vary substantially. Avoiding nonexistent
tools therefore coexists with differences in tool selection and execution
reliability. Together with \textit{IE} and \textit{EF}, these diagnostics
identify whether limitations arise in tool discovery, invocation,
or the use of returned evidence. Calls per assistant step and token
consumption provide complementary information about execution cost.

\paragraph{Excluded Gemini-3.1-Pro run.}
\label{app:gemini_tool_exclusion}
The Tool-Augmented Gemini-3.1-Pro run is excluded from all tool-use and
trading-performance comparisons because it triggered an environment
consistency defect. In the affected trajectory, the agent successfully
submitted a trade and then queried the updated account and market state.
Because state observations lagged behind trade execution, the observation
still reflected the pre-trade positions. The agent interpreted the missing
position update as an execution delay and repeated the same trading action
multiple times. These repeated executions drove the simulator into behavior
not defined by the intended interaction protocol and corrupted the subsequent
account and trajectory metrics. We therefore discard the complete run rather
than treating the resulting behavior as evidence about Gemini's tool-use or
trading capability. The underlying environment-consistency defect has since
been fixed in the platform.

\subsection{Detailed Persistent Memory Results}
\label{app:memory_results}
This section provides supporting evidence for the persistent-memory analysis in Section~\ref{sec:exp_agent_memory}. While the main text summarizes the memory lifecycle from interaction and construction to utilization and downstream effects, we provide here the underlying account-level outcomes, trace analysis of memory construction, temporal comparisons, and judge-level evaluation results.

\subsubsection{Full-Period Performance.} Table~\ref{tab:exp_memory_full_perf} reports the complete 30-day outcomes for the matched w/ Mem and w/o Mem accounts. The most consistent difference appears in downside risk. All five memory-augmented accounts achieve lower \textit{Tail Loss (TL)} than their matched memory-free counterparts: GPT improves from -0.41\% to -0.28\%, DeepSeek from -0.45\% to -0.05\%, Gemini from -0.66\% to -0.61\%, Grok from -1.10\% to -0.27\%, and Qwen from -0.70\% to -0.32\%. Four of five models also improve \textit{Maximum Drawdown (MDD)} and \textit{Consecutive-loss Streaks}, with GPT being the exception for \textit{Maximum Drawdown}.

\begin{figure*}[!t]
\centering
\includegraphics[
    width=0.9\textwidth
]{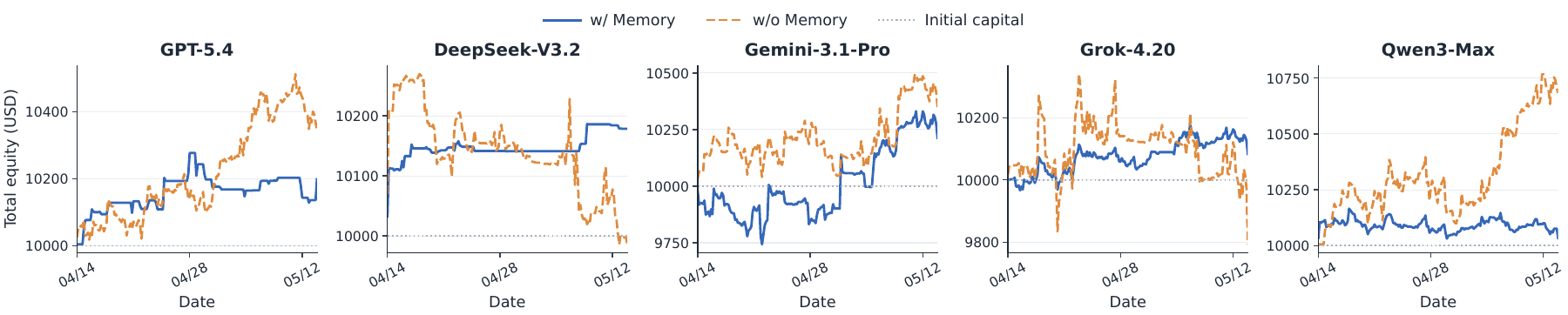}

\caption{
Account-equity trajectories for all ten matched accounts over the full evaluation period.
Solid lines denote \textsc{w/ Mem} and dashed lines denote \textsc{w/o Mem}.
The largest matched downside differences are visible for Grok-4.20 and DeepSeek-V3.2;
}

\label{fig:exp_memory_equity_curves}
\end{figure*}

Return effects are more heterogeneous. Memory improves full-period \textit{PnL} for DeepSeek (+1.78\% vs. -0.12\%) and Grok (+0.84\% vs. -2.07\%), whereas the memory-free accounts obtain higher returns for GPT (+3.45\% vs. +1.99\%), Gemini (+3.46\% vs. +2.14\%), and Qwen (+6.83\% vs. +0.33\%). These results motivate separating downstream risk effects from raw return rather than treating either as a complete measure of memory effectiveness. Figure~\ref{fig:exp_memory_equity_curves} provides the corresponding equity trajectories.

\subsubsection{Memory Construction and Stagnation}
\label{app:memory-stagnation}

The aggregate interaction statistics in Table 1 reveal sharply different memory-construction dynamics despite an identical memory interface. GPT-5.4 performs 332 \texttt{memory\_search} calls but only one \texttt{memory\_add}, leaving a single stored entry after 30 days. Gemini, by contrast, accumulates 93 entries, while DeepSeek, Grok, and Qwen retain 41, 12, and 11 entries, respectively.

Trace analysis helps explain GPT's unusual construction pattern. Its stored rule for the cryptocurrency market can be summarized as \textit{BTC/ETH near recent highs with elevated RSI $\rightarrow$ avoid chasing}. The rule is expressed in terms of a relative market condition rather than a fixed price level. As BTC rises over the trajectory, the reference to a ``recent high'' moves with it, allowing the same rule to remain applicable across changing price levels. In 88\% of the analyzed decisions, the agent concludes that \textit{no additional reusable rule needs to be stored}. GPT therefore provides a concrete example of \textbf{memory stagnation}: \textit{retrieval remains active, while the set of reusable experience ceases to expand}. A similar pattern is observed in the U.S. market, where GPT retrieves a rule favoring modest initial positions in low-drawdown technology leaders supported by AI or earnings catalysts. The model judges the current situation to be covered by this existing rule and therefore considers no additional memory necessary.

Other models exhibit different construction dynamics. DeepSeek and Qwen generate candidate memories more frequently than their final store sizes suggest because similar experiences are removed through deduplication, whereas Gemini continues to accumulate a substantially larger set of entries. These differences show that \textbf{the same persistent-memory interface can induce qualitatively different experience-construction processes across backbones.}

\subsubsection{Temporal Effects}
Because persistent memory evolves throughout the trajectory, we additionally examine outcomes after \textit{April 23}, following at least ten days of memory accumulation. Table~\ref{tab:exp_memory_late_perf} reports the complete late-phase account outcomes.

The relative \textit{Return} effect of memory changes over time for several models. DeepSeek's w/ Mem-w/o Mem PnL differential increases from +1.90 pp over the full period to +2.14 pp in the late phase, while Grok increases from +2.91 to +3.18 pp. Gemini shifts from a -1.32 pp full-period differential to +0.53 pp in the late phase. GPT moves in the opposite direction, from -1.46 to -1.80 pp, while its memory store remains fixed at one entry. Qwen continues to favor the memory-free configuration, although its differential narrows from -6.50 to -4.85 pp. The downside-risk differences also persist into the late phase for most models. These results show that \textbf{the downstream effects of persistent memory are not stationary over the trajectory, motivating evaluation over extended interaction rather than from a single terminal comparison.}

\subsubsection{Judge Robustness}
Table~\ref{tab:exp_memory_judge_scores} reports \textit{Memory Content Quality (MCQ)}, \textit{Memory Usage Quality (MUQ)}, and the aggregate \textit{Memory Score (MS)} for each judge. With the programmatic risk components fixed across judges, the population standard deviation of \textit{MS} across the three judges ranges from approximately 0.014 to 0.038. Model rankings by \textit{MS} are identical across judges (pairwise Kendall's \(\tau=1.0\)), whereas \textit{MUQ} rankings have a mean pairwise Kendall's \(\tau\) of 0.73.

All three judges rank Grok highest and Gemini lowest on \textit{MUQ}, while the relative ordering of GPT, DeepSeek, and Qwen varies. Each judge also assigns Gemini higher \textit{MCQ} than Grok. These patterns support the distinction between memory construction and utilization discussed in Section~\ref{sec:exp_agent_memory}. Table~\ref{tab:memory_lifecycle} reports the judge-averaged \textit{MCQ} and \textit{MUQ} values.

\begin{table*}[!t]
\centering
\caption{
Full-period performance of matched \textsc{w/ Mem} and \textsc{w/o Mem} accounts.
\textit{MDD} denotes \textit{Maximum Drawdown};
\textit{TL} denotes \textit{Tail Loss}, the mean return over the worst-performing $5\%$ of settlement intervals.
\textit{Sharp Losses} and \textit{Sharp Gains} count single-period returns below $-1\%$ and above $+1\%$, respectively, and \textit{Max Loss Streak} denotes the longest sequence of consecutive negative-return periods.
}
\label{tab:exp_memory_full_perf}

\small
\setlength{\tabcolsep}{4pt}
\begin{adjustbox}{max width=\linewidth}
\begin{tabular}{llcccccc}
\toprule
\textbf{Model}
& \textbf{Memory}
& \textbf{Final Return}
& \textbf{MDD}
& \textbf{Sharp Losses}
& \textbf{Sharp Gains}
& \textbf{Max Loss Streak}
& \textbf{TL} \\
\midrule

GPT-5.4
& w/ Mem
& $+$1.99\%
& 1.51\%
& 0
& 0
& 3
& $-$0.28\% \\

& w/o Mem
& $+$3.45\%
& 1.41\%
& 0
& 0
& 5
& $-$0.41\% \\
\midrule

DeepSeek-V3.2
& w/ Mem
& $+$1.78\%
& 0.13\%
& 0
& 1
& 3
& $-$0.05\% \\

& w/o Mem
& $-$0.12\%
& 2.70\%
& 0
& 1
& 6
& $-$0.45\% \\
\midrule

Gemini-3.1-Pro
& w/ Mem
& $+$2.14\%
& 2.36\%
& 1
& 3
& 5
& $-$0.61\% \\

& w/o Mem
& $+$3.46\%
& 2.42\%
& 0
& 2
& 5
& $-$0.66\% \\
\midrule

Grok-4.20
& w/ Mem
& $+$0.84\%
& 1.01\%
& 0
& 0
& 5
& $-$0.27\% \\

& w/o Mem
& $-$2.07\%
& 4.31\%
& 5
& 3
& 6
& $-$1.10\% \\
\midrule

Qwen3-Max
& w/ Mem
& $+$0.33\%
& 1.33\%
& 0
& 1
& 4
& $-$0.32\% \\

& w/o Mem
& $+$6.83\%
& 2.84\%
& 1
& 1
& 5
& $-$0.70\% \\

\bottomrule
\end{tabular}
\end{adjustbox}
\end{table*}
\begin{table*}[!t]
\centering
\caption{
Late-phase performance from April~23 onward, after approximately ten days of memory accumulation.
\textit{TL} denotes \textit{Tail Loss}, and
\textit{MDD} denotes \textit{Maximum Drawdown}.
}
\label{tab:exp_memory_late_perf}

\small
\begin{tabular}{llcccc}
\toprule
\textbf{Model}
& \textbf{Memory}
& \textbf{Return}
& \textbf{TL}
& \textbf{Max Loss Streak}
& \textbf{MDD} \\
\midrule

GPT-5.4
& w/ Mem
& $+$0.56\%
& $-$0.31\%
& 3
& 1.51\% \\

& w/o Mem
& $+$2.36\%
& $-$0.43\%
& 5
& 1.41\% \\
\midrule

DeepSeek-V3.2
& w/ Mem
& $+$0.29\%
& $-$0.04\%
& 2
& 0.10\% \\

& w/o Mem
& $-$1.85\%
& $-$0.48\%
& 6
& 2.23\% \\
\midrule

Gemini-3.1-Pro
& w/ Mem
& $+$2.89\%
& $-$0.51\%
& 5
& 1.56\% \\

& w/o Mem
& $+$2.36\%
& $-$0.66\%
& 5
& 2.42\% \\
\midrule

Grok-4.20
& w/ Mem
& $+$0.24\%
& $-$0.26\%
& 5
& 0.71\% \\

& w/o Mem
& $-$2.94\%
& $-$1.05\%
& 6
& 4.31\% \\
\midrule

Qwen3-Max
& w/ Mem
& $-$0.54\%
& $-$0.33\%
& 4
& 1.00\% \\

& w/o Mem
& $+$4.31\%
& $-$0.73\%
& 5
& 2.84\% \\

\bottomrule
\end{tabular}
\end{table*}
\begin{table*}[!t]
\centering
\caption{
Per-judge persistent-memory scores.
\textit{MCQ} denotes \textit{Memory Content Quality},
\textit{MUQ} denotes \textit{Memory Usage Quality}, and
\textit{MS} denotes \textit{Memory Score}.
For each model and judge, \textit{MS} combines that judge's \textit{MCQ} and \textit{MUQ} with the model's fixed programmatic
$\Delta\mathrm{MDD}_{\mathrm{norm}}$ and
$\Delta\mathrm{TL}_{\mathrm{norm}}$
components defined in Appendix~\ref{app:metric_memory}.
Avg denotes the arithmetic mean across the three judges.
}
\label{tab:exp_memory_judge_scores}

\small
\begin{tabular}{llccc}
\toprule
\textbf{Model}
& \textbf{Judge}
& \textbf{MCQ}
& \textbf{MUQ}
& \textbf{MS} \\
\midrule

GPT-5.4
& GPT
& 1.000
& 0.625
& 0.430 \\

& DeepSeek
& 1.000
& 0.755
& 0.463 \\

& Qwen
& 0.933
& 0.905
& 0.484 \\

& \textbf{Avg}
& \textbf{0.978}
& \textbf{0.762}
& \textbf{0.459} \\
\midrule

DeepSeek-V3.2
& GPT
& 0.927
& 0.630
& 0.700 \\

& DeepSeek
& 0.947
& 0.655
& 0.711 \\

& Qwen
& 0.940
& 0.755
& 0.734 \\

& \textbf{Avg}
& \textbf{0.938}
& \textbf{0.680}
& \textbf{0.715} \\
\midrule

Gemini-3.1-Pro
& GPT
& 0.887
& 0.340
& 0.319 \\

& DeepSeek
& 1.000
& 0.485
& 0.383 \\

& Qwen
& 0.927
& 0.665
& 0.410 \\

& \textbf{Avg}
& \textbf{0.938}
& \textbf{0.497}
& \textbf{0.371} \\
\midrule

Grok-4.20
& GPT
& 0.720
& 0.785
& 0.876 \\

& DeepSeek
& 0.827
& 0.815
& 0.910 \\

& Qwen
& 0.733
& 0.965
& 0.925 \\

& \textbf{Avg}
& \textbf{0.760}
& \textbf{0.855}
& \textbf{0.904} \\
\midrule

Qwen3-Max
& GPT
& 0.833
& 0.620
& 0.589 \\

& DeepSeek
& 0.953
& 0.720
& 0.644 \\

& Qwen
& 0.847
& 0.885
& 0.658 \\

& \textbf{Avg}
& \textbf{0.878}
& \textbf{0.742}
& \textbf{0.630} \\

\bottomrule
\end{tabular}
\end{table*}

\subsection{Additional Rule Following Analysis}
\label{app:rule_compliance_analysis}

\begin{figure*}[t]
  \centering
  \includegraphics[
    width=0.82\textwidth,
    trim=1mm 1mm 1mm 1mm,
    clip
  ]{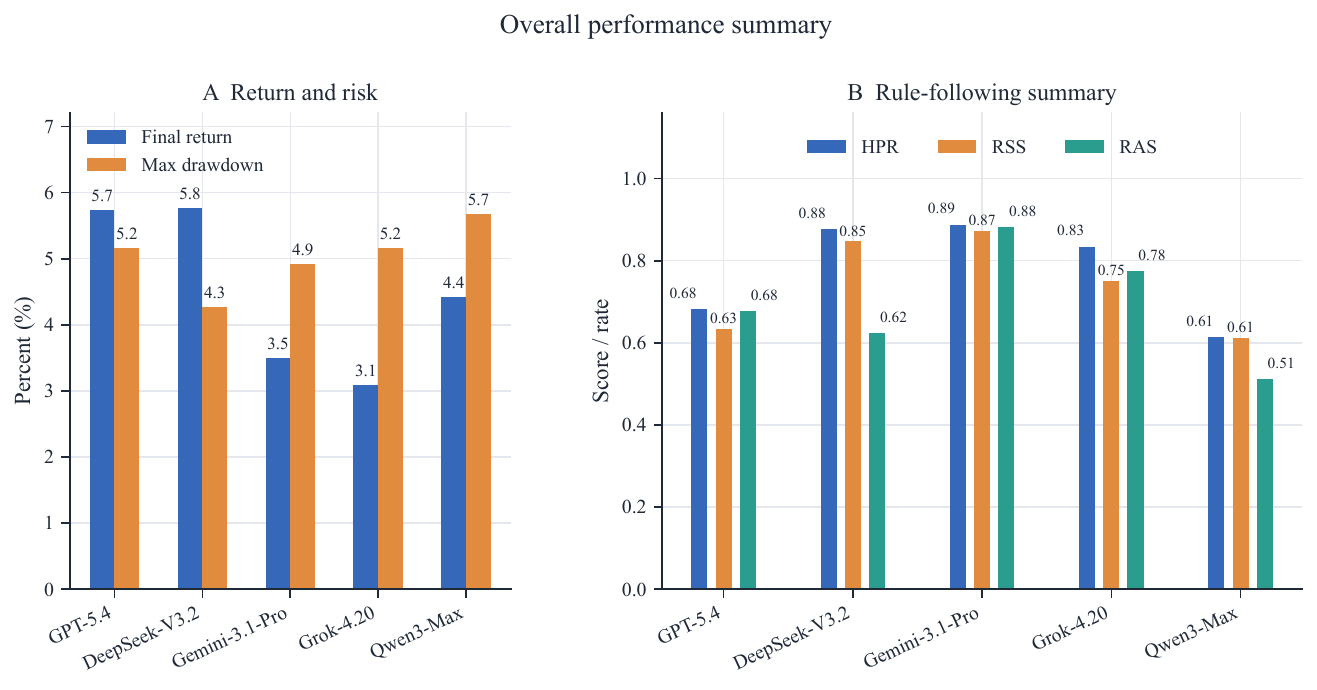}

  \caption{
      Overall trading and rule-following results across models.
      The left panel reports final return and maximum drawdown (MDD);
      the right panel reports \textit{Hard-rule Pass Rate (HPR)},
      \textit{Rule Satisfaction Score (RSS)}, and
      \textit{Rule Auditability Score (RAS)}.
    }
  \label{fig:app_rule_overall_performance}
\end{figure*}

This section reports a cross-auditor robustness analysis and additional rule-level failure statistics and model-level profit--compliance profiles. 
Figure~\ref{fig:app_rule_overall_performance} summarizes the corresponding trading and rule-following metrics.

% \begin{figure*}[t]
%   \centering

%   \begin{subfigure}[t]{0.55\textwidth}
%     \centering
%     \includegraphics[width=\linewidth]{
%       figures/exp_rule/fig2_violation_distribution.pdf
%     }
%     \caption{Rule-specific violations and deviations.}
%     \label{fig:rule_violation_distribution}
%   \end{subfigure}
%   \hfill
%   \begin{subfigure}[t]{0.40\textwidth}
%     \centering
%     \includegraphics[width=\linewidth]{
%       figures/exp_rule/fig3_return_compliance_scatter.pdf
%     }
%     \caption{Profitability and hard-rule safety.}
%     \label{fig:app_rule_return_safety}
%   \end{subfigure}

%   \caption{
%     Rule-specific failure patterns and their relationship to profitability.
%     (a) Hard-rule violations and soft-preference deviations.
%     (b) Final return versus hard-rule failure rate.
%   }
%   \label{fig:app_rule_failure_analysis}
% \end{figure*}

\begin{figure*}[t]
  \centering

  \begin{subfigure}[t]{0.66\textwidth}
    \centering
    \includegraphics[width=\linewidth]{
      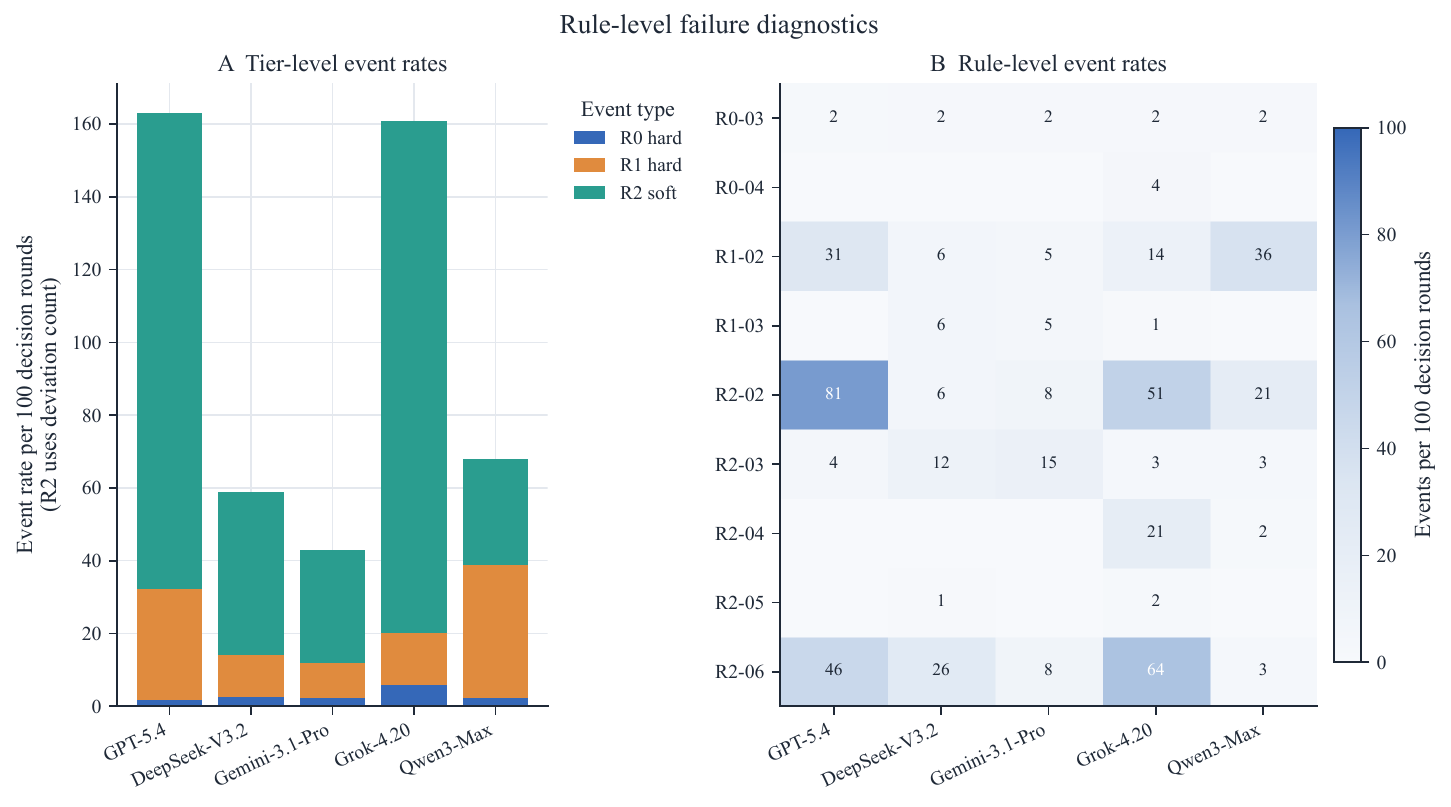
    }
    \caption{Rule-level hard-rule violations and soft-rule deviations.}
    \label{fig:rule_violation_distribution}
  \end{subfigure}

  \par\vspace{0.2em}

  \begin{subfigure}[t]{0.46\textwidth}
    \centering
    \includegraphics[width=\linewidth]{
      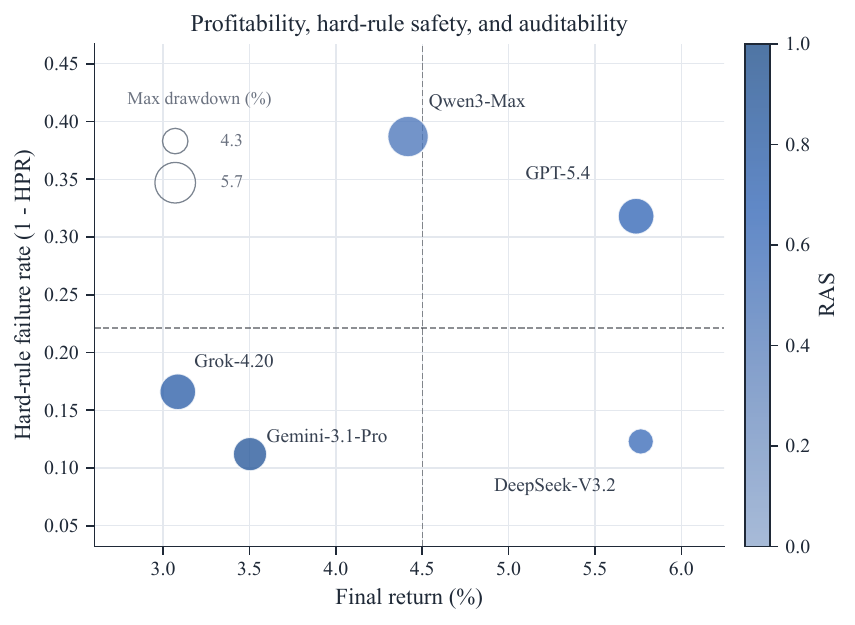
    }
    \caption{Profitability, hard-rule safety, and auditability.}
    \label{fig:app_rule_return_safety}
  \end{subfigure}

  \caption{
    Additional diagnostics of rule-following behavior.
    (a) Rule-level event rates for hard-rule violations and soft-rule deviations,
    highlighting which R0--R2 rules account for the main execution bottlenecks.
    (b) Final return versus hard-rule failure rate, defined as
    $1-\textit{HPR}$, where \textit{HPR} denotes
    \textit{Hard-rule Pass Rate}.
    Marker color represents \textit{Rule Auditability Score (RAS)},
    and marker size represents maximum drawdown (MDD).
  }
  \label{fig:app_rule_failure_analysis}
\end{figure*}

\paragraph{Audit conclusions are broadly consistent across auditor
models.}
The main results use Qwen3.5-397B-A17B without thinking as the primary LLM auditor. To examine whether the conclusions depend on this specific judge, we repeat the same audit procedure using DeepSeek-V4-Pro without thinking and GPT-5.5 with \texttt{reasoning-effort=none}. All three auditors receive the same decision traces, account states, programmatic checker outputs, audit prompt, and scoring rubric.

\begin{table*}[t]
    \centering
    \small
    \setlength{\tabcolsep}{9pt}
    \renewcommand{\arraystretch}{1.12}
    \caption{
        Cross-auditor comparison of \textit{Rule Auditability Score (RAS)}.
        Qwen3.5-397B-A17B and DeepSeek-V4-Pro are evaluated without         thinking, while GPT-5.5 uses \texttt{reasoning-effort=none}. 
        The final column reports the mean \textit{RAS} across the three auditors.
    }
    \begin{tabular}{lcccc}
        \toprule
        \textbf{Audited agent}
        & \textbf{Qwen3.5-397B-A17B}
        & \textbf{DeepSeek-V4-Pro}
        & \textbf{GPT-5.5}
        & \textbf{Mean} \\
        \midrule
        GPT-5.4
        & 0.677
        & 0.744
        & 0.586
        & 0.669 \\

        DeepSeek-v3.2
        & 0.623
        & 0.680
        & 0.516
        & 0.606 \\

        Gemini-3.1-pro
        & \textbf{0.882}
        & \textbf{0.782}
        & 0.645
        & \textbf{0.770} \\

        Grok-4.20
        & 0.775
        & 0.699
        & \textbf{0.652}
        & 0.709 \\

        Qwen3-Max
        & 0.511
        & 0.545
        & 0.428
        & 0.495 \\
        \bottomrule
    \end{tabular}

    \label{tab:rule_cross_auditor_consistency}
\end{table*}

As shown in Table~\ref{tab:rule_cross_auditor_consistency}, the main model-level conclusions remain stable across auditors. Gemini-3.1-pro receives the highest score from Qwen3.5-397B-A17B and DeepSeek-V4-Pro and remains close to the highest score under GPT-5.5, whereas Qwen3-Max is ranked last by all three auditors. 
DeepSeek-V3.2 also consistently receives a lower \textit{Rule Auditability Score (RAS)} despite its strong \textit{Hard-rule Pass Rate (HPR)} and \textit{Rule Satisfaction Score (RSS)}, supporting the distinction between rule execution and auditable rule reasoning.

The average pairwise Spearman rank correlation among the three auditors is \(0.83\). The GPT-5.5 auditor produces the largest shift in absolute score levels and ranks Grok-4.20 slightly above Gemini-3.1-pro. Nevertheless, this variation does not alter the main comparative conclusions, suggesting that the reported audit patterns are not specific to a single auditor model.

\paragraph{Rule-following failures are concentrated in a small number of rule-specific bottlenecks.}
Figure~\ref{fig:rule_violation_distribution} shows that hard-rule violations and soft-preference deviations are concentrated in a small number of rules. 
The dominant hard-rule bottleneck is R1-02, the single-asset concentration limit. GPT-5.4 and Qwen3-Max trigger R1-02 at rates of 31 and 36 events per 100 decisions, respectively.

Trace analysis suggests that these failures do not generally arise from completely ignoring R1-02. Instead, the models inconsistently combine existing positions, incremental orders, and price movements when estimating post-trade exposure. They also sometimes treat the 15\% limit as a target boundary rather than maintaining a safety margin below it.

Among the R2 preferences, R2-02 cash utilization and R2-06 active engagement are the dominant sources of deviation. 
GPT-5.4 and Grok-4.20 trigger R2-02 at rates of 81 and 51 events per 100 decisions, respectively, indicating that their cash ratios frequently fall outside the target range. Grok-4.20 and GPT-5.4 trigger R2-06 at rates of 64 and 46, respectively, reflecting frequent periods of insufficient trading activity.

\paragraph{Different models exhibit distinct profit–rule-following profiles.}
Figure~\ref{fig:app_rule_return_safety}, together with the account traces and trade records, reveals distinct model-level relationships between profitability and rule-following behavior.

DeepSeek-v3.2 relies on frequent rebalancing to keep variables such as the cash ratio within ranges rewarded by the programmatic checker.
However, its lower auditability score \textit{RAS} indicates that mechanical
compliance is not always supported by stable and well-grounded rule
explanations.

Gemini-3.1-pro exhibits the most consistently auditable compliance behavior. Although it does not achieve the highest return, it remains active while effectively handling trade-offs between hard constraints and soft preferences, resulting in the highest rule-satisfaction and auditability scores.

Grok-4.20 follows a more conservative strategy and opens fewer new positions, reducing its overall exposure to hard-rule violations. However, its active entries can still violate constraints such as the single-asset concentration limit.

GPT-5.4 and Qwen3-Max are more profit-oriented and frequently trigger single-asset concentration violations when increasing positions. Both models have difficulty jointly accounting for existing exposure, incremental orders, and price fluctuations. Qwen3-Max, in particular, often treats the 15\% hard limit as a target boundary, contributing to frequent hard-rule violations and the lowest \textit{HPR} and \textit{RSS} among the evaluated models.

\subsection{Additional Multi-Agent Collaboration Analysis}
\label{app:multi_agent_results}

\paragraph{Excluded Grok-4.20 Multi-Agent run.}

The Grok-4.20 Multi-Agent account is excluded from the collaboration and
trading-performance comparisons because it did not produce a complete and
reliable sequence of evaluation checkpoints. During the affected run and
subsequent diagnostic reproductions, some API responses contained extremely
large batches of repetitive tool calls, including a recorded response with
1,363 requested calls to substantially the same function. Although the
harness capped the number of calls per response and removed duplicates, the
behavior recurred in later decision stages, resulting in prolonged timeouts,
retry cycles, and failure to complete the prescribed trading protocol.

Because the resulting trajectory is incomplete, its partial trades and
collaboration events are not included in the reported Multi-Agent statistics.

\paragraph{Role-set sensitivity.}
As a post-hoc sensitivity analysis, we replace CriticAgent with CoderAgent
in the \textit{Role Diversity (RD)} set while leaving
\textit{Evidence Integration (EI)} unchanged. This changes the role component in 176 of 669 trajectories and
shifts the absolute \textit{Observable Collaboration Score (OCS)} values, but preserves the model ordering
DeepSeek-V3.2 $>$ Gemini-3.1-Pro $>$ Qwen3-Max $>$ GPT-5.4
(Table~\ref{tab:app_ocs_robustness}). The principal comparison is therefore
not determined by this role-selection convention.

\begin{table}[H]
    \centering
    \small
    \setlength{\tabcolsep}{5pt}
    \caption{\textit{Observable Collaboration Score (OCS)} robustness. Reported \textit{OCS} uses CriticAgent and excludes
    CoderAgent from role breadth; the alternative reverses that choice.
    The 95\% confidence intervals (CI) apply to the reported \textit{OCS}.}
    \begin{tabular}{@{}lrrrl@{}}
        \toprule
        Model & Traces & Reported \textit{OCS} & Alternative \textit{OCS} & 95\% CI \\
        \midrule
        DeepSeek-V3.2  & 161 & 0.806 & 0.726 & [0.783, 0.830] \\
        Gemini-3.1-Pro & 168 & 0.738 & 0.695 & [0.721, 0.755] \\
        Qwen3-Max      & 170 & 0.516 & 0.485 & [0.475, 0.559] \\
        GPT-5.4        & 170 & 0.265 & 0.265 & [0.255, 0.277] \\
        \bottomrule
    \end{tabular}

    \label{tab:app_ocs_robustness}
\end{table}

\paragraph{Uncertainty and dominance threshold.}
The 95\% confidence intervals in Table~\ref{tab:app_ocs_robustness} are
obtained from 10,000 trajectory-level bootstrap resamples with a fixed
random seed. Varying the role-dominance threshold over
\(\{0.7,0.8,0.9\}\) also preserves the model ordering and changes
model-level \textit{OCS} by at most 0.016. These checks support the stability of the
comparative conclusion while showing that absolute \textit{OCS} values remain
sensitive to metric conventions.

\FloatBarrier
\subsection{Cumulative Ranking Protocol}
\label{app:cumulative_rank_protocol}

\begin{figure}[!ht]
\centering
\includegraphics[width=\linewidth]{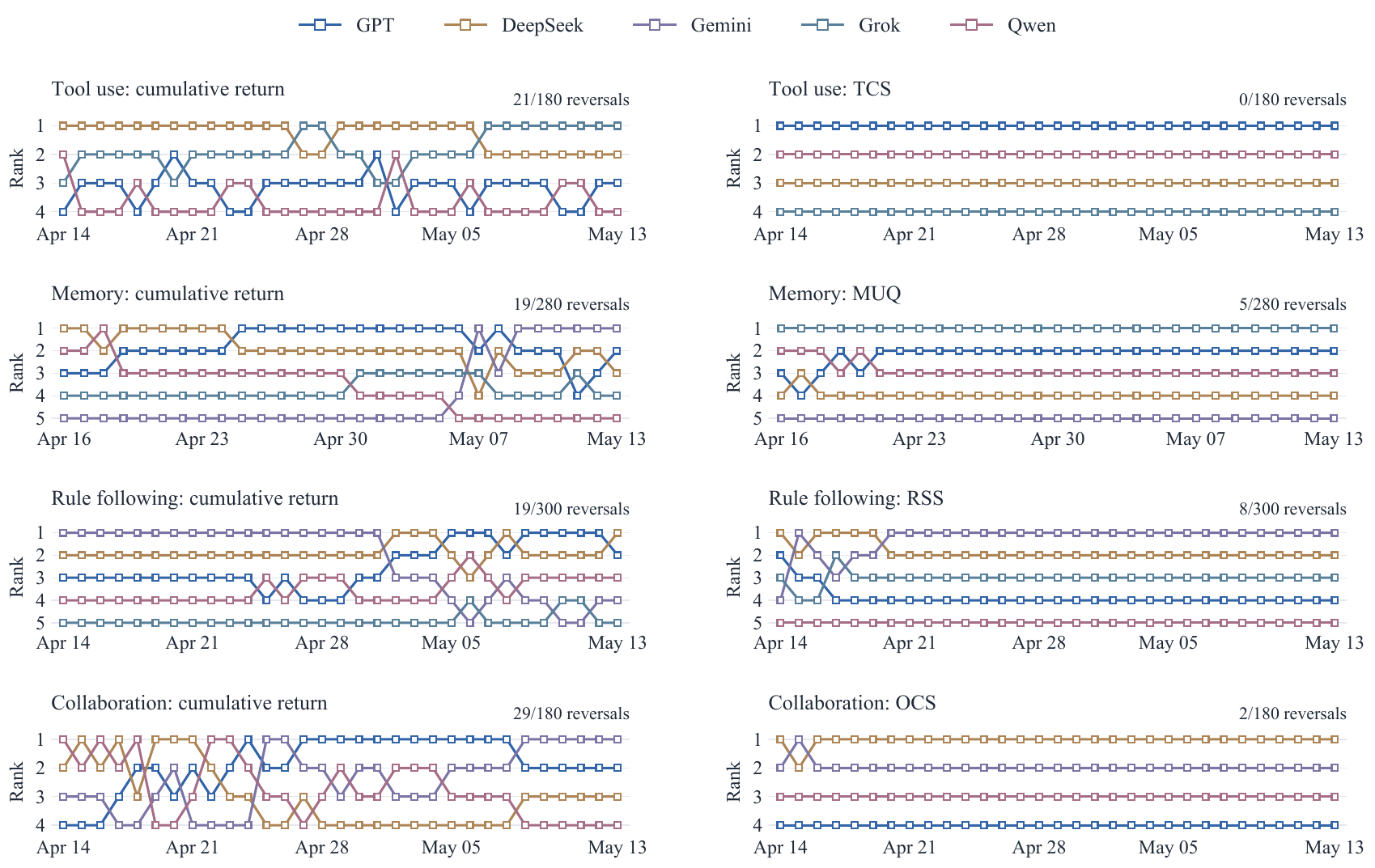}
\caption{Daily cumulative-return and capability rankings for tool use,
memory, rule following, and collaboration. Capability scores are computed
from all available evaluation records from the experiment start through
each observation time: \textit{Tool Call Score (TCS)},
\textit{Memory Usage Quality (MUQ)}, \textit{Rule Satisfaction Score (RSS)},
and \textit{Observable Collaboration Score (OCS)}.
Each row compares matched accounts and observation times; rank 1 is highest.
Annotations count pairwise ranking reversals among comparable pairs.}
\label{fig:supplementary_capability_ranks}
\end{figure}

Figure~\ref{fig:supplementary_capability_ranks} extends the temporal comparisons in Section~\ref{sec:main_results}
to all four capability dimensions, using cumulative records from the experiment start through each observation time. At each observation time, cumulative
return is computed from the latest available hourly account equity relative
to initial equity. \textit{Tool Call Score (TCS)} averages all evaluated traces completed
by that time. Rule satisfaction averages daily scores weighted by the
number of evaluated decisions; each completed day enters at the following
midnight, and the final partial day enters at the final observation.
Memory utilization averages the available evaluated fragments within each
judge, then equally averages the three judges. Fragments are assigned to
observation times using their source trajectory completion times.
Collaboration scores average \textit{Observable Collaboration Score (OCS)} over all completed evaluated traces in
the prefix, with each trace scored as the mean of \textit{Role Diversity (RD)} and \textit{Evidence Integration (EI)}.
The 669 evaluated traces comprise 170 for GPT-5.4, 161 for DeepSeek-V3.2,
168 for Gemini-3.1-Pro, and 170 for Qwen3-Max.

Daily observations run from April 14 through May 13 at 00:00 UTC, followed
by the final observation on May 13 at 16:00 UTC, yielding 31 observations.
Memory comparisons begin on April 16, once every judge has evaluated at
least one fragment for every model, yielding 29 observations. Each
capability--return comparison uses the same models and observation times:
four models each for tool use and collaboration, and five each for memory
and rule following.

A pairwise ranking reversal occurs when two models exchange their relative
ordering between consecutive observations. Comparisons involving ties are
excluded from the reversal denominator; tied scores receive average ranks.
Tool use produces 0 capability-score reversals versus 21 return reversals
across 180 comparable pairs; memory produces 5 versus 19 across 280 pairs;
rule following produces 8 versus 19 across 300 pairs; and collaboration
produces 2 versus 29 across 180 pairs. The overall collaboration profile
uses the full-period mean of the same trace-level \textit{OCS} values.

\section{Case Studies}
\label{app_case}

This appendix presents representative decision traces and case studies
that illustrate the characteristic behaviors, strengths, and failure
modes of the evaluated agent architectures.

\subsection{Tool-Augmented ReAct}
\label{app:case_tool}

\paragraph{Case: Early opportunity discovery and evidence-grounded execution
(GPT-5.4, April~13--27).}
At the first decision round, GPT-5.4 used the routing interface to turn a
broad opportunity search into a full-universe scan. It retrieved the account
state and current snapshots for all 23 allowed assets, downloaded
multi-horizon price histories, used Python to rank momentum, moving-average
position, RSI, drawdown, and volatility, and then validated the leading
candidates with recent news. The trace selected NVDA as the strongest
risk-adjusted setup, citing its multi-horizon uptrend, positive AI-related
news, and confirmation that the U.S. market was open. The agent bought 13
shares at \$188.645, a \$2{,}452.39 position.

\begin{figure*}[!t]
    \centering
    \includegraphics[width=0.96\textwidth]{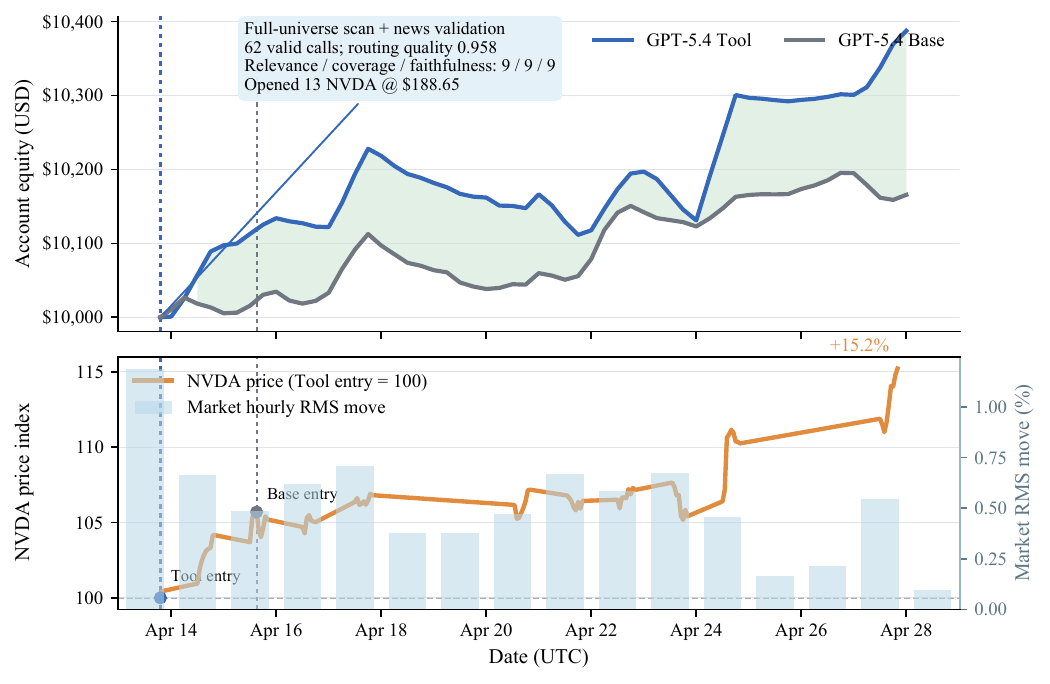}
    \caption{Positive tool-use case study. The upper panel shows the GPT-5.4
    Tool-Augmented and Base ReAct account-equity trajectories. The lower panel
    indexes NVDA to the Tool Agent's April~13 execution price; vertical lines
    mark the Tool and Base entries. Bars report, for each UTC day, the root
    mean square of observed hourly returns across the benchmark asset universe.}
    \label{fig:tool_case_gpt_nvda}
\end{figure*}

The process-level diagnostics confirm that this was a clean routed decision:
all 62 tool calls were unique and valid, with no hallucinated, erroneous, or
no-op calls; routing quality was 0.958. The judge assigned 9/10 for relevance,
9/10 for information coverage, and 9/10 for synthesis/faithfulness. The paired
Base ReAct account did not enter NVDA until April~15, after the price had
already risen to \$199.39, and opened only a \$398.78 position. By April~27,
NVDA had gained 15.2\% from the Tool Agent's execution price. Over the same
window, Tool Agent equity reached \$10{,}387.46 versus \$10{,}165.62 for Base
ReAct, a \$221.84 gap. Figure~\ref{fig:tool_case_gpt_nvda} aligns the two
account trajectories with the realized NVDA path and contemporaneous market
volatility.

\paragraph{Synthesis.}
This episode illustrates the positive pathway from information-gap
decomposition to routing, quantitative comparison, news validation, and
timely execution. The Tool Agent identified the same opportunity earlier and
with greater conviction than its fixed-tool counterpart, and the subsequent
market path rewarded that evidence-grounded decision.

\subsection{Persistent Memory}
\label{app:case_memory}

To complement the aggregate results in Section~\ref{sec:exp_agent_memory}, we analyze the effect of persistent memory on agent behavior at the level of individual tool calls. All figures below are reconstructed from the \texttt{agent\_traces} logs of paired Persistent Memory and Base accounts that run the same backbone under identical market conditions, so that the only controlled difference is access to accumulated memory. The three episodes jointly illustrate the two-sided nature of persistent memory: consistent downside protection in low-conviction or range-bound regimes (Cases~1--2), and a systematic suppression of upside in a trending regime (Case~3). Each account starts from an identical \$10{,}000 balance.

\paragraph{Case 1: Memory-driven volume confirmation (DeepSeek-V3.2, May~7--8).}
Under a regime in which BTC, ETH, and SOL simultaneously exhibited extremely low volume ratios ($<0.01$) with the Fear~\&~Greed index in the fear zone (25--45), the Persistent Memory account retrieved, on two consecutive decision rounds, the same consolidated rule:
\textit{``When BTC/ETH/SOL simultaneously show extremely low volume ratios ($<0.01$), the Fear~\&~Greed index sits in the fear zone, and the macro backdrop is adverse, price action lacks directional conviction and tends to move sideways or downward; wait for volume confirmation ($>0.5\times$ the moving average) before entering, and avoid establishing a directional position until at least two major assets show volume expansion.''}
Acting on this rule, the agent held its entire \$9{,}330 cash balance idle across both rounds rather than chasing the low-volume bounce. The paired Base account instead carried long BTC, SOL, and BNB positions (only \$6{,}711 in cash) and, at 03:00 on May~8, further adjusted its exposure on an anticipated oversold rebound that did not materialize, leaving its holdings under sustained pressure. By 15:00 on May~13 the equity gap had widened to \$185.05 in favor of Persistent Memory (Persistent Memory: \$10{,}178.48; Base: \$9{,}993.43). The episode shows that an accumulated volume-confirmation rule can prevent the agent from opening positions blindly under low-volume, low-conviction conditions.

\paragraph{Case 2: Memory-driven position restraint (Grok-4.20, May~10).}
With BTC up roughly 5\% over 14~days and oscillating near \$79{,}700, the Persistent Memory account retrieved two prior rules:
\textit{``BTC up 5--10\% over 14~days toward \$80k, RSI~$\approx$66, futures-led (spot comparatively weak), approaching resistance under macro uncertainty $\rightarrow$ hold a small position ($\leq$20\%, $1\times$ leverage) to control drawdown''};
and
\textit{``BTC RSI neutral ($\approx$40), up 7--9\% over 14~days and approaching major resistance under macro/geopolitical tension $\rightarrow$ keep the current small position and wait for a breakout or the FOMC reaction.''}
The agent accordingly held, maintaining an $\approx$20\% position (\$2{,}036) with \$8{,}070 in cash untouched. The paired Base account opened additional BTC exposure at an 85\% sizing ratio (\$1{,}766), compressing its cash from \$5{,}200 to \$2{,}078. BTC subsequently ranged between \$79{,}700 and \$80{,}500 and pulled back on May~12, and the concentrated Base exposure left its net asset value under persistent pressure. By 15:00 on May~13 the equity gap had widened to \$221.93 in favor of Persistent Memory (Persistent Memory: \$10{,}106.63; Base: \$9{,}884.70). Here an accumulated ``control size near resistance'' rule suppressed the impulse to chase and protected capital in a choppy regime.

\paragraph{Case 3: Return suppression in a trending market (Qwen3-Max, May~8--9).}
Cases~1--2 show memory protecting capital; Case~3 shows the same mechanism capping returns. Despite issuing 15 \texttt{memory\_add} calls and retaining 11 distinct entries, the Qwen3-Max Persistent Memory account finished with a markedly lower return ($+$0.33\%) than its Base counterpart ($+$6.83\%). The trace data reveal the mechanism. Over May~8--9, SOL exhibited strong upward momentum ($\approx$10\% over 7~days, holding above its MA20 and MA50). The Base account began adding SOL longs at 23:00 on May~8, reasoning from ``strong technical momentum plus the Alpenglow upgrade catalyst,'' and its equity climbed from \$10{,}504 to \$10{,}758 by May~12. Over the same window the Persistent Memory account repeatedly retrieved two rules:
\textit{``When holding a slightly losing crypto position ($-2\%$ to $-4\%$) at $1\times$ leverage, the low leverage removes liquidation risk; hold and wait for the fundamental catalyst to materialize''};
and
\textit{``When holding a slightly profitable position ($+3\%$ to $+8\%$) but RSI~$>80$, the risk--reward ratio becomes asymmetric; consider trimming to lock in profit.''}
Constrained by these rules, the agent continued to hold its ETH position and did not participate in the SOL rally; its equity rose only from \$10{,}069 to \$10{,}100 over the same period.

\paragraph{Synthesis.}
These episodes expose an intrinsic property of the mechanism: what memory accumulates is risk-control experience, not a return-maximizing strategy. The prudent rules it encodes---control drawdown, wait for confirmation, avoid chasing---protect capital effectively in range-bound or declining regimes (Cases~1--2), but in a clearly trending, one-directional advance the same rules systematically cap the upside, as the agent forgoes momentum entries by adhering to its ``do not chase'' rule (Case~3). The value of persistent memory therefore lies in the \emph{consistency of downside-risk control} rather than in dominating the Base configuration across all market regimes---consistent with the aggregate finding that every evaluated model lowers its tail loss under Persistent Memory while absolute returns do not uniformly improve (Section~\ref{sec:exp_agent_memory}). This distinction is reflected in our memory evaluation: Memory Content Quality (MCQ) measures the quality of stored experience, Memory Usage Quality (MUQ) measures whether retrieved experience is subsequently adopted in reasoning or action, and changes in maximum drawdown ($\Delta$MDD) and tail loss ($\Delta$TL) measure downstream risk effects relative to the matched Base configuration. These dimensions separate the quality and use of persistent memory from its realized effect on trading outcomes.

\subsection{Domain-Policy-Guided Rule Compliance}
\label{app:case_rule_aware}

To complement the aggregate analysis in Section~\ref{sec:exp_agent_rule}, we analyze rule-following behavior at the level of individual decision traces. Four episodes are selected to illustrate different combinations of execution-level rule satisfaction and demonstrated auditability, corresponding to the two complementary perspectives captured by the Rule Satisfaction Score (RSS) and Rule Auditability Score (RAS). Within each case, we report one representative round whose trace is fully recorded and whose claimed trades are all verified in the trade ledger; Case~3 also pairs the two models with the highest incidence of R1-02.

\paragraph{Case 1: Silent compliance (DeepSeek-V3.2, May~2).}
Across 30 assistant turns, the agent gathers market data, sells micro positions of SOL (\$3) and ETH (\$2), rebalances \$500 out of BNB, and opens a \$489 BTC position. The programmatic checker records no hard-rule violation and full soft-rule satisfaction, yielding a round-level RSS of 1.0. Yet not one of the thirty turns cites a rule---no [Compliance Audit] block and no R0/R1/R2 identifier appears anywhere in the trace---and the final message is the bare termination token. The auditor assigns a round-level RAS of 0.1, noting that the output contains ``no reasoning, rule citations, or analysis \ldots a complete failure to demonstrate coverage.'' Fourteen of DeepSeek-V3.2's 163 evaluated rounds contain no rule citation in any assistant message; all fourteen have RAS below 0.5. The episode isolates one direction of the execution--auditability separation: an agent can keep every trade inside the hard boundary while leaving no auditable evidence of rule checking.

\paragraph{Case 2: Hallucinated violations and repair trades (DeepSeek-V3.2, April~19).}
The final message here is procedurally exemplary: it enumerates rules with correct identifiers, states the R0 $>$ R1 $>$ R2 priority, and documents three repair trades---selling \$61.88 of ETH ``to reduce from 15.10\% to $\sim$14.58\%'' and \$96.76 of SOL ``from 15.43\% to $\sim$14.57\%,'' redeploying \$169 into BTC. The premise, however, is false: the mechanical checker records zero R0/R1 violations and full R2 satisfaction for this round. The agent asserted ``critical R1-02 violations: ETH (15.10\%) and SOL (15.43\%)'' against a portfolio state that did not contain them, then constructed a coherent compliance narrative around the fabricated breach. The auditor's verdict captures the failure mode: the agent ``severely hallucinated the market state and rule violations \ldots and fabricated specific trade executions to `fix' these non-existent issues'' (round-level RAS = 0.2). Structurally correct auditing is therefore not evidence of reliable rule understanding: the format, the priority reasoning, and the repair logic can all be present while being grounded in an incorrect state estimate. Notably, the phantom repair still ended mechanically compliant---the false premise produced real trades that happened to stay inside the true constraints.

\paragraph{Case 3: Boundary targeting and increment blindness (Qwen3-Max, April~13; GPT-5.4, May~5).}
On the first evaluation day, Qwen3-Max opens ETH sized to exactly the cap---\$1,500 on \$10,000 of initial capital---and by 23:04 the position has drifted to 15.69\% through price appreciation. The trace is unusually articulate about the failure mode: ``I opened this position with a target of exactly 15\% (1500 USD out of 10000 total assets). The issue is that ETH has appreciated by about 5.36\%, causing the allocation to exceed the limit,'' and it observes that the policy ``doesn't specify what to do if price appreciation causes an existing compliant position to exceed the limit.'' The agent trims ETH back ``to exactly 15\%,'' then requests a SOL open ``at 15\% allocation'' followed, in the same round, by a second, identically motivated SOL order sized at \$1,512---again 15\% of equity, on top of the position established moments earlier. The checker records the combined exposure (\$1,284 existing + \$1,512 new, 27.74\% of equity) as an R1-02 violation; no corresponding order or fill appears in the ledger. Having just diagnosed cap-as-target drift in its ETH text, the agent re-commits the same convention on the buy side within the same round: the increment is sized to the cap as if the existing position did not exist. GPT-5.4 exhibits the arithmetic form of the same failure on May~5: ``SOL current exposure is $\sim$13.66\%; adding \$600 keeps SOL below the 15\% cap''---the checker records post-trade exposure of 19.44\%, and a second \$900 order two hours later lifts it to 28.04\%. Both models fail to combine existing exposure, the incremental order, and price movement into a post-trade estimate, and both treat the 15\% ceiling as a target boundary rather than a limit requiring a safety margin. This is precisely the profile that makes R1-02 the dominant hard-rule failure for the two models.

\paragraph{Case 4: Auditable violation repair (DeepSeek-V3.2, May~1).}
At decision time the portfolio actually violates R1-03---the cash reserve stands at 9.92\% against the 10\% minimum---with ETH marginally over the concentration limit at 15.06\%; the round is therefore correctly flagged by the gate. The agent identifies both breaches, quantifies the minimal repair, and executes it in a single \$60 trade: ``Sold \$60 ETH to reduce concentration from 15.06\% to 14.47\% ($\leq$15\%)'' and ``Increased cash ratio from 9.92\% to 10.51\% ($\geq$10\%),'' while reporting the smart-contract theme inside its R2-03 band. The auditor assigns a round-level RAS of 1.0. This is the mirror image of Case~1 and establishes the second direction of execution--auditability separation: a round that begins with a hard-rule violation can still contain a correct, quantified, and state-grounded compliance rationale---precisely the evidence that an execution-only evaluation would not capture.

\paragraph{Synthesis.}
The four episodes make the statistical separation in Section~\ref{sec:exp_agent_rule} concrete. High execution-level rule satisfaction can coexist with very low auditability (Case~1); a structurally complete compliance rationale can be built on a hallucinated state (Case~2); fluent rule vocabulary cannot substitute for correct post-trade arithmetic (Case~3); and a real violation can be identified, quantified, and repaired in a fully auditable way (Case~4). Rule following therefore requires two complementary competencies: policy-consistent execution, summarized by HPR and RSS, and state-grounded, auditable rule reasoning, summarized by RC, CH, and RAS. Strength in either dimension alone does not guarantee strength in the other.

\subsection{Multi-Agent Collaboration}
\label{app:multi_agent_case_study}

To complement the aggregate results in Section~4.3.4, we examine three
representative decision rounds that expose distinct ways in which the manager
uses specialist agents. The cases cover selective delegation, conflict-driven
plan revision, and a coordinated decision in which surfaced risks were not
translated into restraint.

\paragraph{Case 1: Selective delegation with an effective decision
(GPT-5.4, April 16).}
At 15:00 UTC, the manager first delegated a full-universe scan to TradingAgent.
The specialist identified NVDA as the strongest existing trend position, AAPL
as the best new U.S. equity opportunity, and SOL as the weakest leveraged
crypto holding. The manager accepted this evidence and stopped coordinating
after one specialist call: it found no conflicting evidence to reconcile and
no sizing problem that required another role. It then closed SOL, bought one
share of NVDA at \$198.96, and opened one share of AAPL at \$262.28. Over the
following 24 hours, NVDA rose to \$200.41 and AAPL to \$270.95; the Multi-Agent
account gained 3.63\%, compared with 1.28\% for the paired ReAct account. This
round has $R_t=0$, $E_t=0.5$, and OCS $=0.25$ because only TradingAgent was
invoked. The episode illustrates that the manager can terminate coordination
when a localized information need has already been resolved, and that limited
role breadth does not preclude an effective trading decision.

\paragraph{Case 2: Conflict-driven plan revision (Gemini-3.1-Pro, April 14).}
TradingAgent initially recommended increasing the 2$\times$ BTC long to 45\%
of balance while closing the weaker SOL position. NewsAgent then reported a
different market interpretation---capital rotation toward ETH and SOL---along
with an estimated \$2.8B tax-related liquidity drain on April 15 and crowded
SOL positioning. AnalystAgent reconciled the disagreement by retaining the SOL
exit but rejecting the BTC addition, while CriticAgent characterized the
portfolio's 68\% leveraged crypto exposure as the principal risk and also
recommended reducing ETH. The final plan cited the three later evidence items
against the original technical recommendation: the account closed its full SOL
position at \$83.53, reduced ETH by half at \$2,317.10, and held BTC without
adding. With all four scored specialist roles invoked and all four accepted
evidence items referenced, the round obtains $R_t=1$, $E_t=1$, and OCS $=1$.
The trace makes the role of collaboration concrete: additional specialists did
not merely elaborate the initial proposal; they changed both its rationale and
its executed exposure.

\paragraph{Case 3: Surfaced risks without corresponding restraint
(Qwen3-Max, May 11).}
TradingAgent proposed 2$\times$ long positions in ETH and SOL after identifying
breakouts above \$2,300 and \$95. NewsAgent reported large stablecoin outflows,
the first SOL ETF outflows since February, and imminent CPI-related volatility.
AnalystAgent therefore recommended smaller, lower-conviction entries, and
CriticAgent proposed reducing leverage to 1.5$\times$ and entering only half of
the intended size. The manager nevertheless requested a second technical check
and then opened ETH at 22\% of balance and SOL at 14\%, both at 2$\times$
leverage. Its final decision referenced all five evidence items, but the
execution followed the renewed technical recommendation rather than the
specialists' proposed risk reductions. During the next 24 hours, ETH declined
from \$2,330.50 to \$2,294.50; Multi-Agent equity fell by 0.08\%, while the
paired ReAct account rose by 0.64\%. By the final checkpoint, ETH had fallen to
\$2,256.50 and SOL to \$91.84, and the two accounts ended at \$10,243.69 and
\$10,683.04, respectively. Although the round obtains $R_t=1$, $E_t=1$, and
OCS $=1$, the trace shows that broad role participation and explicit evidence
integration do not ensure that warnings materially constrain the final plan.

\paragraph{Synthesis.}
The three episodes clarify the process differences summarized by OCS. GPT-5.4
uses specialists selectively and can act after a single focused delegation;
Gemini-3.1-Pro uses disagreement to revise the proposed exposure; and
Qwen3-Max records a broad evidence trail but ultimately preserves the more
aggressive technical plan. These behaviors explain why observable
collaboration varies substantially across backbones and why it need not move
monotonically with realized outcome. OCS captures how roles and evidence enter
the decision process, while OI records how the resulting Multi-Agent account
performs relative to ReAct over the evaluation period.

\end{document}